\documentclass[aos,preprint]{imsart}

\RequirePackage{amsthm,amsmath,amsfonts,amssymb}
\RequirePackage[numbers,sort&compress]{natbib}
\RequirePackage[colorlinks,citecolor=blue,urlcolor=blue]{hyperref}%% uncomment this for coloring bibliography citations and linked URLs
\RequirePackage{graphicx}%% uncomment this for including figures
\RequirePackage[OT1]{fontenc}

\usepackage{mlmodern}
\usepackage{latexsym}
\usepackage{mathrsfs}
\usepackage{amssymb}
\usepackage{hyperref}
\usepackage{xcolor}
\usepackage{caption, subcaption}
\usepackage{stmaryrd}
\usepackage{dsfont}
\usepackage{ifthen}
\usepackage{enumerate}
\usepackage{bbm}

\usepackage{graphicx}
\usepackage{url} % not crucial - just used below for the URL 
\RequirePackage[OT1]{fontenc}
\usepackage[numbers]{natbib}
\usepackage{amsthm,amsmath}
\usepackage{graphicx,ifthen}

\startlocaldefs
\newtheoremstyle{example}{\topsep}{\topsep}%
     {}%         Body font
     {}%         Indent amount (empty = no indent, \parindent = para indent)
     {\rmfamily}% Thm head font
     {}%        Punctuation after thm head
     {\newline}%     Space after thm head (\newline = linebreak)
     {\thmname{#1}\thmnumber{ #2}\thmnote{ #3}}%         Thm head spec

   \theoremstyle{example}

\def\Mj{\sum_{i=1}^{M_{j}}\gamma_{i,j}}

\numberwithin{equation}{section}
\theoremstyle{plain}
\newtheorem{thm}{Theorem}[section]

\newtheorem{prop}{Proposition}[section]

\newtheorem{rem}{Remark}[section]

\newtheorem{cor}{Corollary}[section]

\newcommand{\Lower}[2]{\smash{\lower #1 \hbox{#2}}}
\newcommand{\ben}{\begin{enumerate}}
\newcommand{\een}{\end{enumerate}}
\newcommand{\bi}{\begin{itemize}}
\newcommand{\ei}{\end{itemize}}

\newcommand{\stirlingfirst}[2]{\genfrac{[}{]}{0pt}{}{#1}{#2}}
\endlocaldefs

\begin{document}

\begin{frontmatter}
\title{Poisson Hierarchical Indian Buffet Processes-With Indications for Microbiome Species Sampling Models}
%\title{A sample article title with some additional note\thanksref{T1}}
\runtitle{PHIBP}
%\thankstext{T1}{A sample of additional note to the title.}

\begin{aug}
%%%%%%%%%%%%%%%%%%%%%%%%%%%%%%%%%%%%%%%%%%%%%%%
%% Only one address is permitted per author. %%
%% Only division, organization and e-mail is %%
%% included in the address.                  %%
%% Additional information can be included in %%
\author[A]{\fnms{Lancelot}~\snm{F. James}\ead[label=e1]{lancelot@ust.hk}},
\author[B]{\fnms{Juho}~\snm{Lee}\ead[label=e2]{juholee@kaist.ac.kr}}
\and
\author[A]{\fnms{Abhinav}~\snm{Pandey}\ead[label=e3]{abhinav.pandey@connect.ust.hk}}
%%%%%%%%%%%%%%%%%%%%%%%%%%%%%%%%%%%%%%%%%%%%%%
%% Addresses                                %%
%%%%%%%%%%%%%%%%%%%%%%%%%%%%%%%%%%%%%%%%%%%%%%
\address[A]{Department of ISOM, HKUST\printead[presep={,\ }]{e1,e3}}

\address[B]{The Graduate School of AI, KAIST\printead[presep={,\ }]{e2}}
\end{aug}

\begin{abstract}
We introduce the Poisson Hierarchical Indian Buffet Process (PHIBP), a new class of species sampling models designed to address the challenges of complex, sparse count data by facilitating information sharing across and within groups. 
Our theoretical developments enable a tractable Bayesian nonparametric framework with machine learning elements, accommodating a potentially infinite number of species (taxa) whose parameters are learned from data. 
Focusing on microbiome analysis, we address key gaps by providing a flexible multivariate count model that accounts for overdispersion and robustly handles diverse data types (OTUs, ASVs). 
We introduce novel parameters reflecting species abundance and diversity. The model borrows strength across groups while explicitly distinguishing between technical and biological zeros to interpret sparse co-occurrence patterns. This results in a framework with tractable posterior inference, exact generative sampling, and a principled solution to the unseen species problem. 
We describe extensions where domain experts can incorporate knowledge through covariates and structured priors, with potential for strain-level analysis. While motivated by ecology, our work provides a broadly applicable methodology for hierarchical count modeling in genetics, commerce, and text analysis, and has significant implications for the broader theory of species sampling models arising in probability and statistics.
\end{abstract}



\end{frontmatter}
%%%%%%%%%%%%%%%%%%%%%%%%%%%%%%%%%%%%%%%%%%%%%%
%% Please use \tableofcontents for articles %%
%% with 50 pages and more                   %%
%%%%%%%%%%%%%%%%%%%%%%%%%%%%%%%%%%%%%%%%%%%%%%
%\tableofcontents

%%%%%%%%%%%%%%%%%%%%%%%%%%%%%%%%%%%%%%%%%%%%%%
%%%% Main text entry area:

%%%%%%%%%%%%%%%%%%%%%%%%%%%%%%%%%%%%%%%%%%%%%%%%%%%%%%%%%%%%%%%%%%%%%%%%%%%%%

\section{Introduction}
The analysis of sparse, grouped count data presents a fundamental challenge across the data-driven sciences. This work introduces a new Bayesian framework to address this challenge, built for modeling the latent structure and diversity in such data.
The engine of this framework is the Poisson Hierarchical Indian Buffet Process (PHIBP), a construction we develop as a a versatile extension to the theory of species sampling models, providing a flexible new building block for hierarchical analysis. To ground this framework, we apply it to the domain of microbiome analysis, addressing core challenges in modeling ecological diversity from complex count data.
The PHIBP is a hierarchical extension of the Poisson Indian Buffet Process (IBP) of ~\cite{Titsias} (see also \cite{James2017, Lo1982, FoFZhou}), designed for complex random sparse count matrices where information is shared within and across groups. 
In this, it is an analogue to other hierarchical nonparametric models, such as the Hierarchical Dirichlet Process (HDP)~\cite{HDP} for clustering, building on foundational IBP work (e.g., the Bernoulli IBP of~\cite{GriffithsZ}, its hierarchical extension~\cite{Thibaux}, and recent applications~\cite{MasoeroBiometrika}). The HDP, itself a more flexible version of seminal models like STRUCTURE in population genetics~\cite{pritchard2000inference} and Latent Dirichlet Allocation (LDA) in topic modeling~\cite{BleiLDA}, provides a prior for modeling related groups as unique mixtures over a shared, global set of clusters. The PHIBP, in contrast, models each group by the specific subset of features it possesses from a shared, global dictionary. This shifts the modeling paradigm from assigning proportional group membership to cataloging the presence of shared, overlapping attributes.

This paper elaborates on the preliminary work in~\cite{hibp23}, developing the complete theoretical results required for modeling complex real-world phenomena. Specifically, while the primary contribution of this paper is the theoretical analysis and development of this methodological framework, we also outline its interpretation and how it significantly extends modeling capabilities for complex count models arising in recent microbiome and ecological species sampling studies (see, e.g., \citep{balocchi2024bayesian,franzolini2023bayesian,ren2017bayesian,sankaran2024semisynthetic,willis2022estimating}).

Our Bayesian framework is distinguished by its ability to handle a potentially infinite number of species and their parameters, inducing complex dependencies within a sparse count matrix. A key achievement of this work (Proposition~\ref{prop2}, Theorem~\ref{sumsampleprop}; Section~\ref{PoissonHIBPconstruction}) is the derivation of a tractable generative process from components already present in the literature, though not all broadly familiar. This tractability is what makes the PHIBP a versatile building block for domain experts to incorporate covariates and other information into more intricate models, much like the role the Dirichlet process~\cite{Ferg1973} plays in Bayesian nonparametrics (BNP). To achieve these results, we draw on methodologies from Bayesian latent feature models, random occupancy models, and excursion theory (e.g., \cite{James2017,JLP2,kingman1975,Kolchin,Pit97,Pit02}). Pitman's unpublished manuscript~\cite{PitmanPoissonMix} was pivotal in linking the mixed Poisson species sampling framework of Fisher and McCloskey with works such as \cite{JLP2,FoFZhou,Kolchin,Pit97}; while our treatment is self-contained, see \cite{JamesStick} for further details. Our discussion aligns with broader species sampling perspectives (e.g., \cite{IJ2003,Pit96,Pit02,Pit06}) applied to complex microbiome settings. We also address, within a microbiome setting, the important problem of unseen species, a challenge originating with Fisher's answer to Corbet's question on butterfly diversity~\cite{fisher}, which remains a subject of contemporary research~\cite{balocchi2024bayesian}. Our contributions to this topic are detailed as special applications of our results in Section~\ref{posteriorPoissonHIBP}. Finally, our general multivariate compound Poisson results in Section~\ref{PoissonHIBPconstruction} generalize the Negative Binomial framework of~\cite{JMQ}, offering greater flexibility to model the complex overdispersion characteristic of modern microbiome datasets. The framework in~\cite{JMQ} provides an empirical justification for a model in which the number of colonies is distributed according to a Poisson distribution, while the bacterial counts within each colony independently follow the logarithmic series distribution introduced by~\cite{fisher}.
Despite its technical depth, we aim for this work to be accessible, particularly for practitioners in microbiome and ecological studies. We adopt a pseudo-expository style, providing insights into the practical translations of our results before and after the more technical developments. While this work addresses key modeling gaps in these fields, we emphasize that our primary expertise is not in these specific application domains. Our approach is to encourage the integration of these models as components within more sophisticated, domain-specific frameworks, highlighting how our technical developments facilitate their application, much like innovations seen with other foundational BNP processes.

A key innovation of this work is the development of novel computational schemes that operationalize our theoretical results, culminating in a practically implementable Bayesian Machine Learning framework. As a highlight, we apply our framework to the sparse microbiome data of \citep{willis2022estimating}, where only about 86 of 1,425 unique microbial species (ASVs) are common across all samples. This demonstration shows how our principled handling of zeros avoids the well-known sensitivity of diversity estimates to the arbitrary choice of pseudocounts and the statistical artifacts this data perturbation can introduce. We will present further computational results based on both synthetic and real data, along with freely available resources to validate our model's performance.
To motivate the theoretical developments that follow, we first turn our attention to the practical problems our framework is designed to solve. The upcoming section therefore serves as a conceptual guide, detailing how the architecture of the PHIBP, in itself a foundational extension of species sampling models, directly addresses the core challenges of overdispersion, robust diversity estimation, and the prediction of unseen species in microbiome analysis. With this domain-specific context established, we will then proceed to the model's formal mathematical construction in Section~\ref{PoissonHIBPconstruction}.

Throughout we will use the notation \([n]=\{1,2,\ldots,n\}\) for an integer \(n\).

\section{A Hierarchical Bayesian Model for Microbiome Counts}
\label{sec:model_description}

Microbiome datasets present a unique set of statistical challenges, including extreme sparsity, overdispersion, and the formidable problem of species that are present in the ecosystem but unobserved in any sample \citep{franzolini2023bayesian, sankaran2019latent, sankaran2024semisynthetic}. The Poisson Hierarchical Indian Buffet Process (PHIBP) model is particularly well-suited to address these challenges. Here, we detail the core components of our Bayesian nonparametric/machine learning framework, demonstrating how it provides a principled and robust approach for analyzing multivariate species counts across multiple groups. A group \(j\) might represent a geographical location or treatment condition, and may consist of multiple samples, \(M_j\). Our framework makes three primary contributions: (1) a flexible hierarchical structure that captures severe overdispersion, (2) a novel Bayesian definition of ecological diversity that is robust to sparse counts, and (3) a predictive method for quantifying unseen species.

\subsection{A Two-Level Hierarchy of Abundance}

Our model begins with an idealized, infinite catalog of unique species (taxa), which in practice correspond to entities like Amplicon Sequence Variants (ASVs). These are denoted as pairs \( (Y_{l}, \lambda_{l})_{l \geq 1} \), which are the points of a Poisson random measure, where \( \lambda_{l} \) is interpreted as the prior global mean abundance rate of species \( Y_{l} \) across the entire meta-community. From this, the model generates a prior local mean abundance rate, \( \sigma_{j,l}(\lambda_{l}) \), for each species \( Y_{l} \) within each specific group \( j \). This two-step randomization is key to capturing severe over-dispersion. As we will demonstrate, this layered mixed Poisson framework offers significant advantages over more conventional Gamma-Poisson (i.e., Negative Binomial) models—whose use for microbiological counts has origins in the work of \citep{JMQ}—or Dirichlet-Multinomial models, which may not be ideal for capturing the complex variance patterns and extreme sparsity inherent in microbiome species counts.

The link to the data is formed through the observed counts. We assume each species \( Y_{l} \) may appear \( N_{j,l} \in \{0,1,2,\ldots\} \) times in group \( j \). In our full hierarchical construction, these counts can be decomposed further: \( N_{j,l} = \sum_{k} C_{j,k,l} \), where \( C_{j,k,l} \) represents the subcounts corresponding to finer-grained operational taxonomic units (OTUs) of species \( Y_l \). In many applications, these sub-counts are latent. The latent rates \( \lambda_l \) and \( \sigma_{j,l}(\lambda_l) \) are the (dependent) random mean intensities that generate these counts. This structure is central to handling sparse data, as it provides a mechanism to infer the nature of observed zeros. Based on information borrowed across the hierarchy, the model's posterior inference reflects whether a zero count \( N_{j,l}=0 \) likely corresponds to a structural zero (where the posterior for the latent rate \( \sigma_{j,l} \) becomes concentrated at zero) or a sampling zero (where the posterior for \( \sigma_{j,l} \) maintains significant mass at positive values, suggesting the species was present but undetected). After observing the data, we obtain posterior distributions for the latent abundance rates for the \( \varphi=r \) species that were actually observed. We denote the posterior global mean abundance rate as \( H_l \) and the corresponding posterior local mean abundance rate in group \( j \) as \( \tilde{\sigma}_{j,l}(H_l) \). These posterior rates serve as the basis for all subsequent inference.

\subsection{Bayesian Measures of Ecological Diversity}
\label{sec:diversity}

A primary contribution of our work is a novel Bayesian framework for defining and estimating ecological diversity. Our approach is grounded in the model's generative process. We define diversity indices as functionals of the random components of our model. Since the number of realized species, $\varphi$, and their latent rates, $\tilde{\sigma}_{j,l}(H_{l})$, are random variables, the diversity indices are themselves random variables with well-defined prior distributions. This allows one to simulate from the "prior diversity" to understand the model's assumptions before any data is observed. While our model can address traditional measures, as discussed in the literature \citep{RICOTTA2021, willis2022estimating}, its core innovation is this generative definition.

Within this framework, among other possibilities, we define alpha-diversity (within-group diversity) via the Shannon entropy, which is the random variable:
\begin{equation}
\mathscr{D}_{j} := -\sum_{l=1}^{\varphi} \frac{\tilde{\sigma}_{j,l}(H_{l})}{\sum_{t=1}^{\varphi} \tilde{\sigma}_{j,t}(H_{t})} \log\left(\frac{\tilde{\sigma}_{j,l}(H_{l})}{\sum_{t=1}^{\varphi} \tilde{\sigma}_{j,t}(H_{t})}\right).
\label{Shannon1}
\end{equation}
For beta-diversity (between-group diversity), we define a dissimilarity based on the Bray-Curtis index, which is the random variable:
\begin{equation}
\mathscr{B}_{j,v} := \frac{\sum_{l=1}^{\varphi} |\tilde{\sigma}_{j,l}(H_{l})-\tilde{\sigma}_{v,l}(H_{l})|}{\sum_{l=1}^{\varphi} (\tilde{\sigma}_{j,l}(H_{l})+\tilde{\sigma}_{v,l}(H_{l}))}, \quad j\neq v \in[J].
\label{BrayCurtis1}
\end{equation}
For inference, we condition the entire generative model on the observed data. This provides posterior distributions for the latent rates of the $\varphi=r$ species that were realized in the sample. We denote the posterior global mean rate for an observed species $l$ as $H_l$, which carries the subtle interpretation of the rate conditional on that species being present in the sample at least once. The posterior distributions of $\mathscr{D}_{j}$ and $\mathscr{B}_{j,v}$ are then computed by evaluating their definitions using posterior samples of the corresponding local rates, $\tilde{\sigma}_{j,l}(H_l)$.

The power of this formulation lies in its handling of sparse counts. Even if an observed count \( N_{j,l} \) is zero, the posterior for the latent rate \( \tilde{\sigma}_{j,l}(H_l) \) is a full distribution, not a point mass at zero. This allows the model to borrow strength across its hierarchical structure to infer that a species may be present at low abundance, providing a more robust assessment of diversity. As we will show, the posterior distributions for $\mathscr{D}_{j}$ and $\mathscr{B}_{j,v}$ can be sampled exactly. This approach directly mitigates the well-known problem where sparse co-occurrence artificially inflates beta-diversity estimates.

\subsection{Predicting Unseen Species}
\label{unseen}

The underlying infinite species model allows for a principled approach to prediction. The global species catalog, \( (\lambda_{l}, Y_{l})_{l \geq 1} \), may be divided into the set of unseen species, \( (\lambda'_{l}, Y'_{l})_{l \geq 1} \), and the set of observed species, \( (H_{1}, \tilde{Y}_{1}), \ldots, (H_{r}, \tilde{Y}_{r}) \). The challenge of predicting these unseen species is the ultimate manifestation of the sparsity problem. In this work, we derive the predictive distributions of our processes to provide a novel approach to this issue, in the tradition of the foundational work of Fisher, Corbet, and Williams \citep{fisher}. This problem continues to pose many challenges for modern statistical methods \citep{balocchi2024bayesian, jeganathan2021statistical}. We provide tractable distributions for a host of unseen quantities in future samples—either from an existing group \(j\) or a new community \(J+1\)—given the initial observations, as detailed in Section~\ref{predict}.

By applying our model to challenging datasets, including those of the type presented in works such as \citep{willis2022estimating}, we will validate its ability to provide robust diversity estimates, capture complex overdispersion, and offer principled predictions for unseen species, showcasing its utility for modern microbiome research and for other challenging count based problems with similar structure.

\section{The PHIBP Construction}\label{PoissonHIBPconstruction}

In Section~\ref{sec:model_description}, we introduced our model intuitively through its hierarchy of global (\(\lambda_l\)) and local (\(\sigma_{j,l}(\lambda_{l})\)) abundance rates. We now provide the formal generative construction. The framework is built from a hierarchy of Poisson random measures, which define the distributions of the underlying completely random measures (CRMs) that underpin the model. The key innovation is to specify the base measure of a Poisson Indian Buffet Process (IBP) as itself being a CRM, which induces the desired sharing of species across groups.

We begin by recalling the definition of a group-specific Poisson IBP, as discussed in \citep{James2017, Titsias}. For each group \(j\in[J]\), let \(B_{j}\) be a completely random measure defined as 
$B_{j} = \sum_{k=1}^{\infty} s_{j,k} \delta_{w_{j,k}} \sim \mathrm{CRM}(\tau_{j},B_{0})$. Here, the distribution of the jumps \((s_{j,k})\) and atoms \((w_{j,k})\) is specified by a Poisson random measure with mean measure \(\tau_{j}(s)dsB_{0}(d\omega)\). The measure \(B_{0}\) is a finite measure on a Polish space \(\Omega\), which in our final hierarchical construction will itself be specified as a random measure. The atoms are drawn i.i.d. from the normalized measure \(\bar{B}_{0}(dw) := B_{0}(dw)/B_{0}(\Omega)\). The L\'evy density \(\tau_{j}(s)\) on \((0,\infty)\) determines the distribution of the jumps \(s_{j,k}\) and must satisfy \(\int_{0}^{\infty}\min(s,1)\tau_{j}(s)ds<\infty.\) The total mass \(B_{j}(\Omega):=\sum_{k=1}^{\infty}s_{j,k}\) is an infinitely divisible random variable with Laplace exponent \(B_{0}(\Omega)\psi_{j}(t) := \int_{0}^{\infty}(1-e^{-ts})\tau_{j}(s)ds.\)

Furthermore, for a countable number of disjoint sets $(Q_{l})_{l\ge 1}$, $B_{j}(Q_{l})$ are independent with respective laws indicated by the Laplace exponents $B_{0}(Q_{l})\psi_{j}(t)$ for $l=1,2,\ldots.$ Hence the collection of jumps $\{s_{j,k} : w_{j,k} \in Q_{l}, k\ge 1\}$, are independent across each $j,l,$ and specified by Lévy densities $B_{0}(Q_{l})\tau_{j}$, with atoms selected iid proportional to $\mathbb{I}_{\{\omega\in Q_{l}\}}B_{0}(d\omega).$ 
\begin{rem} Throughout we use $\mathscr{P}(\lambda)$ to denote a $\mathrm{Poisson}(\lambda)$ random variable which may otherwise be indexed by $i,j,k,l$ etc.
\end{rem}

Following \citep{James2017,Titsias,FoFZhou}, we define the Poisson IBP for group \(j\) of \(M_j\) customers as a collection of Poisson Processes:
$$
Z^{(i)}_j = \sum_{k=1}^{\infty}\mathscr{P}^{(i)}_{j,k}(\gamma_{i,j}s_{j,k})\delta_{w_{j,k}} \quad \text{where} \quad Z^{(i)}_j | B_j \overset{ind}{\sim} \mathrm{PoiP}(\gamma_{i,j}B_{j}),
$$
and \(B_j \sim \mathrm{CRM}(\tau_j,B_0).\) The term \(\gamma_{i,j}>0\) reflects within-group heterogeneity, and importantly may correspond to time or space. The counts \(\mathscr{P}^{(i)}_{j,k}\) are conditionally independent Poisson variables with means \(\gamma_{i,j}s_{j,k}\).

In analogy to the construction of other hierarchical nonparametric models, such as the hierarchical IBP of Thibaux and Jordan \citep{Thibaux}, the Hierarchical Dirichlet Process \citep{HDP}, and the single-customer ($M_j=1$) Poisson-Gamma-Gamma HIBP \citep{ZhouPadilla2016}, we create our general PHIBP by specifying the base measure \(B_0\) to be a discrete random measure itself. This fosters the sharing of species (atoms) across the \(J\) groups. Specifically, we set \(B_0=\sum_{l=1}^{\infty}\lambda_l\delta_{Y_{l}} \sim \mathrm{CRM}(\tau_0,F_0)\), where \(F_0\) is a non-atomic probability distribution on \(\Omega\). This yields the full three-level PHIBP generative process for \(i\in[M_{j}], j\in[J]\): 
$$
Z^{(i)}_j|B_j \overset{ind}{\sim} \mathrm{PoiP}(\gamma_{i,j}B_{j}), \quad B_{j}|B_{0}\sim \mathrm{CRM}(\tau_{j},B_{0}), \quad B_{0}\sim \mathrm{CRM}(\tau_{0},F_{0}).
$$

It follows that we may write \( B_{j} = \sigma_{j} \circ B_{0} \), in terms of compositions of CRMs, where for a set \( Q \), \( \sigma_{j} \circ B_{0}(Q) := \sigma_{j}(B_{0}(Q)) \). Each \( \sigma_{j} = (\sigma_{j}(s): s \ge 0) \) is a subordinator, that is, a stationary independent increment process with jumps \( (s_{j,k})_{k \ge 1} \) specified by \( \tau_{j} \), and satisfying \( \sigma_{j}(\lambda+s) - \sigma_{j}(s) \overset{d}{=} \sigma_{j}(\lambda) \) for \( s < \lambda \), with Laplace transform
$
\mathbb{E}[e^{-t\sigma_{j}(\lambda)}] = e^{-\lambda\psi_{j}(t)}.
$
This formal construction connects directly to the intuitive rates from Section~\ref{sec:model_description}. The global rates are the jumps of the base measure, \((\lambda_l)\). The local rate for species \(l\) in group \(j\) is the sum of all jumps in \(B_j\) whose atoms correspond to the global species \(Y_l\):
$
\sigma_{j,l}(\lambda_l) := \sum_{k=1}^{\infty}s_{j,k,l}=\sum_{k=1}^{\infty} s_{j,k} \mathbb{I}_{\{w_{j,k}=Y_l\}}
$
where the individual jumps $(s_{j,k,l})_{k\ge 1}$ contributing to this sum are distributionally equivalent to the jumps of $\sigma_{j}$ over an interval of length $\lambda_{l}$, with a corresponding L\'evy density of $\lambda_{l}\tau_{j}$ for each fixed $\lambda_{l}.$ That is, $\sigma_{j,l}(\lambda_{l})\overset{d}=\sigma_{j}(\lambda_{l})$. The individual jumps \(s_{j,k,l}\) can be interpreted as the rates of distinct OTUs, which allows us to formally define the latent OTU frequency measure, $F_{j,l}$:
$$
F_{j,l} = \sum_{k=1}^{\infty} \frac{s_{j,k,l}}{\sigma_{j,l}(\lambda_l)} \delta_{U_{j,k,l}},
$$
where $(U_{j,k,l})\overset{iid}\sim \mathrm{Uniform}[0,1]$ are latent markers for these OTU-level clusters.

Since \(B_0\) is discrete, we can express the conditional measure \(B_j|B_0\) as a sum over the shared atoms: \(B_{j}|B_{0}\overset{d}=\sum_{l=1}^{\infty}\sigma_{j,l}(\lambda_{l})\delta_{Y_{l}}\). This leads to a key mixed Poisson representation for the observed data, which makes the role of the latent abundances explicit:
\begin{equation}\label{eq:mixed_poisson_rep}
(Z^{(i)}_{j},i\in[M_{j}])\overset{d}=(\sum_{l=1}^{\infty}\mathscr{P}^{(i)}_{j,l}(\gamma_{i,j}\sigma_{j,l}(\lambda_{l}))\delta_{Y_{l}},i\in[M_{j}]).
\end{equation}
This representation reveals that the observed counts for species \(Y_l\) arise from a Poisson distribution whose intensity is the product of a sample-specific effect \(\gamma_{i,j}\) and the latent group-level abundance rate \(\sigma_{j,l}(\lambda_l)\), thus formalizing the intuitive model described in Section~\ref{sec:model_description}. While the mixed Poisson representation in \eqref{eq:mixed_poisson_rep} is appealing in terms of interpretative modelling, it presents many challenges from a technical point of view, including those for practical implementation. As a key contribution of this work, we develop and exploit compound Poisson type representations, involving distributions of latent OTU counts and related quantities, which are crucial for the theoretical and practical development of our framework.

Before we proceed to our formal analysis, we mention that the work of \citep{Masoerotrait} produces perhaps the first posterior calculations for a large class of hierarchical latent feature processes based on integer-valued entries, including a Poisson-based variant; those results are also presented in \citep{BerahaFavscaled}. Results of a similar general resolution could be obtained for the PHIBP by a straightforward application of the theory for the multivariate IBP in \citep[Section 5]{James2017}. However, as we shall show, a much more detailed and careful analysis is needed to obtain results yielding the tractable forms and rich interpretability that we develop for the species sampling models discussed here. Specifically, among other things,  we introduce an Allocation process and exploit key compound Poisson vector representations that we develop in this section.  See also \cite{hibp23} for more general results and the idea of coupling our Poisson based HIBP to obtain other spike and slab models including variants of the Bernoulli based HIBP of~\cite{Thibaux}.

Throughout, we will use \( X \sim \mathrm{tP}(s) \) to denote a zero-truncated \(\mathrm{Poisson}(s)\) distribution where \(
\mathbb{P}(X=x|s) = \frac{s^{x} e^{-s}}{(1 - e^{-s})x!}\) for  $x =1, 2, \ldots$ A \(\mathrm{MtP}\) variable is defined by mixing appropriately over \(s\). Multivariate extensions are defined in a similar way, see for instance \citep{Todeschini2020}. All such variables appear in different contexts in the literature. $G_{b}\sim \mathrm{Gamma}(b,1)$ denotes a gamma random variable with shape $b$ and scale $1.$ Additionally, a variable, say \( T_{\alpha}(y) \), is a (simple) generalized gamma random variable if its Laplace transform is of the form 
$
\mathbb{E}[e^{-s T_{\alpha}(y)}] = e^{-y \left( (1+s^{\alpha}) - 1 \right)}
$
for \( 0 < \alpha < 1 \), and otherwise has the density of an exponentially tilted stable variable. This is a commonly used variable and corresponding  process that yields power law behaviour. Note also that $T_{\alpha}(G_{b})\overset{d}=G_{b\alpha}.$ We will show stark contrasts between results for corresponding Gamma and Generalized Gamma processes.

\subsection{Properties of the Poisson IBP}
We now proceed to provide descriptions of the relevant marginal, posterior and predictive distributions. We first begin with an interpretation of some of the relevant results in \cite{James2017}[Sections 3,4.2, 5] for the non-hierarchical setting, which will also help us introduce some key quantities and distributions, where $B_{0}$ is considered to be given. Our description below is based on decomposing the relevant space into a set where the $M_{j}$ customers do not sample from and its complement. This reveals a compound Poisson structure which identifies the appropriate marginal, joint and conditional distributions of the variables. We believe this leads to a more clear description of the results and underlying mechanisms in~\cite{James2017}, which can be applied more broadly.

\begin{prop}\label{IBPpost}
\cite{James2017}[Sections 3,4.2,5]
Let the observed counts for group $j$ be generated as $(Z^{(i)}_j |B_j \overset{ind}{\sim} \mathrm{PoiP}(\gamma_{i,j}B_j), i\in[M_{j}])$, where the latent measure $B_{j}\sim \mathrm{CRM}(\tau_{j},B_{0})$. The jumps of $B_j$, with intensities $(s_{j,k})_{k \ge 1}$, represent the mean abundance rates of corresponding species. The model's structure is derived by thinning these jumps based on whether a species is observed by any customer in group $j$.

\begin{enumerate}
    \item[1. Lévy Density Decomposition via Thinning.] The probability that a species with mean abundance rate $s$ is not observed by any of the $M_j$ customers is $e^{-s\Mj}$. This induces a decomposition of the L\'evy density $\tau_j$ into two parts:
    \begin{itemize}
        \item[i.] The L\'evy density for unobserved species: $\tau^{(M_{j})}_{j}(s) := e^{-s\Mj}\tau_{j}(s)$.
        \item[ii.] The (unnormalized) density for observed species, which gives rise to a proper probability density function (pdf) for their mean abundance rates:
        $f_{S_{j}}(s) = \frac{(1-e^{-s\Mj})\tau_{j}(s)}{\psi_{j}(M_{j})}$, where $\psi_{j}(\Mj)=\int_{0}^{\infty}(1-e^{-s\Mj})\tau_{j}(s)ds$ is the total rate of observed species.
    \end{itemize}

    \item[2. Decomposition of the Latent Measure.] This partitioning of the L\'evy density implies a decomposition of $B_{j}$ itself into an unobserved process and a compound Poisson process of observed species:
    \begin{equation}
    \label{mudecomp}
        B_{j}\overset{d}{=}B_{j,M_{j}}+\sum_{\ell=1}^{\xi_{j}}S_{j,\ell}\delta_{\omega_{j,\ell}}
    \end{equation}
    where $B_{j,M_{j}}\sim \mathrm{CRM}(\tau^{(M_{j})}_{j},B_{0})$, $\xi_{j}\sim \mathrm{Poisson}(\psi_{j}(\Mj)B_{0}(\Omega))$ is the number of observed species, the $(\omega_{j,\ell})$ are their iid locations from $\bar{B}_{0}$, and the $(S_{j,\ell})$ are their iid mean abundance rates from the pdf $f_{S_{j}}(s)$.
    
    \item[3. Marginal Representation of Observed Counts.] The observed count measures are constructed solely from the $\xi_j$ observed species, leading to the marginal representation:
    \begin{equation}
    \label{item:key sum}
        (Z^{(i)}_{j}, i\in[M_{j}])\overset{d}{=}\left(\sum_{\ell=1}^{\xi_{j}}\tilde{C}^{(i)}_{j,\ell}\delta_{\omega_{j,\ell}},i\in[M_{j}]\right).
    \end{equation}

    \item[4. Distribution of Counts for an Observed Species.] For each observed species $\ell \in [\xi_{j}]$, the vector of counts across the $M_j$ customers, $(\tilde{C}^{(1)}_{j,\ell}, \ldots, \tilde{C}^{(M_{j})}_{j,\ell})$, is governed by the following distributions:
    \begin{itemize}
        \item[a.] Conditional on its mean abundance rate $S_{j,\ell} = s$, the count vector follows a multivariate zero-truncated Poisson distribution, $\mathrm{tP}(s\gamma_{1,j},\ldots,s\gamma_{M_{j},j})$, with probability mass function:
        \[
        \mathbb{P}(\tilde{C}^{(1)}_{j,\ell}=c^{(1)}_{j,\ell},\ldots,\tilde{C}^{(M_{j})}_{j,\ell}=c^{(M_{j})}_{j,\ell}|S_{j,\ell}=s) = \frac{s^{c_{j,\ell}}e^{-s\Mj}\prod_{i=1}^{M_{j}}\gamma^{c^{(i)}_{j,\ell}}_{i,j}}{(1-e^{-s\Mj})\prod_{i=1}^{M_{j}}c^{(i)}_{j,\ell}!}
        \]
        for $c_{j,\ell}=\sum_{i=1}^{M_{j}}c^{(i)}_{j,\ell} \ge 1$.
        \item[b.] The joint distribution of the count $\tilde{C}_{j,\ell}=\sum_{i=1}^{M_j} \tilde{C}^{(1)}_{j,\ell}$and the species' mean abundance rate $S_{j,\ell}$ is given by:
        \begin{equation}
        \label{joint1}
        \mathbb{P}(\tilde{C}_{j,\ell}=c_{j,\ell}, S_{j,\ell} \in ds) = \frac{{(\Mj)}^{c_{j,\ell}}s^{c_{j,\ell}}e^{-s\Mj}\tau_{j}(s)ds}{\psi_{j}(\Mj)c_{j,\ell}!}.
        \end{equation}
        \item[c.]$\tilde{C}_{j,\ell}\sim \mathrm{MtP}(\tau_{j},\Mj).$
    \end{itemize}
    
    \item [5. Posterior Distribution.] The posterior of $B_{j}$ given the data $(Z^{(i)}_{j};i\in[M_{j}])$ is characterized by updating only the observed part of the decomposition in \eqref{mudecomp}. The unobserved process $B_{j,M_{j}}$ remains unchanged, while the posterior for each species' mean abundance rate $S_{j,\ell}$ is derived from its joint distribution with the observed counts $(\tilde{C}^{(i)}_{j,\ell}, i\in[M_j])$.
\end{enumerate}
\end{prop}

\begin{rem}
In the non-hierarchical result above the $(\tilde{C}^{(i)}_{j,\ell})$ are considered to be observed counts of OTUs for each individual sample $Z^{(i)}_{j}.$ In the hierarchical scenario that follows, these quantities will be latent.     
\end{rem}

\subsection{Structure of the PHIBP}

Now, in our PHIBP setting, the base measure is itself a discrete random measure, \(B_{0}=\sum_{l=1}^{\infty}\lambda_{l}\delta_{Y_{l}}\sim \mathrm{CRM}(\tau_{0},F_{0})\). We can therefore derive the key distributional properties of our model by applying the logic of Proposition~\ref{IBPpost} conditionally on \(B_0\). Specifically, since the properties of a CRM hold over disjoint sets, we can analyze the process over each atom \(\delta_{Y_l}\) with mass \(\lambda_l\). This leads to the following crucial representations for each group \(j\):

\begin{enumerate}
    \item The total number of latent OTUs in group \(j\), \(\xi_j\), which has a \(\mathrm{Poisson}(\psi_{j}(\Mj)B_{0}(\Omega))\) distribution, can be decomposed as a sum over the contributions from each species: \(\xi_{j}\overset{d}=\sum_{l=1}^{\infty}\xi_{j,l}\), where the \(\xi_{j,l}\) are independent and \(\xi_{j,l}\sim \mathrm{Poisson}(\psi_{j}(\Mj)\lambda_{l})\).
    \item This decomposition of latent OTUs induces a corresponding compound Poisson representation for the observed data, organized by species:
    \[(Z^{(i)}_{j},i\in[M_{j}])\overset{d}=\left(\sum_{l=1}^{\infty}\left[\sum_{k=1}^{\xi_{j,l}}\tilde{C}^{(i)}_{j,k,l}\right]\delta_{Y_{l}},i\in[M_{j}]\right).\]
    \item For each species \(Y_{l}\) and latent OTU \(k \in [\xi_{j,l}]\), the vector of latent counts across samples, \((\tilde{C}^{(1)}_{j,k,l}, \ldots, \tilde{C}^{(M_j)}_{j,k,l})\), and its associated mean abundance rate \(S_{j,k,l}\) are conditionally independent and have the joint distribution described in \eqref{joint1}.
\end{enumerate}

\subsection{The Species Allocation process}
We now provide an analysis for the key \textit{species allocation process} 

\[
\mathscr{A}_{J} \overset{d}{=} \left( \sum_{l=1}^{\infty} \xi_{j,l} \delta_{Y_{l}}, \; j \in [J] \right),
\]
which is a component of $((Z^{(i)}_{j},i\in[M_{j}]), j\in[J]),$ and is a multivariate Poisson IBP in the sense of \cite{James2017}[Section 5]. This process constitutes the allocation of multiplicities of unique species between the \( J \) groups and, as we shall show, depends only on the components of \( B_{0} \) and the samples \( (M_{j}, j \in [J]) \) through \( (\psi_{j}(\Mj), j \in [J]) \). That is to say, regardless of the type of entries that would appear in a nonhierarchical general spike and slab IBP, in this case, counts \( \tilde{C}^{(i)}_{j,\ell} \), as seen in \cite{hibp23}[Proposition 2.2].

The next result describes the specific properties of this multivariate Poisson \(\mathrm{IBP}\) process, which, although not immediately obvious, plays a crucial role throughout, as it serves as a multivariate extension of the discretization process in \cite{Pit97}; see also \cite{James2017}[Remark 4.2]. Our approach utilizes a description based on a method of decompositions, while relying on the results in \cite{James2017}[Section 5] for the multivariate IBP. We will present the result first, followed by a detailed elaboration on its interpretation within a microbiome species sampling context.

\begin{prop}\label{prop2}
Let $\mathscr{A}_{J}\overset{d}{=}(\sum_{l=1}^{\infty}\xi_{j,l}\delta_{Y_{l}}, j\in[J])$ be the species allocation process, where the counts $\xi_{j,l}$ are independent $\mathrm{Poisson}(\psi_{j}(\Mj)\lambda_{l})$ variables conditional on the jumps $(\lambda_l)$ of the baseline CRM, $B_0$. The structure of $\mathscr{A}_{J}$ is determined by a decomposition of $B_0$ based on whether its jump locations (species) are observed in the $J$ samples. This leads to the following properties:

\begin{enumerate}
    \item [1. Decomposition of the Baseline Process.] The thinning of species, based on the event that a species $l$ is not observed in any sample, $\{ \xi_{j,l}=0, \forall j\in[J] \}$, induces a decomposition of the L\'evy density $\tau_0$ of $B_0$:
    \begin{itemize}
        \item[i.] The L\'evy density of jumps corresponding to unobserved species is $\tau_{0,J}(\lambda) := e^{-\lambda\sum_{j=1}^{J}\psi_{j}(\Mj)}\tau_{0}(\lambda)$.
        \item[ii.] The jumps for observed species follow a proper probability density $f_H(\lambda)$, given by
        \[ f_{H}(\lambda) := \frac{\left(1-e^{-\lambda\sum_{j=1}^{J}\psi_{j}(\Mj)}\right)\tau_{0}(\lambda)}{\Psi_{0}\left(\sum_{j=1}^{J}\psi_{j}(\Mj)\right)}. \]
    \end{itemize}
    This results in a decomposition of the $B_{0}$ itself into an unobserved part and an observed part:
    \begin{equation}
    \label{postdisint}
        B_{0} \overset{d}{=} B_{0,J} + \sum_{l=1}^{\varphi} H_{l}\delta_{\tilde{Y}_{l}},
    \end{equation}  
    where:
    \begin{itemize}
        \item[a.] $B_{0,J}\sim \mathrm{CRM}(\tau_{0,J},F_{0})$ represents the process of unobserved species.
        \item[b.] $\varphi \sim \mathrm{Poisson}\left(\Psi_{0}\left(\sum_{j=1}^{J}\psi_{j}(\Mj)\right)\right)$ is the total number of distinct species observed across all $J$ groups.
        \item[c.] The $(\tilde{Y}_{l})$ are the iid locations/species tags ($F_0$) of these observed species, and the $(H_{l})$ are their iid mean abundance rates, drawn from $f_H$.
    \end{itemize}

    \item [2. The Observed Allocation Process.]The measures $\mathscr{A}_{J}$ are constructed solely from the $\varphi$ observed species (and their group counts):
    \begin{equation}
    \label{realizedAllocationprocess}
    \mathscr{A}_{J} \overset{d}{=} \left(\sum_{l=1}^{\varphi}X_{j,l}\delta_{\tilde{Y}_{l}}, j\in[J]\right).
    \end{equation}
    For each observed species $\tilde{Y}_{l}$, conditional on its abundance rate $H_l = \lambda$, the vector of counts $(X_{1,l}, \dots, X_{J,l})$ follows a multivariate zero-truncated Poisson distribution, which ensures that $\sum_{j=1}^J X_{j,l} \ge 1$:
    \[ (X_{1,l}, \dots, X_{J,l}) \mid H_l=\lambda \sim \mathrm{tP}(\lambda \psi_{1}(\sum_{i=1}^{M_{1}}\gamma_{i,1}),\ldots,\lambda\psi_{J}(\sum_{i=1}^{M_{J}}\gamma_{i,J})).\]

    \item[3. Conditional Structure of the Allocation.] The allocation of counts for any observed species $l$ has the following nested structure:
    \begin{itemize}
        \item[a. Total Count:] The total number of unique subspecies/markers for species $l$, $\tilde{X}_{l} := \sum_{j=1}^{J}X_{j,l}$, conditional on $H_l=\lambda$, follows a univariate zero-truncated Poisson distribution:
        \[ \tilde{X}_{l} \mid H_l=\lambda \sim \mathrm{tP}\left(\lambda \sum_{j=1}^{J}\psi_{j}(\Mj)\right). \]
        \item[b.] $(\tilde{X}_{l})\overset{iid}\sim\mathrm{MtP}(\tau_{0},\sum_{j=1}^{J}\psi_{j}(\Mj))$
        \item[c. Distribution Across Groups:] Given the total count $\tilde{X}_{l}=x_l$, this total is distributed among the $J$ groups according to a Multinomial distribution:
        \[ (X_{1,l}, \dots, X_{J,l}) \mid \tilde{X}_{l}=x_l \sim \mathrm{Multi}\left(x_l; q_{1},\ldots, q_{J}\right), \]
        where the probabilities $q_{j} = \frac{\psi_{j}(\Mj)}{\sum_{v=1}^{J}\psi_{v}(\sum_{i=1}^{M_{j}}\gamma_{i,v})}$ represent the relative sampling effort for each group.
    \end{itemize}
\end{enumerate}
\end{prop}
We next describe the posterior distribution of $B_{0}|\mathscr{A}_{J}$ which will play several important roles in the sequel. We see it's posterior has the same form as the univariate Poisson IBP in~\cite{James2017}[Section 4.2], as it depends only on 
$(\sum_{l=1}^{\infty}[\sum_{j=1}^{J}\xi_{j,l}]\delta_{Y_{l}})|B_{0}\sim \mathrm{PoiP}([\sum_{j=1}^{J}\psi_{j}(\Mj)]B_0).$
\begin{prop}\label{propN} The posterior distribution of $B_{0}|\mathscr{A}_{J}$ only depends on the $(\tilde{X}_{l}=x_{l},\tilde{Y}_{l},l\in[\varphi],\varphi=r)$ 
 and follows the decomposition of $B_{0}$ in \eqref{postdisint} with given quantities as indicated, the distribution of $B_{0,J}\sim \mathrm{CRM}(\tau_{0,J},F_{0})$ unchanged, and $H_{l}|\tilde{X}_{l}=x_{l}$ having density proportional to $\lambda^{x_{l}}e^{-\lambda\sum_{j=1}^{J}\psi_{j}(\Mj)}\tau_{0}(\lambda)$, conditionally independent for $l\in[r].$
\end{prop}

\subsubsection{Interpretations of the Allocation Process $\mathscr{A}_{J}$ in the Microbiome Setting}\label{Remark:splitdata}

Proposition~\ref{prop2} indicates that, in the $(M_{j}, j\in[J])$ samples, \(\varphi\) distinct species \((\tilde{Y}_{l}, l\in[\varphi])\) will be generated and distributed across the \(J\) groups. These species correspond to \(X_{j,l}\) unique subspecies, which we can map to markers \((\tilde{U}_{j,k,l}, k\in[X_{j,l}])\) associated with corresponding OTU clusters or counts. Additionally, \(\tilde{X}_{l}=\sum_{j=1}^{J}X_{j,l}\) measures the number of distinct subspecies (markers) linked to specific OTUs across the \(J\) communities contributing to the counts of $\tilde{Y}_{l}.$ As previously mentioned, \(H_{l}\) represents the mean abundance rates of the observed species \(\tilde{Y}_{l}\). Furthermore, from Proposition~\ref{propN}, we have \(B_{0,J} = \sum_{l=1}^{\infty} \lambda'_{l} \delta_{Y'_{l}}\), where \((\lambda'_{l}, Y'_{l})_{l\geq 1}\) represents the species \((Y'_{l})_{l \geq 1}\) that do not appear in the \((M_{j}, j \in [J])\) samples. Here, \((\lambda'_{l})_{l \geq 1}\) are the corresponding random rates determined by the L\'evy density \(\tau_{0,J}(\lambda) = e^{-\lambda \sum_{j=1}^{J} \psi_{j}(\Mj)} \tau_{0}(\lambda)\). For general but related interpretations of \((H_{l}, \tilde{X}_{l})_{l\ge 1}\), see~\cite{Pit97}[Section 3] and
~\cite{James2017}[Proposition 3.3]. Note that in many applications within our present setting, \(\tilde{X}_{l}\) is not observed directly.

\subsection{Exact sampling of the PHIBP marginal distribution}\label{exactsampling}
We now establish one of our main results which allows to sample exactly from the PHIBP marginal process. Note despite the relative simplicity of sampling, these lead to exact sampling of quite complex multi-dimensional distributions. See section~\ref{marginal} for precise descriptions of their apparently complex joint distributions. As noted, this may be used as a basis to describe a generative process for data arising in microbiome or other species sampling contexts. Important for our general exposition,  for $\tilde{C}_{j,k,l}:=\sum_{i=1}^{M_{j}}\tilde{C}^{(i)}_{j,k,l},$ the joint distribution of $S_{j,k,l}, \tilde{C}_{j,k,l}$ may be expressed as, for $c_{j,k,l}=1,2,\ldots,$
\begin{equation}
\label{MtPsimple}
\frac{s^{c_{j,k,l}}e^{-s\Mj}\tau_{j}(s)}{\psi^{(c_{j,k,l})}_{j}(\Mj)}\times 
\frac{{(\Mj)}^{c_{j,k,l}}\psi^{(c_{j,k,l})}_{j}(\Mj)}{\psi_{j}(\Mj)c_{j,k,l}!}
\end{equation}
where \(\psi^{(c_{j,k,l})}_{j}(\Mj)=\int_{0}^{\infty}s^{c_{j,k,l}}e^{-s\Mj}\tau_{j}(s)ds\) provides different interpretations of~\eqref{MtPsimple}, as detailed in~\cite{Pit97}. We denote \(\tilde{C}_{j,k,l} \sim \mathrm{MtP}(\tau_{j},\Mj)\), where \(\tilde{C}_{j,k,l} | S_{j,k,l} = s \sim \mathrm{tP}(s\Mj)\), which indicates a zero-truncated \(\mathrm{Poisson}(s\Mj)\) variable, and \(S_{j,k,l}\) has a marginal density \(f_{S_{j}}\) as in Proposition~\ref{IBPpost}. See~\cite{James2017}[Section 4.2].

In our species sampling context, $(S_{j,k,l},\tilde{C}_{j,k,l},\tilde{U}_{j,k,l})$ are respectively the mean abundance rate of OTU, total counts of OTU, and their corresponding markers $\tilde{U}_{j,k,l},$ associated with an observed species $\tilde{Y}_{l}.$ 
As we mentioned, $X_{j,l}$ counts the number of distinct markers for OTU clusters in community $j$. Additionally, also important for our general exposition, is the work of \cite{Pit97}[Proposition 12, Corollary 13], as noted in ~\cite{James2017}[Remark 4.1], also gives the variables $(H_{l},\tilde{X}_{l})$ another meaning in the Poisson setting, in terms of decomposing the space. From that work the joint distributions $(H_{l},\tilde{X}_{l})$ can be expressed in terms of $H_{l}|\tilde{X}_{l}=x_{l}$ and $\tilde{X}_{l},$ as,  
\begin{equation}
\label{HXdecomp}
\frac{\lambda^{x_{l}}e^{-\lambda\sum_{j=1}^{J}\psi_{j}(\Mj)}\tau_{0}(\lambda)}{\Psi^{(x_{l})}_{0}(\sum_{j=1}^{J}\psi_{j}(\Mj))}\times \frac{{(\sum_{j=1}^{J}\psi_{j}(\Mj))}^{x_{l}}\Psi^{(x_{l})}_{0}(\sum_{j=1}^{J}\psi_{j}(\Mj))}
{x_{l}!\Psi_{0}(\sum_{j=1}^{J}\psi_{j}(\Mj))}
\end{equation}
for $\Psi^{(x_{l})}_{0}(\sum_{j=1}^{J}\psi_{j}(\Mj))=\int_{0}^{\infty}
\lambda^{x_{l}}e^{-\lambda\sum_{j=1}^{J}\psi_{j}(\Mj)}\tau_{0}(\lambda)d\lambda.$

\begin{rem}
While these distributions may appear to be somewhat exotic, they appear in the works of \cite{James2017,Pit97,Titsias,FoFZhou} and elsewhere and are easily sampled. See, for instance~\cite{James2017}[Section 4.2.1, p. 2030] for  the gamma and generalized gamma cases, which we elaborate on further in our simulations. The gamma case for $\tilde{C}_{j,k,l}$ corresponds to logarithmic distributions in \cite{fisher,JMQ}.
\end{rem}

\begin{thm}\label{sumsampleprop}
Let for each $j$, $(Z^{(i)}_{j}| B_{j} \overset{ind}\sim \mathrm{PoiP}(\gamma_{i,j}B_{j})$, for $i\in[M_{j}],$ where $B_{j} | B_{0} \sim \mathrm{CRM}(\tau_{j}, B_{0})$ conditionally independent across $j\in[J]$. Then consider the sum process $(\sum_{i=1}^{M_{j}} Z^{(i)}_{j} , j\in[J])$, where $\sum_{i=1}^{M_{j}} Z^{(i)}_{j} | B_{j} \sim \mathrm{PoiP}(\Mj B_{j}),$ and $B_{0}\sim~
\mathrm{CRM}(\tau_{0},F_{0}).$
    \begin{enumerate}
    \item Then, for $\varphi \sim \mathrm{Poisson}(\Psi_{0}(\sum_{j=1}^{J}\psi_{j}(\Mj))),$ the joint marginal distribution of the sum process $(\sum_{i=1}^{M_{j}} Z^{(i)}_{j},j\in [J])$ is equal in distribution to
    \begin{equation}\label{prop:sumequation}
(\sum_{l=1}^{\varphi} N_{j,l}\delta_{\tilde{Y}_{l}}, j\in [J])\overset{d}=(\sum_{l=1}^{\varphi} \sum_{k=1}^{X_{j,l}}\tilde{C}_{j,k,l}\delta_{\tilde{Y}_{l}}, j\in [J])
    \end{equation}
\item where $(\tilde{C}_{j,k,l}=\sum_{i=1}^{M_{j}} \tilde{C}^{(i)}_{j,k,l}, k\in [X_{j,l}]) \overset{iid}\sim \mathrm{MtP}(\tau_{j},\Mj)$ with the $X_{j,l}$ components independent across $l$.
\item $\tilde{X}_{l}\sim \mathrm{MtP}(\tau_{0},\sum_{j=1}^{J}\psi_{j}(\Mj))$
\item $(X_{j,l},j\in[J])|\tilde{X}_{l}=x_{l}\sim \mathrm{Multi}(x_{l}; q_{1},\ldots, q_{J})$, for $q_{j}=\frac{\psi_{j}(\Mj)}{\sum_{v=1}^{J}\psi_{v}(\sum_{i=1}^{M_{v}}\gamma_{i,v})}.$   
\item  $(\sum_{k=1}^{X_{j,l}} \tilde{C}^{(i)}_{j,k,l}, i\in [M_{j}]) |N_{j,l}:=\sum_{k=1}^{X_{j,l}} \tilde{C}_{j,k,l} = n_{j,l}$ is $\mathrm{Multi}(n_{j,l};\frac{\gamma_{1,j}}{\Mj},\ldots,\frac{\gamma_{M_{j},j}}{\Mj})$, for each $j,l$.
    \end{enumerate}
\end{thm}

Theorem~\ref{sumsampleprop} indicates that one can sample from the marginal distribution of $(Z^{(i)}_{j},i\in[M_{j}], j\in[J]),$ by first sampling $((N_{j,l}, j\in [J]),\tilde{Y}_{l},l\in[\varphi],\varphi)$ from the sum process as in~\eqref{prop:sumequation} and then applying Multinomial sampling as in item 5. See \cite{hibp23}[Theorem 3.1 and Proposition 3.2] for indications on how to sample more general spike and slab HIBP and how to couple them to the PHIBP.

\section{Posterior distribution for PHIBP}\label{posteriorPoissonHIBP}
We now address the posterior distribution of the PHIBP, which exhibits remarkable properties
Here the given structure is  $\mathbf{Z}_{J}:=(\sum_{k=1}^{X_{j,l}}\tilde{C}^{(i)}_{j,k,l},\tilde{Y}_{l}, l\in[r],\varphi=r, i\in [M_{j}], j\in J).$ However in view of Theorem~\ref{sumsampleprop}, it suffices to work with the information in the sum process $(N_{j,l},\tilde{Y}_{l}, l\in[r],\varphi=r, j\in J),$ where again $N_{j,l}=\sum_{k=1}^{X_{j,l}}\tilde{C}_{j,k,l},$ for descriptions of the posterior distribution. The task then becomes to describe distributions of $(\tilde{C}_{j,k,l}, k\in [X_{j,l}], X_{j,l})$ given this information, which in general is a difficult problem, see \citep{hibp23}[Section 4 and Theorem 4.1]. 

One of our innovations is to note that conditional on \( (N_{j,l}, j \in [J]) \) and \( H_{l} = \lambda \), the variables are independent across the indices \( (j,l) \). This allows for the application of foundational results from \cite{Kolchin} and \cite{Pit97}, as stated in \cite{Pit06} (Theorems 1.2), to establish the following result, which describes the conditional distribution of the blocks of a finite Gibbs partition, governed by a finite Gibbs Exchangeable Partition Probability Function (EPPF), \( p^{[n_{j,l}]} \), arranged in exchangeable order:

\begin{prop}\label{prop:GibbsEPPF}
The distribution of the latent counts $(\mathbf{C}_{j,l}, X_{j,l})$ conditional on the total count \(N_{j,l} = n_{j,l}\) and the process parameter \(H_{l} = \lambda\) has the following properties:
\begin{enumerate}
    \item The distributions are independent across the indices $(j,l)$.

    \item For any group $j$ with $n_{j,l} > 0$, the joint probability mass function for the ordered composition of counts $\mathbf{c}_{j,l}:=(c_{j,1,l},\ldots,c_{j,x_{j,l},l})$ is, for $\sum_{k=1}^{x_{j,l}}c_{j,k,l}=n_{j,l}$:
    \begin{equation} \label{eq:ordered_counts_pmf}
    \mathbb{P}(\mathbf{C}_{j,l} = \mathbf{c}_{j,l} | N_{j,l}=n_{j,l}, H_l=\lambda) = \frac{n_{j,l}!\lambda^{x_{j,l}}}{x_{j,l}!\Xi^{[n_{j,l}]}_j(\lambda,\Mj)} \frac{\prod_{k=1}^{x_{j,l}}\psi^{(c_{j,k,l})}_{j}(\Mj)}{\prod_{k=1}^{x_{j,l}} c_{j,k,l}!}
    \end{equation}
    This distribution corresponds to arranging the blocks of a finite Gibbs partition in exchangeable order. 
    \item The conditional probability of the underlying unordered partition is given by the finite Gibbs EPPF, 
    $$p^{[n_{j,l}]}(\mathbf{c}_{j,l}|\lambda\tau_{j},\Mj) = \frac{\lambda^{x_{j,l}}}{\Xi^{[n_{j,l}]}_j(\lambda,\Mj)} \prod_{k=1}^{x_{j,l}}\psi^{(c_{j,k,l})}_{j}(\Mj)$$.

    \item The normalization constant $\Xi^{[n_{j,l}]}_j$ is the sum of the $x_{j,l}$-part partition functions:
    $$
    \Xi^{[n_{j,l}]}_j(\lambda,\Mj)=\sum_{x_{j,l}=1}^{n_{j,l}}\Xi^{[n_{j,l}]}_{j,x_{j,l}}(\lambda\tau_{j},\Mj).
    $$
    It also has the key representation:
    $$
    \mathbb{E}\left[ \sigma^{n_{j,l}}_{j,l}(\lambda)e^{-\sigma_{j,l}(\lambda)\Mj}\right]=e^{-\lambda\psi_{j}(\Mj)} \Xi^{[n_{j,l}]}_j(\lambda\tau_{j},\Mj).
    $$
    Each component $\Xi^{[n_{j,l}]}_{x_{j,l}}$ may be expressed as, 
    \begin{equation}
    \label{xisumrep}
    \Xi^{[n_{j,l}]}_{j,x_{j,l}}(\lambda\tau_{j},\Mj):=\frac{n_{j,l}!\lambda^{x_{j,l}}}{x_{j,l}!}\sum_{\mathbf{c}_{j,l}}\frac{\prod_{k=1}^{x_{j,l}}\psi^{(c_{j,k,l})}_{j}(\Mj)}{\prod_{k=1}^{x_{j,l}} c_{j,k,l}!}
    \end{equation}
    where the sum is over all compositions $\mathbf{c}_{j,l}$ such that $\sum_{k=1}^{x_{j,l}}c_{j,k,l}=n_{j,l}$.

    \item The marginal conditional distribution for the number of blocks, $X_{j,l}$, is given by:
    $$
    \mathbb{P}(X_{j,l}=x_{j,l} | N_{j,l}=n_{j,l}, H_l=\lambda) = \frac{\Xi^{[n_{j,l}]}_{x_{j,l}}(\lambda\tau_{j},\Mj)}{\Xi^{[n_{j,l}]}_j(\lambda,\Mj)},
    $$
    and $x_{j,l}=0,$ if $n_{j,l}=0.$
\end{enumerate}
These forms equate to expressions in \cite{JLP2}[Proposition 4] and \cite{FoFZhou}[Sections 2.3, 3], and the general relationships can be read from~\cite{Pit06}[Exercise 1.5.2, p. 33].
\end{prop}

\subsection{Special Cases and Connections to Stirling Numbers}
The general result in Proposition~\ref{prop:GibbsEPPF} simplifies in important special cases, revealing connections to known combinatorial quantities. Since \(\tilde{C}_{j,k,l} \sim \mathrm{MtP}(\lambda\tau_{j},\Mj)=\mathrm{MtP}(\tau_{j},\Mj)\), for any $\lambda,$  we have the general relation
$$
\Xi^{[n_{j,l}]}_{x_{j,l}}(\tau_{j},\Mj)=\frac{n_{j,l}!{(\psi_{j}(\Mj))}^{x_{j,l}}}{{(\Mj)}^{n_{j,l}}x_{j,l}!}
\mathbb{P}\left(\sum_{k=1}^{x_{j,l}}\tilde{C}_{j,k,l}=n_{j,l}\right).
$$
This and other relationships can be read from~\cite{Pit06}[Exercise 1.5.2, p. 33].

[Generalized Gamma Case:] For $\tau_{j}(s)=\frac{\alpha_{j}}{\Gamma(1-\alpha_{j})}s^{-\alpha_{j}-1}e^{-s\zeta_{j}},$ the distribution of the sum $\sum_{k=1}^{x_{j,l}}\tilde{C}_{j,k,l}$ is related to the stable case ($\zeta_j=0$). For the stable case, the sum's distribution is given by generalized Stirling numbers of the first kind, $S_{\alpha_{j}}(n,k)$:
\begin{equation}
\label{genStirling}
\mathbb{P}\left(\sum_{k=1}^{x_{j,l}}C_{j,k,l}=n_{j,l}\right)=\frac{\alpha^{x_{j,l}}_{j}x_{j,l}!}{n_{j,l}!}S_{\alpha_{j}}(n_{j,l},x_{j,l}).
\end{equation}
See also~\cite{JamesStick,Pit97,FoFZhou}.

[Gamma Case:] For $\tau_{j}(s)=\theta_{j}s^{-1}e^{-s\zeta_{j}}$ (Gamma process), the EPPF is that of a Dirichlet process, see \citep{JLP2}[Remark 2, p. 86]:
$$
p^{[n_{j,l}]}(\mathbf{c}_{j,l}|\lambda\tau_{j},\Mj)=p_{0,\theta_{j}\lambda}(\mathbf{c}_{j,l}):=\frac{\Gamma(\theta_{j}\lambda)\theta_{j}^{x_{j,l}} \lambda^{x_{j,l}}}{\Gamma(\theta_{j}\lambda+n_{j,l})}\prod_{k=1}^{x_{j,l}} \Gamma(c_{j,k,l}).
$$
The distribution of the number of blocks $\Pi_{n_{j,l}}$ involves the unsigned Stirling numbers of the first kind, $\stirlingfirst{n}{k}$:
$$
\mathbb{P}(\Pi_{n_{j,l}}=x_{j,l}|\lambda)=\frac{\Gamma(\theta_{j}\lambda)\theta_{j}^{x_{j,l}} \lambda^{x_{j,l}}}{\Gamma(\theta_{j}\lambda+n_{j,l})}\stirlingfirst{n_{j,l}}{x_{j,l}},
$$
which informs the computational strategy in~\cite{ZhouCarin2015,ZhouPadilla2016}. See~\citep{CraneEwens,Pit96,Pit02,PY97,TavareEwens} for more on random partitions and related results for the one and two parameter Poisson Dirichlet families, otherwise referred to as Pitman-Yor processes~\citep{IJ2001,IJ2003}. 

\subsection{Descriptions of the posterior distribution}
We now present an implementable description of the posterior distribution for PHIBP.

\begin{thm}\label{postPoissonHIBP}
A description of the posterior distributions of $(B_{j}\in [J],B_{0})|\mathbf{Z}_{J}$ is as follows,
\begin{enumerate}
\item The joint distribution of $(B_{j},j\in[J]),B_{0}|\mathbf{Z}_{J}$ is such that component-wise and jointly, is equivalent in distribution to,
  \begin{equation}
  \label{postsumrep}
  (B_{j,M_{j}}+\sum_{l=1}^{r}\left[\hat{\sigma}_{j,l}(H_{l})+\sum_{k=1}^{X_{j,l}}S_{j,k,l}\right]\delta_{\tilde{Y}_{l}}, j\in[J]),B_{0,J}+\sum_{l=1}^{r}H_{l}\delta_{\tilde{Y}_{l}} 
  \end{equation}
for $B_{0,J}\sim \mathrm{CRM}(\tau_{0,J},F_{0}),$ $B_{j,M_{j}}\sim\mathrm{CRM}(\tau^{(M_{j})}_{j},B_{0,J})$, and 
\item $(\hat{\sigma}_{j,l}(H_{l}),j\in[J])|H_{l}=\lambda$ are independent, and independent of $(S_{j,k,l}),$ with density $\eta^{[0]}_{j}(t_{j}|\lambda,\Mj)=e^{-t_{j}\Mj}\eta_{j}(t_{j}|\lambda)e^{\lambda\psi_{j}(\Mj)}.$  
\item $S_{j,k,l}|\tilde{C}_{j,k,l}=c_{j,k,l}$ has density $\frac{s^{c_{j,k,l}}e^{-s\Mj}\tau_{j}(s)}{\psi^{(c_{j,k,l})}(\Mj)}.$ 
\item $(\tilde{C}^{(i)}_{j,k,l}, i\in[M_{j}])|\tilde{C}_{j,k,l}=c_{j,k,l} \sim \mathrm{Multi}(c_{j,k,l}, \frac{\gamma_{1,j}}{\Mj}, \ldots, \frac{\gamma_{M_{j},j}}{\Mj}).$
\item  $H_{l}|\tilde{X}_{l}=x_{l}$ has density $\frac{\lambda^{x_{l}}e^{-\lambda\sum_{j=1}^{J}\psi_{j}(\Mj)}\tau_{0}(\lambda)}{\Psi^{(x_{l})}_{0}(\sum_{j=1}^{J}\psi_{j}(\Mj))}$, where $\tilde{X}_{l}=\sum_{j=1}^{J}X_{j,l}$
\item \label{latentlaws} $(X_{j,l},(\tilde{C}_{j,k,l},k\in [X_{j,l}], l\in[r], j\in J),$ given $(N_{j,l}=n_{j,l}, j\in[J]),l\in[r])$ are conditionally independent over $l\in[r],$ with distributions proportional to the joint distributions $\mathbb{P}(X_{j,l}=x_{j,l}; j\in[J])\times \prod_{j=1}^{J}
\prod_{k=1}^{x_{j,l}}\mathbb{P}(\tilde{C}_{j,k,l}=c_{j,k,l}),$ specified in~Theorem~\ref{sumsampleprop},
where $\sum_{k=1}^{x_{j,l}}c_{j,k,l}=n_{j,l},$ $x_{l}=\sum_{j=1}^{J}x_{j,l},$ and noting $X_{j,l}=0$ if $N_{j,l}=0.$
\end{enumerate}
\end{thm}
A description of the posterior distributions  of $\bar{B}_{0},$ $\bar{B}_{j}$ and $((F_{j,l}, j\in[J])_{l\ge 1}|\mathbf{Z}_{J}$ follows as an immediate consequence of Theorem~\ref{postPoissonHIBP} and can be read from 
the meaning of~\eqref{postsumrep} along with descriptions in Section~\ref{MicrobiomeInterpet:PostHIBP}. We omit further details for brevity.
The following proposition provides two descriptions of the posterior distribution of the mean abundance rates, denoted as $(\tilde{\sigma}_{j,l}(H_{l}), j\in[J])$, of $\tilde{Y}_{l}$ selected in the sample for each group $j\in[J]$. This result follows from the randomization of the posterior distribution of the total mass in \cite{JLP2}[Theorem 1], see also \cite{Pit02}.
\begin{prop}\label{poissonequivalence}
Consider the setting in Theorem~\ref{postPoissonHIBP}. Let \((\tilde{\sigma}_{j,l}(H_{l}), j \in [J])\) represent the mean abundance rates of the selected species \(\tilde{Y}_{l}\) for \(l \in [\varphi]\) in the sample \(\mathbf{Z}_{J}\). Given that \((N_{j,l} = n_{j,l}, j \in [J], H_{l} = \lambda)\) for each \(l \in [r]\), the vectors are conditionally independent across \(j\) and \(l\), with density \(\eta^{[n_{j,l}]}_{j}(t_{j} | \lambda, \Mj) = \frac{t_{j}^{n_{j,l}} \eta^{[0]}_{j}(t_{j} | \lambda, \Mj)}{\Xi^{[n_{j,l}]}(\lambda\tau_{j}, \Mj)}
\), leading to the distributional equivalence:
\begin{equation}
\label{splitsigma}
\tilde{\sigma}_{j,l}(\lambda) \overset{d}{=} \hat{\sigma}_{j,l}(\lambda) + \sum_{k=1}^{X_{j,l}} S_{j,k,l}.
\end{equation}
\end{prop}
See~\cite{JamesStick} for more on such variables, and also \cite{JLP2} for more examples.  As we shall see in Section~\ref{unseen}, the decomposition on the right hand side of~\eqref{splitsigma} is crucial for obtaining refined interpretations in the  microbiome setting.

\subsection{Some Interpretations with Respect to Microbiome Data}\label{MicrobiomeInterpet:PostHIBP}
We note that \(B_{j,M_{j}} \sim \mathrm{CRM}(\tau^{(M_{j})}_{j}, B_{0,J})\) indicates that 
\(
B_{j,M_{j}} \overset{d}{=} \sum_{l=1}^{\infty} \sigma'_{j,l}(\lambda'_{l}) \delta_{Y'_{l}},
\)
where \(\sigma'_{j,l}(\lambda'_{l})\) is interpreted as the mean abundance rate of the unseen species \(Y'_{l}\), with \((\lambda'_{l}, Y'_{l})\) defined as in Remark \ref{Remark:splitdata}. Furthermore, given \(\lambda'_{l} = \lambda\), the density of \(\sigma'_{j,l}(\lambda)\) is given by 
\(
e^{-[t_{j}\Mj - \psi_{j}(\Mj)]} \eta_{j}(t_{j} | \lambda)\). Moreover, \(
\tilde{\sigma}_{j,l}(H_{l}) \overset{d}{=} \hat{\sigma}_{j,l}(H_{l}) + \sum_{k=1}^{X_{j,l}} S_{j,k,l},
\)
in~\eqref{splitsigma} of Proposition~\ref{poissonequivalence}, and Theorem~\ref{postPoissonHIBP}, corresponds to the posterior distribution of the mean abundance rate of \(\tilde{Y}_{l}\) in the \(j\)-th group based on the sample \((M_{j}, j \in [J])\). This can be used to sample the posterior distribution of \((\mathscr{D}_{j}, j \in [J]) | (N_{j,l}, j \in [J], l \in [r])\), as otherwise defined in \eqref{Shannon1}, among many other possibilities. In the Gamma case, where \(\tau_{j}(s) = \theta_{j} s^{-1} e^{-s\zeta_{j}}\) and hence  \(
\tilde{\sigma}_{j,l}(H_{l}) \sim \mathrm{Gamma}(\theta_{j} H_{l} + n_{j,l}, \Mj + \zeta_{j}),
\) it follows that 

\[
\left(\frac{\tilde{\sigma}_{j,l}(H_{l})}{\sum_{t=1}^{r} \tilde{\sigma}_{j,t}(H_{t})}, l \in [r]\right) \sim \mathrm{Dirichlet}(\theta_{j} H_{l} + n_{j,l}; l \in [r]),
\]

leading to fairly simple sampling from \((\mathscr{D}_{j}, j \in [J]) | (N_{j,l}, j \in [J], l \in [r], \varphi = r)\). However, the corresponding generalized gamma case specified by $
\tau_j(s) = \frac{\theta_j}{\Gamma(1-\alpha_j)} s^{-\alpha_j-1} e^{-\zeta_j s}$, $j \in [J],$ yields
$(\Mj+\zeta_{j})\hat{\sigma}_{j,l}(\lambda)\overset{d}=T^{(j,l)}_{\alpha}(\frac{\theta_{j}}{\alpha_{j}}(\Mj+\zeta_{j})^{\alpha_{j}}\lambda),$ and leads to a more detailed structure, where it follows that for each $j\in[J]$, and $l\in[r],$
\begin{equation}
\frac{\tilde{\sigma}_{j,l}(H_{l})}{\sum_{t=1}^{r} \tilde{\sigma}_{j,t}(H_{t})} \overset{d}=
\frac{T^{(j,l)}_{\alpha_{j}}(\frac{\theta_{j}}{\alpha_{j}}{(\Mj+\zeta_{j})}^{\alpha_{j}}H_{l}) + G_{n_{j,l} - X_{j,l}\alpha_{j}}}
{T^{(j)}_{\alpha_{j}}(\frac{\theta_{j}}{\alpha_{j}}{\Mj+\zeta_{j})}^{\alpha_{j}}\sum_{t=1}^{r}H_{t}) + G_{n^{(+)}_{j} - X^{(+)}_{j}\alpha_{j}}},
\label{gengamgroupseenprobabilities}
\end{equation}

where \(T^{(j)}_{\alpha_{j}}(\frac{\theta_{j}}{\alpha_{j}}{(\Mj+\zeta_{j})}^{\alpha_{j}}\sum_{l=1}^{r}H_{l}) \overset{d}{=} \sum_{l=1}^{r} 
T^{(j,l)}_{\alpha_{j}}(\frac{\theta_{j}}{\alpha_{j}}{(\Mj+\zeta_{j})}^{\alpha_{j}}H_{l})\) and also \( G_{n^{(+)}_{j} - X^{(+)}_{j}\alpha_{j}}\overset{d}{=} \sum_{l=1}^{r} G_{n_{j,l} - X_{j,l}\alpha_{j}}\). \(n^{(+)}_{j}=\sum_{l=1}^{r}n_{j,l}\) denotes the total abundance of the observed species 
\((\tilde{Y}_{l}, l \in [r])\) in group \(j\), \(X^{(+)}_{j}=\sum_{l=1}^{r}X_{j,l}\) is the number of distinct OTUs in group $j,$ in the sample $(Z^{(i)}_{j}, i\in[M_{j}], j\in[J]).$ Furthermore, \((S_{j,k,l}, k \in [X_{j,l}])\) corresponds to the mean abundance rates of the markers \((\tilde{U}_{j,k,l}, k \in [X_{j,l}])\), each reflecting the abundance of the corresponding OTU with counts \((\tilde{C}_{j,k,l}, k \in [X_{j,l}])\) contributing to the counts \((N_{j,l})\) of \(\tilde{Y}_{l}\).  Meanwhile, 
$
\hat{\sigma}_{j,l}(H_{l}) = \sum_{k=1}^{\infty} s^{(H_{l},')}_{j,k,l},
$
denotes the mean abundance rate of \(\tilde{Y}_{l}\) in group \(j\) based on markers of OTUs, with respective mean abundance rates \((s^{(H_{l},')}_{j,k,l})_{k \geq 1}\) that have not yet contributed to the counts of \(\tilde{Y}_{l}\) in group \(j\). This means that these OTUs do not appear in the sample for group \(j\).
%\begin{prop}\label{PropgenPGG}
%Set $B_{0}\sim \mathrm{GG}(\alpha_{0},\theta_{0},\zeta_{0}, F_{0})$ denoting a %generalized gamma process $\mathrm{CRM}$ with L\'evy density for $0<\alpha_{0}%<1$:
%$
%\tau_{0}(\lambda)=\theta_{0}\alpha_{0}\lambda^{-\alpha_{0}-1}e^{-%\zeta_{0}\lambda}/\Gamma(1-\alpha_{0})$ and 
%$\Psi_{0}(t)=\theta_{0}[(\zeta_{0}+t)^{\alpha_{0}}-\zeta^{\alpha_{0}}_{0}].$ %Then given $\mathbf{Z}_{J}$: 
%1. $B_{0,J}\sim \mathrm{GG}(\alpha_{0},\theta_{0},\zeta_{0}+\sum_{j=1}^{J}\psi_{j}(M_{j}))$
%2. $H_{l}|\tilde{X}_{l}\sim \mathrm{Gamma}(\tilde{X}_{l}-%\alpha_{0},\zeta_{0}+\sum_{j=1}^{J}\psi_{j}(M_{j}))$ and 3.
%for $\tau_{j}(s) =\frac{ \alpha_{j}\theta_{j}e^{-s\zeta_{j}}s^{-\alpha_{j}-1}}%{\Gamma(1-\alpha_{j})}$, for $j\in[J],$ $((\tilde{C}_{j,k,l}=c_{j,k,l},k\in %[X_{j,l}],X_{j,l}=x_{j,l}) j\in J),$ given $(N_{j,l}, j\in[J])$
%has, for $x_{l}=\sum_{j=1}^{J}x_{j,l},$ with $x_{j,l}=0$ when $n_{j,l}=0,$ and %otherwise $x_{j,l}\in[n_{j,l}],$ distribution proportional to,
%\begin{equation}
%\frac{\Gamma(x_{l} - \alpha_{0})}%{\prod_{j=1}^{J}\Gamma(x_{j,l})}\frac{\prod_{j=1}^{J}\alpha_{j}\theta^{x_{j,l}}_%{j}{(\zeta_{j}+M_{j})}^{x_{j,l}\alpha_{j}}}{(\zeta_{0} + \sum_{j=1}^{J} %\theta_{j} [(\zeta_{j} + M_j)^{\alpha_{j}} - \zeta_{j}^{\alpha_{j}}])^{x_{l} - %\alpha_{0}}} \prod_{j=1}^{J}p_{\alpha_{j},0}(\mathbf{c}_{j,l})
%\label{specialstable}
%\end{equation}
%\end{prop}

\subsection{The marginal distribution of $(N_{j,l},j\in[J])$}\label{marginal}
We now provide descriptions of the general marginal process. First note that $(N_{j,l},j\in[J])|(\tilde{\sigma}_{j,l}(H_{l})=t_{j}, j\in[J]),H_{l}$ is $\mathrm{tP}(t_{1}\sum_{i=1}^{M_{1}}\gamma_{i,1},\ldots,t_{J}\sum_{i=1}^{M_{J}}\gamma_{i,J}),$ and hence there is a joint distribution of  $(N_{j,l},\tilde{\sigma}_{j,l}(H_{l}),j\in[J]),H_{l},$ with $n_{l}:=\sum_{j=1}^{J}n_{j,l}=1,2,\ldots,$ 
$$
\frac{\prod_{j=1}^{J} (\Mj)^{n_{j,l}} t_{j}^{n_{j,l}} e^{-t_{j} \Mj}}{(1 - e^{-\sum_{j=1}^{J} \Mj t_{j}}) \prod_{j=1}^{J} n_{j,l}!} \cdot \frac{(1 - e^{-t_{j} \sum_{j=1}^{J} \Mj}) \prod_{j=1}^{J} \eta_{j}(t_{j} | \lambda)}{(1 - e^{-\lambda \sum_{j=1}^{J} \psi_{j}(\Mj)})} f_{H_{l}}(\lambda),
$$
which can be deduced as a special case of results in~\citep[Section 5]{James2017}.
For the next result define vectors, $\mathbf{k_J} = (k_1, k_2, \ldots, k_J) \in \mathbb{N}^J$, and $\mathbf{n}_l = (n_{1,l}, \ldots, n_{J,l})$. Where, $k_j(n_{j,l}) = 0$ if $n_{j,l} = 0$ and $k_j(n_{j,l}) \in [n_{j,l}]$ otherwise. Let the collection of all such vectors $\mathbf{k_J}$ be denoted by $\mathcal{K}_J({\mathbf{n}_l})$. For the results that follow, recall the identity, whose sum form will facilitate practical implementation, 
\begin{align}
    \mathbb{E}\left[ (\sigma_{j}(\lambda))^{n_{j,l}} e^{-\sigma_{j}(\lambda)\psi_{j}(\Mj)} \right] 
    &= \int_0^{\infty} t^{n_{j,l}} e^{-t \Mj} \eta_j(t|\lambda) \,dt \notag \\
    &= e^{-\lambda\psi_{j}(\Mj)} \Xi^{[n_{j,l}]}(\lambda\tau_{j}, \Mj), \label{eq:xi_moment_identity_brief}
\end{align}
where $\Xi^{[0]}(\lambda\tau_{j}, \Mj)=\Xi^{[0]}_{0}(\lambda\tau_{j}, \Mj)=1,$ and  $\sum_{x_{j,l}=1}^{n_{j,l}} \Xi^{[n_{j,l}]}_{x_{j,l}}(\lambda\tau_{j}, \Mj)$ if $n_{j,l}>0$

\begin{prop}\label{post:marginalofN}
There are the following descriptions of the marginal and conditional distributions of $(N_{j,l}=n_{j,l}, j\in[J])$ defined for $n_{l}=\sum_{j=1}^{J}n_{j,l}=1,2,\ldots,,$ and each $l\in[\varphi]$,for $\varphi\sim \mathrm{Poisson}(\Psi_{0}(\sum_{j=1}^{J}\psi_{j}(\Mj))).$
\begin{enumerate} 
    \item The conditional distribution of $(N_{j,l}=n_{j,l}, j\in[J])| H_{l}=\lambda$ can be expressed as,
    $$
   \frac{\prod_{j=1}^{J}{(\Mj)}^{n_{j,l}}}{\prod_{j=1}^{J}n_{j,l}!} \frac{e^{-\lambda\sum_{j=1}^{J}\psi_{j}(\Mj)}\prod_{j=1}^{J}\Xi^{[n_{j,l}]}(\lambda\tau_{j},\Mj)}{(1-e^{-\lambda\sum_{j=1}^{J}\psi_{j}(\Mj)})}
    $$
    \item The joint distribution may be expressed as 
    for a multi-index $\vec{n}_{l} = (n_{1,1}, \dots, n_{J,l}),$ such that $\sum_{j=1}^{J}n_{j,l}=1,2,\ldots$ and a parameter vector $\vec{\gamma} = (\sum_{i=1}^{M_{1}}\gamma_{i,1}, \dots, \sum_{i=1}^{M_{J}}\gamma_{i,J})$, the formula is:
\begin{equation}
\label{Ngroupldistforfrag}
\mathbb{P}(N_{j,l}=n_{j,l};j\in[J]) = \frac{\prod_{j=1}^{J}{(\Mj)}^{n_{j,l}}}{\prod_{j=1}^{J}n_{j,l}!}
\frac{{(\Psi_{0}\circ \sum_{j=1}^{J}\psi_{j})}^{(\vec{n}_{l})}(\vec{\gamma})}{\Psi_{0}(\sum_{j=1}^{J}\psi_{j}(\Mj))}.
\end{equation}
where,
$
\left(\Psi_{0}\circ \sum_{j=1}^{J}\psi_{j}\right)^{(\vec{n}_{l})}(\vec{\gamma})
:= \int_{0}^{\infty} \mathbb{E}\left[ \left(\prod_{j=1}^{J}[\sigma_{j}(\lambda)]^{n_{j,l}}\right)e^{-\sum_{j=1}^{J}\sigma_{j}(\lambda)\Mj} \right] \tau_{0}(\lambda)\,d\lambda.
$
\item The joint distribution of  $(N_{j,l}=n_{j,l}, j\in[J])$ has also the following sum representation, with $x_{j,l}=0$  when $n_{j,l}=0,$ and $x_{l}:=\sum_{j=1}^{J} x_{j,l},$:

 $$
\frac{\prod_{j=1}^{J}{(\Mj)}^{n_{j,l}}}{\prod_{j=1}^{J}n_{j,l}!}\frac{\sum_{\mathbf{x}_{j,l} \in \mathcal{K}_J(\mathbf{n}_l)}\Psi^{(x_{l})}_{0}(\sum_{j=1}^{J} \psi_j(\Mj)) \prod_{j=1}^{J}\Xi^{[n_{j,l}]}_{x_{j,l}}(\tau_{j}, \Mj) }
{\Psi_0\left(\sum_{j=1}^{J} \psi_j(\Mj)\right)}
$$

which is the same as $\sum_{\mathbf{x}_{j,l} \in \mathcal{K}_J(\mathbf{n}_l)}\mathbb{P}(X_{j,l}=x_{j,l}; j\in[J])\times \prod_{j=1}^{J}\mathbb{P}(\sum_{k=1}^{x_{j,l}}\tilde{C}_{j,k,l}=n_{j,l})$
\end{enumerate}
\end{prop}
\begin{rem}
Our work provides a multivariate extension of the univariate distribution described in Proposition 12 of~\cite{Pit97}. The representation for $J=1$ corresponds to the case of composition of two subordinators, and is of independent interest. In general, the expression in~\eqref{Ngroupldistforfrag} applies to a special class of  multivariate L\'evy processes tethered by a single subordinator~\citep{barndorff2001multivariate,James2017}, yielding explicit formulae for the component totals $(N_{j,l},j\in[J])$. Furthermore, the sum representation in 3 of Proposition~\ref{post:marginalofN} is computationally tractable and reveals a connection to permanental processes~\cite{CraneEwens,McCullaghMoller2006}.
\end{rem}

\subsection{Prediction rules}\label{predict}
We present the prediction rule for a new vector of observations $(Z^{(M_{j}+1)}_{j}, j\in[J])$, representing one new customer/sample per group, conditional on the existing data $\mathbf{Z}_{J}$. Further descriptions follow from Theorems~\ref{postPoissonHIBP} and Proposition~\ref{sumsampleprop}.

\begin{prop}\label{Prediction}
Let $Z^{(M_{j}+1)}_{j}|B_{j}\sim \mathrm{PoiP}(\gamma_{M_{j}+1,j}B_{j})$ for each $j\in [J]$, where $B_{j}\sim \mathrm{CRM}(\tau_{j},B_{0})$ and $B_{0}\sim \mathrm{CRM}(\tau_{0},F_{0})$. The predictive distribution of the new observation for group $j$, denoted $Z_j^{(new)}$, is a sum of three components:
\[
Z_j^{(new)} \overset{d}{=} \mathcal{N}_j^{(1)} + \mathcal{N}_j^{(2)} + \mathcal{N}_j^{(3)}
\]
These components represent completely new species, new group-level observations of existing species, and additional counts for previously observed species, respectively.

\begin{enumerate}
    \item [1. Completely New Species ($\mathcal{N}_j^{(1)}$):] A process involving new species, identified by $(Y^{*}_{v}) \overset{iid}{\sim} F_{0}$, which have not been previously observed in any group. For each group $j$, this component is:
    \[
    \mathcal{N}_j^{(1)} = \sum_{v=1}^{\varphi^{*}}N^{*}_{j,v}\delta_{Y^{*}_{v}}=\sum_{v=1}^{\varphi^{*}}\sum_{k=1}^{X^{*}_{j,v}}\tilde{C}^{(*,1)}_{j,k,v}\delta_{Y^{*}_{v}}
    \]
    where:
    \begin{itemize}
        \item[a.] The number of new species clusters, $\varphi^{*}$, is Poisson distributed:
        \[
        \varphi^{*}\sim \mathrm{Poisson}\left(\Psi_{0}\left(\sum_{j=1}^{J}\psi_{j}(\sum_{i=1}^{M_{j}+1}\gamma_{i,j})\right)-\Psi_{0}\left(\sum_{j=1}^{J}\psi_{j}(\Mj)\right)\right).
        \]
        \item[b.] For each new species $Y^{*}_{v} \sim F_0$, the number of features per group, $(X^{*}_{1,v}, \ldots, X^{*}_{J,v})$, is allocated via a Multinomial distribution conditional on their total $X^{*}_{v}$:
        \[
        (X^{*}_{1,v}, \ldots, X^{*}_{J,v}) | X^{*}_{v} \sim \mathrm{Multi}(X^{*}_{v}; q_{1}, \ldots, q_{J})
        \]
        with probabilities $q_j$ representing the relative rate of discovery in group $j$:
        \[
        q_{j}=\frac{\psi_{j}(\sum_{i=1}^{M_{j}+1}\gamma_{i,j})-\psi_{j}(\Mj)}{\sum_{t=1}^{J}[\psi_{t}(\sum_{i=1}^{M_{t}+1}\gamma_{i,t})-\psi_{t}(\sum_{i=1}^{M_{t}}\gamma_{i,t})]}.
        \]
        \item[c.]$X^{*}_{v}:=\sum_{j=1}^{J}X^{*}_{j,v}\sim \mathrm{MtP}(\tau_{0,J},\sum_{j=1}^{J}
[\psi_{j}(\sum_{i=1}^{M_{j}+1}\gamma_{i,j})-\psi_{j}(\Mj)]),$  such that the iid pairs say
\(((H^{*}_{v},X^{*}_{v}), v \in [\varphi^{*}])\) have common joint distribution
$$
\frac{{\lambda^{x_{v}}e^{-\lambda\sum_{j=1}^{J}\psi_{j}(\sum_{i=1}^{M_{j}+1}\gamma_{i,j})}\tau_{0}(\lambda)(\sum_{j=1}^{J}[\psi_{j}(\sum_{i=1}^{M_{j}+1}\gamma_{i,j})-
\psi_{j}(\Mj)])}^{x_{v}}}
{x_{v}![\Psi_{0}(\sum_{j=1}^{J}\psi_{j}(\sum_{i=1}^{M_{j}+1}\gamma_{i,j}))-\Psi_{0}(\sum_{j=1}^{J}\psi_{j}(\Mj))]}.
$$
        \item[d.] The counts $(\tilde{C}^{(*,1)}_{j,k,v})\overset{iid}\sim\mathrm{MtP}( \tau^{(M_{j})}_{j},\gamma_{M_{j}+1,j})$ are iid draws representing new observations. Hence, there are iid pairs \( (\tilde{S}^{(*,1)}_{j,k,v}, \tilde{C}^{(*,1)}_{j,k,v}) \) with a common joint distribution given by, for $s\in(0,\infty)$ and $c=1,2,\ldots$ :
\[
\frac{s^{c}e^{-s[\sum_{i=1}^{M_{j}+1}\gamma_{i,j}]} \tau_{j}(s)}{\psi^{(c_{j})}_{j}(\sum_{i=1}^{M_{j}+1}\gamma_{i,j})} \times \frac{
{(\gamma_{M_{j}+1,j})}^{c}
\psi^{(c)}_{j}(\sum_{i=1}^{M_{j}+1}\gamma_{i,j})}{[\psi_{j}(\sum_{i=1}^{M_{j}+1}\gamma_{i,j}) - \psi_{j}(\Mj)] c!}.\]
    \end{itemize}
    \item [2. New Group-Level Occurrences of Existing Species ($\mathcal{N}_j^{(2)}$):] A process for new distinct OTUs (subspecies) that are new to a group but belong to one of the $r$ established global clusters, corresponding to existing species $(\tilde{Y}_l)$. For each group $j$, this component is:
    \[
    \mathcal{N}_j^{(2)} = \sum_{l=1}^{r}\sum_{k=1}^{\mathscr{P}^{*}_{j,l}(H_{l})}\tilde{C}^{(*,2)}_{j,k,l}\delta_{\tilde{Y}_{l}}
    \]
    where:
    \begin{itemize}
        \item[a.] The posterior intensity $H_l$ for each species cluster $l\in[r]$ is from Theorem~\ref{postPoissonHIBP}.
        \item[b.] For each existing species $\tilde{Y}_l$, the number of new features, $(\mathscr{P}^{*}_{1,l}(H_{l}), \ldots, \mathscr{P}^{*}_{J,l}(H_{l}))$, is allocated via a Multinomial distribution, conditional on their Poisson-distributed sum, with mean  $H_{l}\sum_{t=1}^{J}[\psi_{t}(\sum_{i=1}^{M_{t}+1}\gamma_{i,t})-\psi_{t}(\sum_{i=1}^{M_{t}}\gamma_{i,t})],$ using the same probabilities $(q_1, \ldots, q_J)$ as in component (1).
        \item[c.] The counts $(\tilde{C}^{(*,2)}_{j,k,l})$ follow the same distribution as those in component (1).
    \end{itemize}
\item [3. Additional Counts for Existing Species ($\mathcal{N}_j^{(3)}$):] Additional counts for species $(\tilde{Y}_{l})$ that have already been observed within group $j$. For each group $j$, this component is:
    $$
    \mathcal{N}_j^{(3)} = \sum_{l=1}^{r} \sum_{k=1}^{X_{j,l}} C^{(new)}_{j,k,l} \delta_{\tilde{Y}_{l}}
    $$
    where the new counts $C^{(new)}_{j,k,l}$ are drawn from a Poisson distribution with a rate determined by the new customer's sampling effort and the species' posterior intensity:
    $$
    C^{(new)}_{j,k,l} \sim \mathrm{Poisson}(\gamma_{M_{j}+1,j}S_{j,k,l}).
    $$
    The posterior intensities $(S_{j,k,l})$ and feature counts $(X_{j,l})$ for $l\in[r], k\in[X_{j,l}]$ are given in Theorem~\ref{postPoissonHIBP}.
\end{enumerate}
\end{prop}

If one is interested in adding a new group, $J+1,$ as in~\cite{ZhouPadilla2016}, the prediction rule simplifies as follows.

\begin{cor}
Suppose that $Z^{(1)}_{J+1}|B_{J+1}\sim \mathrm{PoiP}(\gamma_{1,J+1}B_{J+1})$ and $B_{J+1}\sim \mathrm{CRM}(\tau_{J+1},B_{0}).$ Then the predictive distribution for the first observation in this new group, $Z_{J+1}^{(new)}|\mathbf{Z}_{J}$, consists of only two components, as there are no species previously observed in this group:
\[
Z_{J+1}^{(new)} \overset{d}{=} \mathcal{N}_{J+1}^{(1)} + \mathcal{N}_{J+1}^{(2)}
\]
where:
\begin{enumerate}
    \item[1. Completely New Species ($\mathcal{N}_{J+1}^{(1)}$):] Observations of species not seen in any of the previous $J$ groups. The number of such species, $\varphi^{*}_{J+1}$, is Poisson distributed:
    \[
    \varphi^{*}_{J+1}\sim \mathrm{Poisson}\left(\Psi_{0}\left(\psi_{J+1}(\gamma_{1,J+1})+\sum_{j=1}^{J}\psi_{j}(\Mj)\right)-\Psi_{0}\left(\sum_{j=1}^{J}\psi_{j}(\Mj)\right)\right).
    \]
with $X^{*}_{J+1,v}\sim \mathrm{MtP}(\psi_{J+1}(\tau_{0,J},\gamma_{1,J+1})),$ $\tilde{C}^{(*,1)}_{J+1,k,v}\sim \mathrm{MtP}(\tau_{J+1},\gamma_{1,J+1}),$ $Y^{*}_{v}\sim F_{0}$
    \item[2. New Group-Level Occurrences ($\mathcal{N}_{J+1}^{(2)}$):] Observations of species that exist in the global pool (groups $1, \ldots, J$) but are recorded for the first time in group $J+1$. For each existing species $\tilde{Y}_l$, the number of new features is Poisson:
    \[
    \mathscr{P}^{*}_{J+1,l}(H_{l})\sim \mathrm{Poisson}(H_{l}\psi_{J+1}(\gamma_{1,J+1})).
    \]
\end{enumerate}
The corresponding counts within these components  $\tilde{C}^{(*,2)}_{J+1,k,l}\sim \mathrm{MtP}(\tau_{J+1},\gamma_{1,J+1}).$ To obtain the simpler rule corresponding to the setting of~\cite{ZhouPadilla2016}, where $\mathbf{Z}_{J}:=(Z^{(1)}_{j},j\in[J]),$ set $M_{j}=1$ for all $j\in[J].$
\end{cor}

\subsection{Using the Prediction Rule for Unseen Species in Microbiome Studies}\label{unseen}
The problem of estimating the number of unseen species has a long history, originating with~\cite{fisher}. In the context of Bayesian nonparametrics, this problem is typically formulated for a single population under the assumption that a sample of a fixed, deterministic size is observed~\cite{favaro2009bayesian,lijoi2007bayesian,balocchi2024bayesian}. While this assumption is well-suited for discrete sampling experiments, it does not directly translate to the analysis of modern microbiome samples. In sequencing a microbiome sample, there is no natural, pre-existing count of "individuals"; the total number of sequencing reads is an outcome of the experimental process, not a fixed property of the underlying biological community. Our hierarchical framework is designed for this reality. It extends the classical setup in two fundamental ways. First, we address the challenge of modeling multiple populations simultaneously. Second, and crucially, we treat the total number of observations as a random variable. This removes the artificial constraint of a fixed sample size, providing a more physically grounded model for data from sequencing-based studies and addressing known challenges in the field~\cite{jeganathan2021statistical}.

The predictive distribution from Proposition~\ref{Prediction} provides a rich, model-based engine for exploring the "unseen." The three components directly map to different aspects of discovery in a new sample:
\begin{itemize}
    \item[$\mathcal{N}_j^{(1)}$: Arrival of Completely New Species.] This term governs the arrival of $\varphi^{*}$ new species, labeled $(Y_v^*, v \in [\varphi^*])$, which have never been observed in any group. When such a species $Y_v^*$ appears in group $j$, it is composed of $X_{j,v}^*$ new, distinct OTUs (subspecies), whose initial counts are given by $(\tilde{C}^{(*,1)}_{j,k,v})$ and whose mean rates are $(\tilde{S}^{(*,1)}_{j,k,v})$.

    \item[$\mathcal{N}_j^{(2)}$: Discovery of New OTUs for Existing Species.] This term accounts for the discovery of new subspecies for a previously documented species $\tilde{Y}_l$. Specifically, it generates $\mathscr{P}^{*}_{j,l}(H_{l})$ new, distinct OTUs for species $\tilde{Y}_l$ within group $j$. These new OTUs contribute counts of $(\tilde{C}^{(*,2)}_{j,k,l})$ and have mean rates of $(\tilde{S}^{(*,2)}_{j,k,l})$.

    \item[$\mathcal{N}_j^{(3)}$: Additional Counts for Existing OTUs.] This final term describes an increase in observations for OTUs that are already known to exist within group $j$. For a species $\tilde{Y}_l$, the count $\tilde{C}_{j,k,l}$ for each of its existing OTUs, $k \in [X_{j,l}]$, is incremented by an amount determined by the process $  C^{(new)}_{j,k,l} \overset{d}=\mathscr{P}^{(M_{j}+1)}_{j,k,l}(\gamma_{M_{j}+1,1}S_{j,k,l})$.
\end{itemize}

This structure allows us to form posterior predictive distributions to answer classical ecological questions, such as those discussed in~\cite{balocchi2024bayesian}:
\begin{itemize}
    \item[\textbf{Q1:}] What is the expected population frequency of a species with frequency $t \ge 1$ in the sample?
    \item[\textbf{Q2:}] How many previously unobserved species will be discovered in an additional sample?
    \item[\textbf{Q3:}] How many species with frequency $t \ge 1$ in the sample will be re-observed?
\end{itemize}

To demonstrate, we propose a novel approach to Q2 by defining a parameter for the diversity of the truly unseen species. We introduce $\mathscr{U}_{j,M_{j}+1}$, the predictive Shannon entropy of the completely new species that will appear in the next sample for group $j$. It is defined on the mean abundance rates $(\tilde{\sigma}^{*}_{j,v})$ of the $\varphi^*$ new species from component $\mathcal{N}_j^{(1)}$:
\begin{equation}
\mathscr{U}_{j,M_{j}+1} := -\sum_{v=1}^{\varphi^*} \frac{\tilde{\sigma}^{*}_{j,v}(H^{*}_{v})}{\sum_{t=1}^{\varphi^*} \tilde{\sigma}^{*}_{j,t}(H^{*}_{t})} \ln{\left(\frac{\tilde{\sigma}^{*}_{j,v}(H^{*}_{v})}{\sum_{t=1}^{\varphi^{*}} \tilde{\sigma}^{*}_{j,t}(H^{*}_{t})}\right)}, \quad j \in [J]
\label{ShannonUnseen}
\end{equation}
The distribution of this random measure of unseen diversity is determined by the posterior of the model. For the flexible generalized gamma case, we have the following result.

\begin{prop}
Consider the unseen entropy measure $\mathscr{U}_{j,M_{j}+1}$ in~\eqref{ShannonUnseen}. Let the Lévy intensities be of the generalized gamma (GG) type: $\tau_j(s) = \frac{\theta_j}{\Gamma(1-\alpha_j)} s^{-\alpha_j-1} e^{-\zeta_j s}$ for $j \in [J]$, and $\tau_0(\lambda) = \frac{\theta_0}{\Gamma(1-\alpha_0)} \lambda^{-\alpha_0-1} e^{-\zeta_0 \lambda}$. Let $\tilde{\zeta}_j := \zeta_j + \sum_{i=1}^{M_j+1} \gamma_{i,j}$ be the updated rate for group $j$, and let the updated global rate be $\tilde{\zeta}_0 := \zeta_{0} + \sum_{t=1}^{J} \frac{\theta_{t}}{\alpha_{t}} ( \tilde{\zeta}_t^{\alpha_t} - \zeta_t^{\alpha_t} )$. Then, conditional on the data $\mathbf{Z}_J$, the scaled abundance rate for a new species $Y_v^*$ in group $j$ is
\[
\tilde{\zeta}_{j}\times \tilde{\sigma}^{*}_{j,v}(H^{*}_{v}) \overset{d}{=} T^{(j,v)}_{\alpha_{j}}(\rho_{j}G_{X^{*}_{v}-\alpha_{0}}) + G_{N^{*}_{j,v} - X^{*}_{j,v}\alpha_{j}},
\]
where $T^{(j,v)}_{\alpha}(y)$ is a GG variable, $G_a \sim \mathrm{Gamma}(a,1)$, and the global intensity $H_v^*$ and scaling proportion $\rho_j$ are given by
\begin{align*}
\tilde{\zeta}_0 H^{*}_{v} \overset{d}{=} G_{X^{*}_{v}-\alpha_{0}} \quad \text{and} \quad
\rho_{j} = \frac{1}{\tilde{\zeta}_0} \left( \frac{\theta_{j}}{\alpha_{j}} \tilde{\zeta}_j^{\alpha_j} \right).
\end{align*}
These results provide a direct path to sample the predictive distribution of $\mathscr{U}_{j,M_{j}+1}$ and related measures.
\end{prop}

\section{Extensions}
\label{sec:Extensions}
As we have shown the PHIBP framework can be used to model many key aspects of microbiome or more general phenomena. We now provide indications for further extensions where experts can incorporate domain specific knowledge via covariates and latent variable modelling. In particular we describe three general types of extensions. Note we may repeat notation which have different meaning in each context. [1. Technical variation adjusments] to model within group heterogeneity which adjusts, for example, for sequencing depth, region size etc. can be instituted by specifying $\gamma_{i,j} = \exp\left(\mathbf{V}_i^T \boldsymbol{\beta}_j\right)$ where 
$\mathbf{V}_i^T$ represent covariates and $\boldsymbol{\beta}_j$ are parameters. [2. Group level taxon-specific covariate effects] can be incorporated using GLMs as follows: given $\tilde{Y}_{l}, N_{j,l}, l=1,\ldots,\varphi,$ For taxon $Y_l$ in sample $i$ from group $j$:
$
W_{i,j,l} \mid Y_l \sim F\left(g^{-1}(\eta_{i,j,l})\right), \quad \eta_{i,j,l} = \alpha_l + \mathbf{U}_j^T \boldsymbol{\beta}_l + \mathbf{R}_i^T \boldsymbol{\nu}_l,
$
where: $F$: Exponential family distribution (e.g., Negative Binomial for counts), $g$: Link function (e.g., log for count data, logit for binary traits), $\mathbf{U}_j$: Group-level covariates (e.g., treatment), $\mathbf{R}_i$: Sample-level covariates (e.g., pH which measures how acidic or alkaline an environment is). [3. Strain-Trait Modeling] The PHIBP framework leverages Theorem 3.1's latent components $(S_{j,k,l}, \tilde{C}_{j,k,l})$ for OTUs with $\tilde{C}_{j,k,l} > 0$. When strain counts $\tilde{C}_{j,k,l}$ are directly observed (e.g., via \textsc{DADA2} ASVs), these serve as fixed inputs for inferring fitness rates $S_{j,k,l}$. Otherwise, $\tilde{C}_{j,k,l}, X_{j,l}$ are sampled given $N_{j,l}$. The fitness rates $S_{j,k,l}$ quantify per-OTU ecological contributions, enabling direct modeling of strain-trait relationships:
\[
W_{j,k,l} \mid S_{j,k,l}, \mathbf{U}_j \sim \mathrm{F}_{\mathrm{trait}}\bigl(g(S_{j,k,l}, \mathbf{U}_j)\bigr),
\]
where: $\mathrm{F}_{\mathrm{trait}}$ is the trait distribution (e.g., $\mathrm{Bernoulli}$ for binary virulence, $\mathrm{Poisson}$ for colonization counts). $g$ is a link function, with examples: Linear: $g(S_{j,k,l}, \mathbf{U}_j) = \beta_0 + \beta_1 S_{j,k,l} + \boldsymbol{\beta}_2^\top \mathbf{U}_j,$
Logistic: $g(S_{j,k,l}) = \mathrm{logit}^{-1}(\beta_0 + \beta_1 S_{j,k,l}).$ \textsc{DADA2} resolves exact sequence variants but lacks ecological parameters ($S_{j,k,l}$). Whereas PHIBP, provides complementary hierarchical modeling of strain fitness whether $\tilde{C}_{j,k,l}$ are observed (ASVs) or latent (aggregate counts). See also~\citep{Scheinthesis} for some general ideas of possible extensions.

\section{Experiments}
In this section, we present an empirical study of the GG PHIBP model using both simulated and real microbiome population data. Specifically, we choose
\[
\tau_0(\lambda) = \frac{\theta_0}{\Gamma(1-\alpha_0)}\lambda^{-\alpha_0-1}e^{-\zeta_0\lambda}, \quad
\tau_j(s) = \frac{\theta_j}{\Gamma(1-\alpha_j)}s^{-\alpha_j-1}e^{-\zeta_j s} \text{ for } j \in [J],
\]
where $\alpha_0, (\alpha_j)_{j\in [J]} \in [0, 1)$, $\theta_0, (\theta_j)_{j\in[J]} > 0$. We fix $\zeta_0 = \{\zeta_j\}_{j\in [J]} = 1,$ and $\gamma_{i,j}=1,$ for simplicity, but inferring them does not add extra components to the algorithm. When $\alpha_0$ and $\alpha_j$ are fixed to 0, the model reduces to a Gamma PHIBP.
Due to space constraints, we defer the details of the MCMC algorithm for posterior inference and the procedure for computing predictive likelihoods based on Proposition 4.4 to Section B in the supplementary materials.

\subsection{Simulated Data}
\begin{figure}
    \centering
    \includegraphics[width=0.28\linewidth]{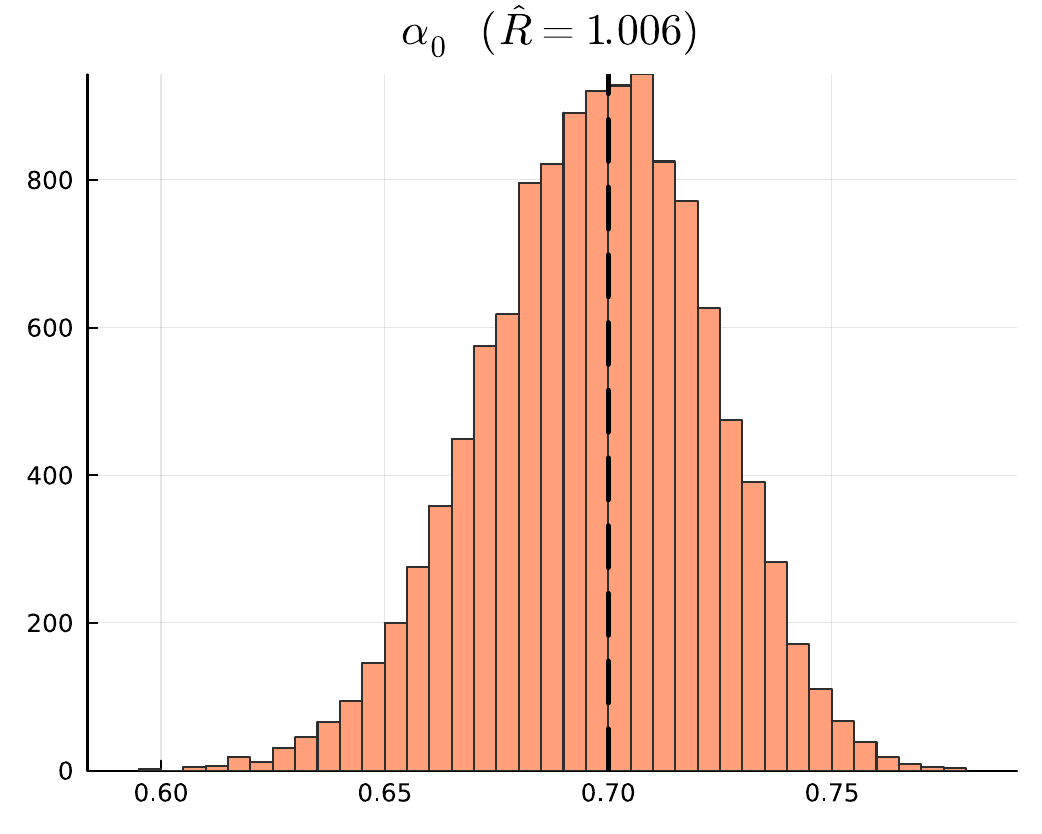}
    \includegraphics[width=0.28\linewidth]{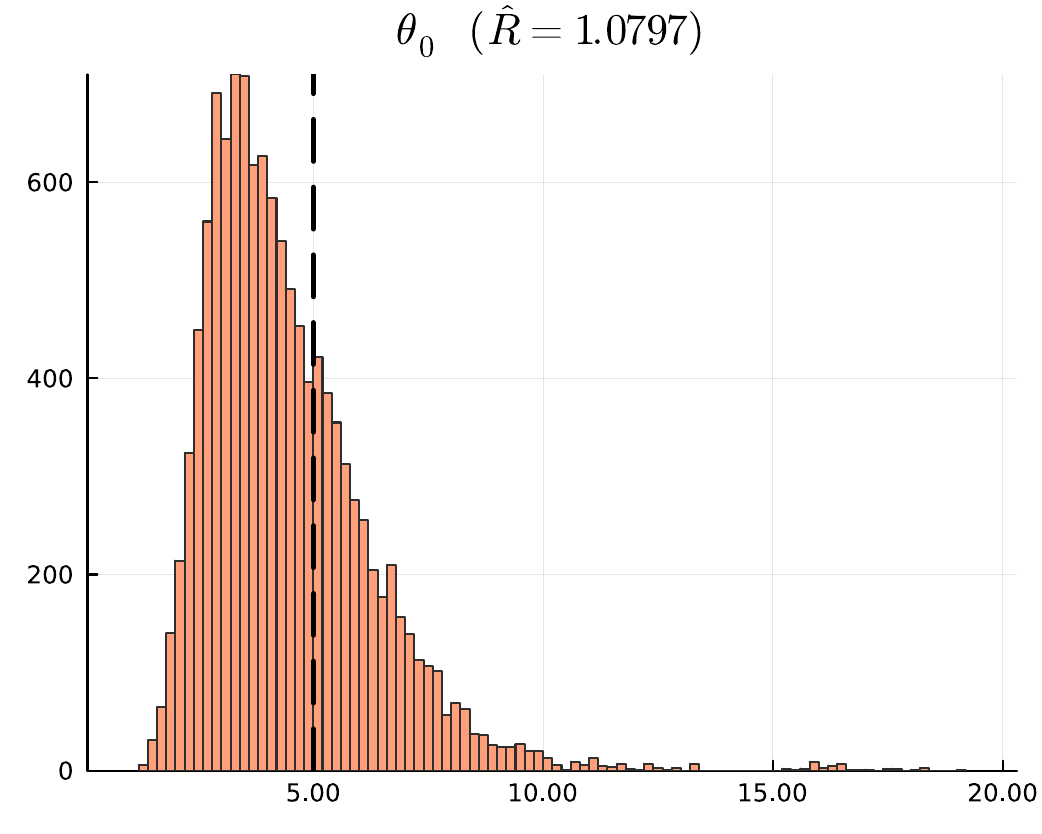} 
    \includegraphics[width=0.36\linewidth]{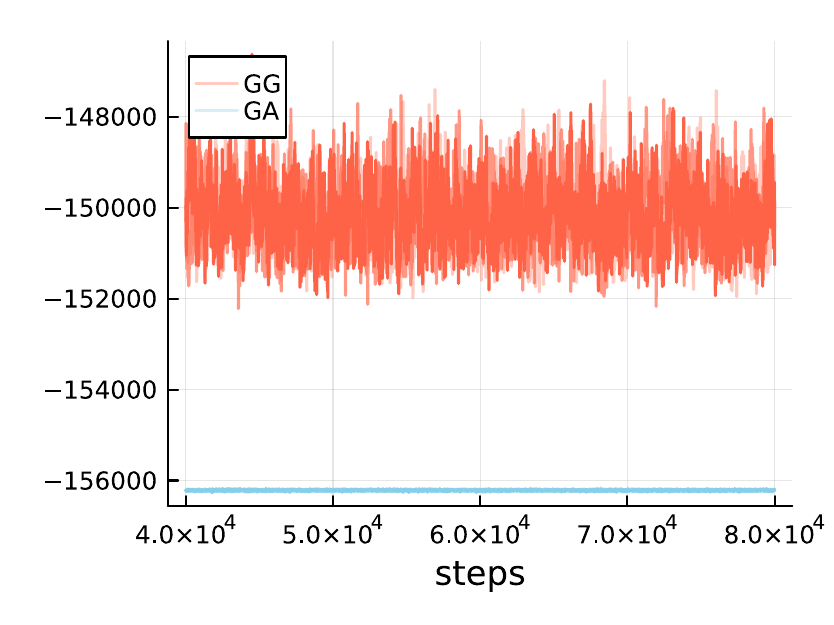}
    \caption{Simulated data experiments.  Left and middle: Posterior samples for $\alpha_0$ and $\theta_0$, with black dotted lines indicating the true values. $\hat{R}$ statistics~\citep{GelmanInference,BrooksGeneral} are reported to assess convergence. Right: Test log-likelihood values for the GG and Gamma PHIBP models.}
    \label{fig:simulated_posterior_gg_partial}
\end{figure}

As a sanity check, we first generate simulated data from our PHIBP model with ground truth parameters chosen as follows:
\[
\alpha_0=0.7, \theta_0=5.0, \,\, (\alpha_j, \theta_j) = \begin{cases}
    (0.3, 1.0) & \text{if } j \text{ is even} \\
    (0.6, 2.0) & \text{if } j \text{ is odd}
\end{cases},
\]
where $J = 4$. We then generate $M_j \sim \mathscr{P}(1000)$ documents for each group from the PHIBP model. We run three independent chains MCMC algorithm described in Section B.I to infer the parameters, where each chain is run for 80,000 steps with 40,000 burn-in steps. The samples are collected from three chains with a thinning interval of 10 steps. From the same model, we also generate test data with $m_j \sim \mathscr{P}(1000)$, and compare our samplers with GG prior and Gamma prior for their predictive ability. Due to the space limit, we defer most of the results to the supplementary materials. Figure \ref{fig:simulated_posterior_gg_partial} shows the partial results, showing that the posterior distributions of $\alpha_0$ and $\theta_0$ recover the true values, and the GG PHIBP has higher test log-likelihood values for test data.

\subsection{Real Data}
We apply our PHIBP model to a real-world microbiome dataset from \citep{willis2022estimating}, originally collected by \citep{lee2015microbial} (see details at: \citep{astrobiomike}). The dataset consists of 12 microbial samples collected from the seafloor rock at the Dorado Outcrop, with 4 from ``glassy'' basalts and 8 from ``altered'' basalts. In the original analysis, which relies on log-ratios, the high proportion of zero counts (42\% of the data) presents a fundamental challenge. Their approach requires perturbing the data by adding a "pseudocount" to every observation, a fix they note can drastically alter the resulting diversity estimates. Our primary goal is to demonstrate how our framework overcomes this very limitation by treating zeros as informative sampling outcomes. For a broader comparison with other prominent ecological models on this data, we refer the interested reader to the original analysis. Our focus here is therefore not to replicate that comparison, but to specifically isolate the contribution of the flexible GG prior versus the standard Gamma prior within our hierarchical framework. We also note that our results do not involve the use of covariates, although we have provided some indications of how they might be incorporated in Section~\ref{sec:Extensions}. 

The resulting dataset is a count matrix consisting of $\varphi = 1425$ unique ASVs (columns) across $J=12$ samples (rows). As noted by \citep{willis2022estimating}, only about 6\% of the ASVs are common across all 12 samples. This extreme sparsity, combined with the known sensitivity of competing methods to zero-handling, makes this dataset an ideal test case for the robustness of our model. In the context of the PHIBP, in the sequel, we may refer to the sample/community $j$ by group $j$.

\begin{figure}
    \centering
        \includegraphics[width=0.25\linewidth]{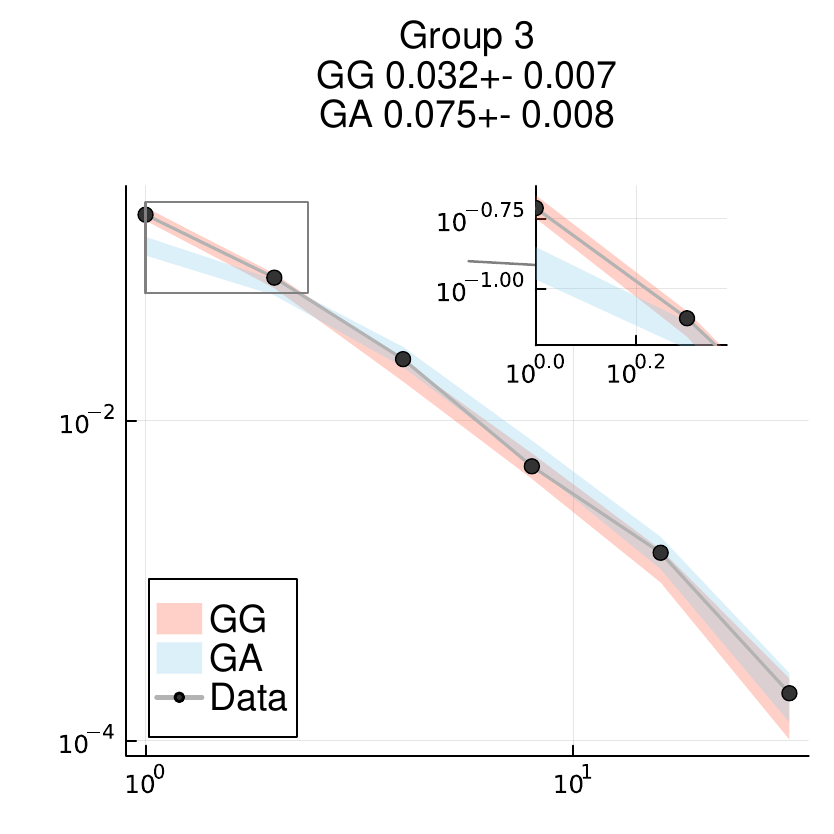}
    \includegraphics[width=0.25\linewidth]{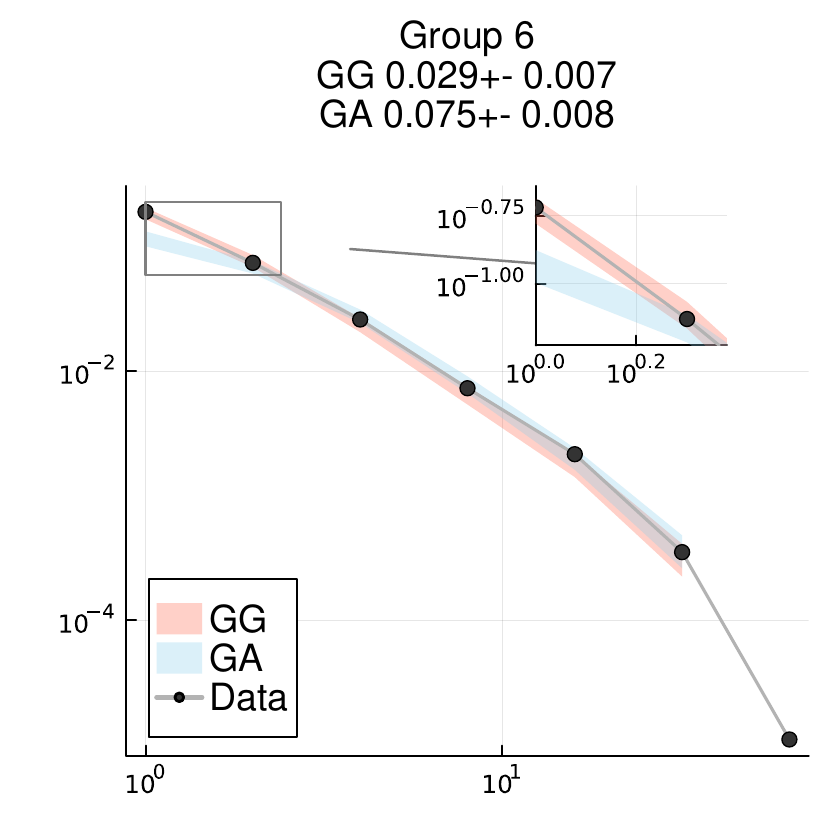}   
    \includegraphics[width=0.4\linewidth]{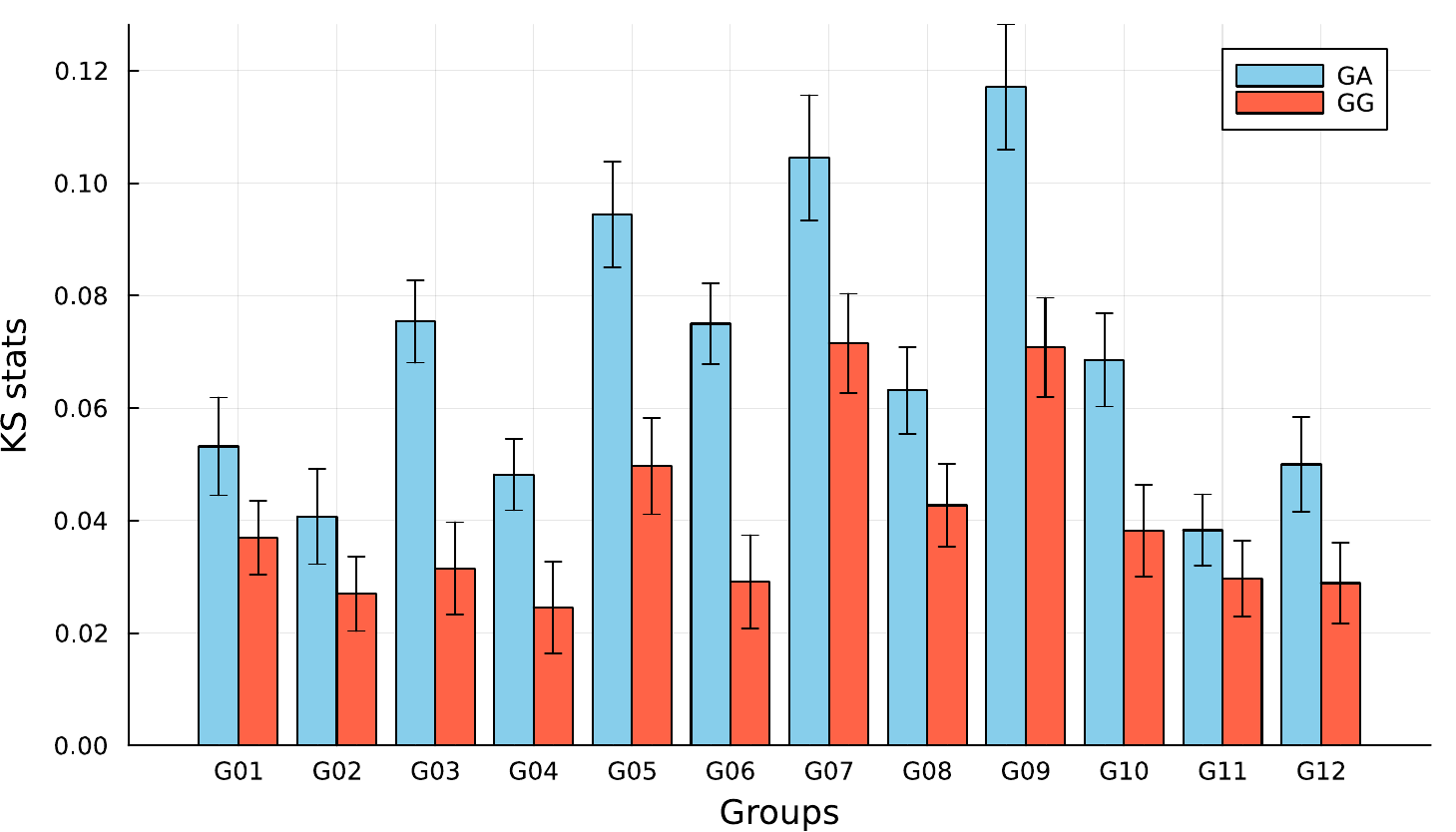}
    \caption{Left and Middle: Comparison of FoF distributions of the simulated data (under GG and gamma priors) and the real test data for Group 3 and Group 6. Right: Average Kolmogorov–Smirnov statistics between the FoF distributions of the simulated and real test data across all groups.}    
    \label{fig:microbiome_predictive_partial}
\end{figure}

To evaluate and compare the predictive performance of our models, we constructed an artificial split of the data by dividing each group into training and test subsets. Specifically, for each group $j$, given the observed count $N_{j,l}$ of species $l$, we draw a training count $N_{j,l}^{(\text{tr})} \sim \mathrm{Binom}(N_{j,l}, M_j / (M_j + m_j))$, where $M_j$ and $m_j$ denote the number of training and test subsamples in group $j$, respectively. We fit the PHIBP model using the training subsamples under a GG prior and a gamma prior ($\alpha_j$ fixed to $0$ for all $j \in \{0\} \cup [J]$). For each model, we run three independent MCMC chains, each for 40,000 steps, discarding the first 20,000 as burn-in and applying a thinning interval of 10. Posterior distributions of key parameters are provided in the supplementary materials. 
\paragraph{Comparing simulated data to observed test data.}
Using the posterior samples obtained from the training data, we generate posterior abundance parameters via Theorem 3.1. Conditional on these abundance parameters, we simulate species count data for the test set and compare their statistical properties to those of the observed test data. In particular, for each group $j$, we examine the frequency-of-frequency (FoF) distribution, defined as the fraction $p_{j,k}$ of species with count $k$, akin to a degree distribution in network analysis. The FoF distribution of the real test data follows a power-law, a common pattern across many domains. The model with a GG prior accurately captures this power-law behavior, whereas the model with a gamma prior fails to do so. The left and middle panels of Figure 2 show the FoF distributions of the observed and simulated test data for both models. As highlighted in the plots, the Gamma PHIBP underestimates the occurrence of rare species—those with small counts—while the GG PHIBP successfully recovers them. To provide a quantitative comparison, we compute the Kolmogorov–Smirnov statistic between the FoF distributions of the real and simulated test data. The results, summarized in the right panel of Figure \ref{fig:microbiome_predictive_partial}, show that the GG PHIBP model consistently outperforms the Gamma PHIBP across all groups.

\paragraph{Comparing model fit on the test data}
In the appendix, we further compare the predictive likelihoods of the test data for the GG and Gamma PHIBP models, using the procedure outlined in Section C in the supplementary materials. The results show that the GG PHIBP model consistently achieves significantly higher predictive likelihoods than the Gamma PHIBP model, indicating that the GG prior provides a substantially better fit to the overall structure of the dataset.

\paragraph{Diversity measures}
We next examine diversity measures based on the posterior abundance parameters. Since these parameters are drawn from the posterior distribution, all computed diversity measures naturally come with associated uncertainty, represented as distributions. As a representative example of $\alpha$ diversity, we compute the Shannon entropy (Eq. \ref{Shannon1}) for each group, and defer the results for the remaining groups to Section C in the supplementary materials. Figure \ref{fig:microbiome_diversities_partial} displays the resulting distributions for Groups 3 and 6, where the GG PHIBP model yields consistently higher entropy values than the Gamma PHIBP model. This aligns with earlier observations: the GG prior successfully captures rare species, resulting in richer and more diverse populations, while the gamma prior fails to do so. We also compute $\beta$ diversities for all pairs of groups, and plot the average differences between the values computed from the gamma and GG PHIBP models in Figure \ref{fig:microbiome_diversities_partial}. Interestingly, the Gamma PHIBP model tends to produce larger $\beta$-diversity values than the GG PHIBP model. This result confirms our hypothesis from Section \ref{sec:diversity}: the gamma model, by failing to capture the full extent of rare but shared species, suffers from the exact problem of sparse co-occurrence leading to an artificial inflation of beta-diversity. In contrast, the GG PHIBP model provides a more robust and nuanced measure of dissimilarity by more faithfully representing the underlying abundances.

\begin{figure}
    \centering
    \includegraphics[width=0.31\linewidth]{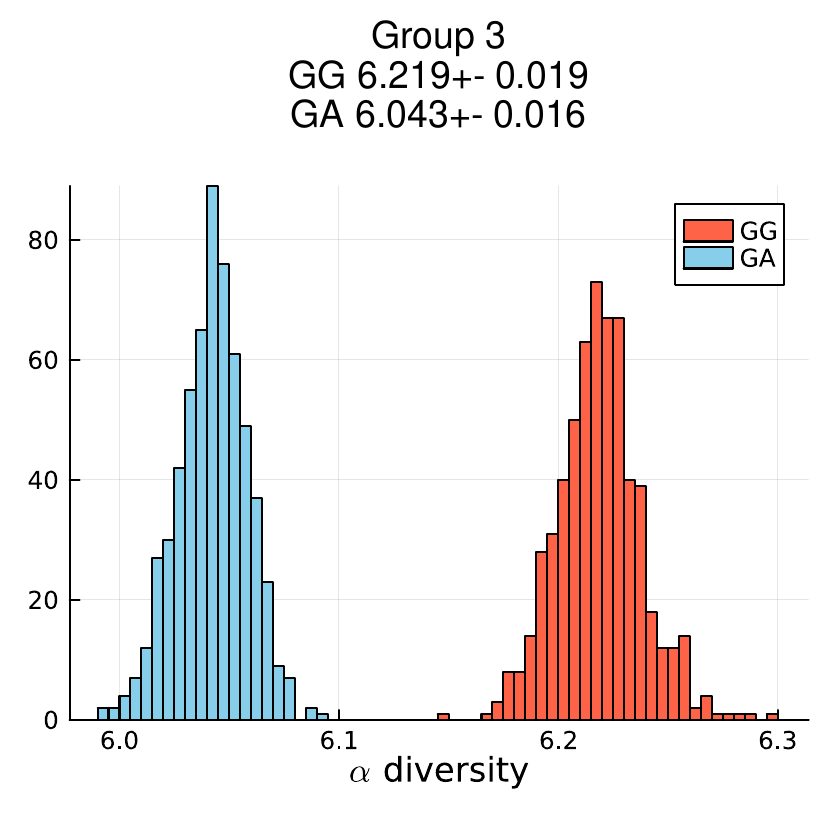}
    \includegraphics[width=0.31\linewidth]{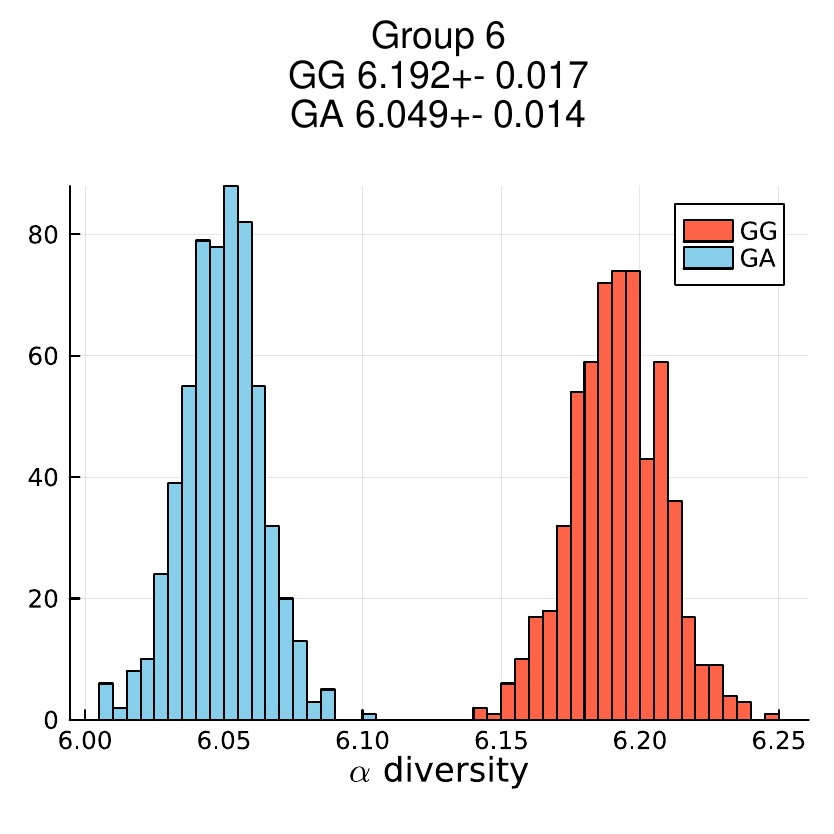}    
    \includegraphics[width=0.33\linewidth]{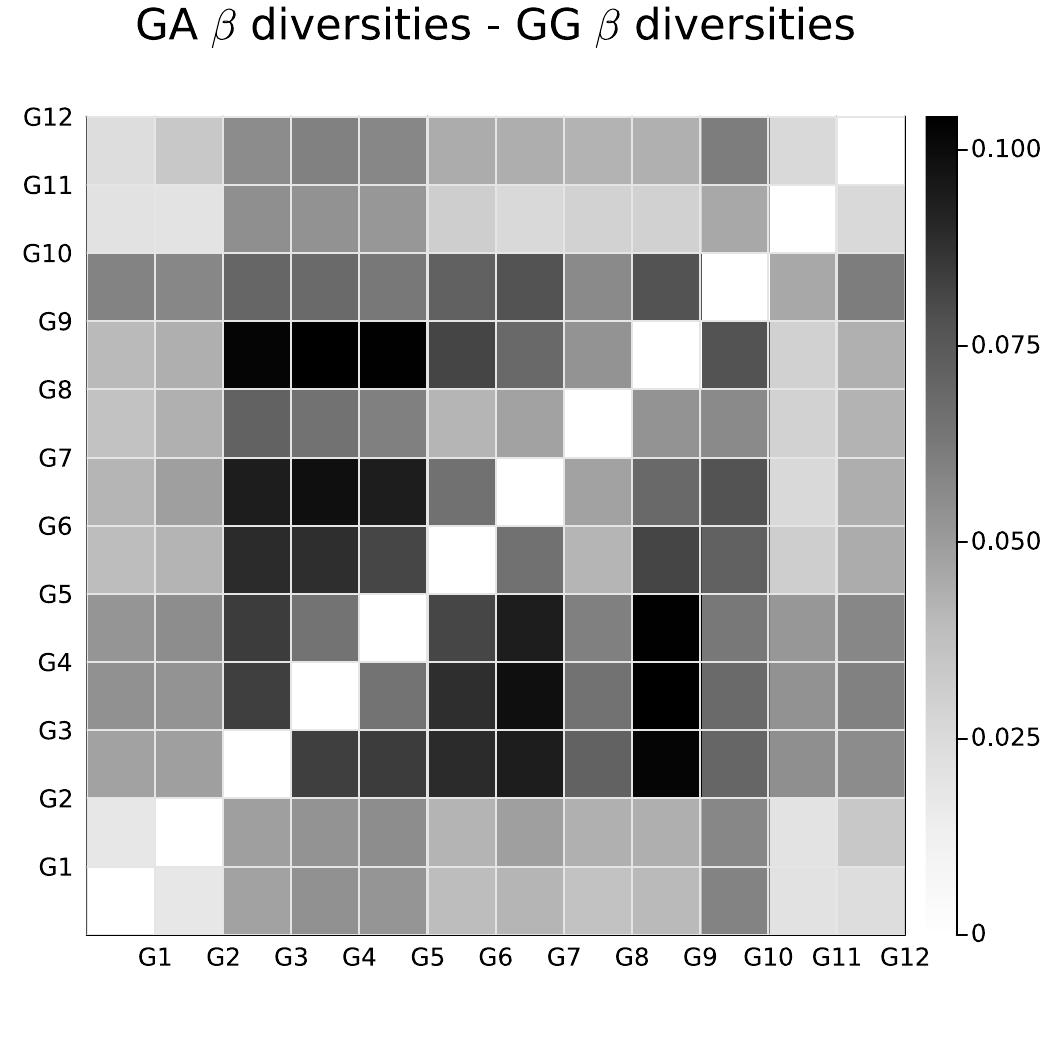}
    \caption{Left and Middle: posterior distributions of $\alpha$ diversities for Group 3 and Group 6. Right: Average differences of pairwise $\beta$ diversities from gamma and GG HIBP models.}   
    \label{fig:microbiome_diversities_partial}
\end{figure}

\begin{funding}
L.F. James was supported in part by grants RGC-GRF 16301521 of the Research Grants Council (RGC) of the Hong Kong SAR. 
\end{funding}

\appendix

\section{Proofs and discussions}
For historical context and to establish priority, we note that the initial technical concepts for this work appeared as part of a broader collection of results for other models in our unpublished manuscript~\cite{hibp23}. The present article provides the first focused and complete development of these specific ideas, with all results derived in a fully self-contained and updated manner, and with a focus on the microbiome that is native to this manuscript. An interesting insight in our paper is that hierarchical processes can be treated as multivariate IBP which can be handled by methods in ~\cite[Section 5]{James2017} but with quite specific care and focus, in particular, the development and exploitation of general compound Poisson representations, to obtain the best refinements. 

\subsection{Results in section 3} 
To make our approach clear, we will explain the method of decompositions as a translation of results in~\cite{James2017} as it relates to Proposition 3.1. We will continue throughout  as these mechanisms, which underlie the work in~\cite{James2017}, remain the same regardless of complexity. Here, when we mention IBP generically, we refer to general spike and slab IBPs in~\cite{James2017} but only deal with mixed Poisson based ones and their multivariate extensions. 

\begin{proof}[Proof of Proposition 3.1] Proposition 3.1 as mentioned in the main text, is mostly a rephrasing of the specific results in~\cite{James2017}[Sections 3 and 4.2]. However, the fact that we use inhomogeneous parameters $(\gamma_{i,j})$ makes this more general, yielding a fairly simple multivariate version of a Poisson IBP. Furthermore, any IBP based on $i\in[M_{j}]$ customers can be expressed as a multivariate process, which points to a treatment via the multivariate setting in~\cite{James2017}[Section 5]. First by construction, and as stated, the overall mean measure of the corresponding Poisson random measure governing the jumps and atoms of the process is 
$$
\tau_{j}(s)B_{0}(d\omega)ds=B_{0}(\Omega)\tau_{j}(s)\bar{B}_{0}(d\omega)ds
$$
where $\bar{B}_{0}(d\omega)=B_{0}(d\omega)/B_{0}(\Omega).$ Now borrowing a slight variation of the notation from~\cite{James2017}[Section 5], we set 
$A_{j,0}|s=(\mathscr{P}^{(i)}_{j}(\gamma_{i,j}s); i\in[M_{j}]),$ with 
  \[
   \mathbb{P}(\mathscr{P}^{(1)}_{j}(\gamma_{1,j}s)=a^{(1)}_{j},\ldots,\mathscr{P}^{(M_{j})}_{j}(\gamma_{M_{j},j}s)=a^{(M_{j})}_{j}|s) = \frac{s^{\sum_{i=1}^{M_{j}}a^{(i)}_{j}}e^{-s\Mj}\prod_{i=1}^{M_{j}}\gamma^{a^{(i)}_{j}}_{i,j}}{\prod_{i=1}^{M_{j}}a^{(i)}_{j}!}
        \]
The key is then the spike distribution $\pi_{A_{j,0}}(s)$, where
$$
1-\pi_{A_{j,0}}(s)=\mathbb{P}(\mathscr{P}^{(1)}_{j}(\gamma_{1,j}s)=0,\ldots,\mathscr{P}^{(M_{j})}_{j}(\gamma_{M_{j},j}s)=0|s)=e^{-s\sum_{i=1}^{M_{j}}\gamma_{i,j}}
$$
This "zero set" dictates the decomposition of the L\'evy density:
$$
\tau_{j}(s)=e^{-s\sum_{i=1}^{M_{j}}\gamma_{i,j}}\tau_{j}(s)+(1-e^{-s\sum_{i=1}^{M_{j}}\gamma_{i,j}})\tau_{j}(s)
$$
where
$$
\int_{0}^{\infty}(1-e^{-s\sum_{i=1}^{M_{j}}\gamma_{i,j}})\tau_{j}(s)ds=\psi_{j}(\sum_{i=1}^{M_{j}}\gamma_{i,j})
$$
leading to the decomposition 
$$
B_{j}=B_{j,M_{j}}+\sum_{\ell=1}^{\xi_{j}}S_{j,\ell}\delta_{\omega_{j,\ell}}
$$

indicating there are $\xi_{j} \sim \mathrm{Poisson}\left(B_{0}(\Omega)\psi_{j}(\sum_{i=1}^{M_{j}}\gamma_{i,j})\right)$ non-zero vectors of observations from $(Z^{(i)}_{j}; i \in [M_{j}])$. The collection $(S_{j,\ell})$ are iid with common density
$$
f_{S_{j}}(s)=\frac{1-e^{-s\Mj})\tau_{j}(s)}{\psi_{j}(\sum_{i=1}^{M_{j}}\gamma_{i,j})}
$$
and $(\omega_{j,\ell})\overset{iid}\sim \bar{B}_{0}.$

The multivariate slab distribution, expressed generally in~\citep[Section 5.1, p. 2036]{James2017}, specializes in this case to a multivariate truncated Poisson distribution, $\mathrm{tP}(s\gamma_{1,j},\ldots,s\gamma_{M_{j},j})$. This leads to the vector 
  \[
        \mathbb{P}(\tilde{C}^{(1)}_{j,\ell}=c^{(1)}_{j,\ell},\ldots,\tilde{C}^{(M_{j})}_{j,\ell}=c^{(M_{j})}_{j,\ell}|S_{j,\ell}=s) = \frac{s^{c_{j,\ell}}e^{-s\Mj}\prod_{i=1}^{M_{j}}\gamma^{c^{(i)}_{j,\ell}}_{i,j}}{(1-e^{-s\Mj})\prod_{i=1}^{M_{j}}c^{(i)}_{j,\ell}!}
        \]
        for $c_{j,\ell}=\sum_{i=1}^{M_{j}}c^{(i)}_{j,\ell} \ge 1$. and the marginal representation
  \begin{equation}
    \label{item:key sum}
        (Z^{(i)}_{j}, i\in[M_{j}])\overset{d}{=}\left(\sum_{\ell=1}^{\xi_{j}}\tilde{C}^{(i)}_{j,\ell}\delta_{\omega_{j,\ell}},i\in[M_{j}]\right).
    \end{equation}
which otherwise can be read from \cite[Prooposition 5.2]{James2017}, The remaining ingredients follow from applications of  Bayes rule as stated.
\end{proof}

\begin{rem}
Note, for a fixed $B_{0}$, the overall mean measure is $B_{0}(\Omega)\tau_{j}(s)\bar{B}_{0}(d\omega)$. The decomposition is dependent on $\tau_{j}$, not the nature of $\bar{B}_{0}$. That is, it doesn't matter whether $\bar{B}_{0}$ is discrete or continuous.
\end{rem}

\begin{proof}[Proof of Proposition 3.2, and 3.3] The allocation process 
\[
\mathscr{A}_{J} \overset{d}{=} \left( \sum_{l=1}^{\infty} \xi_{j,l} \delta_{Y_{l}}, \; j \in [J] \right),
\] 
is also a multivariate Poisson IBP of vector length $J$, based on the random base measure $B_{0}=\sum_{l=1}^{\infty}\lambda_{l}\delta_{Y_{l}}$, and the mean measure in question is $\tau_{0}(\lambda)F_{0}(dy)d\lambda$. The vector-valued components are
\begin{equation}
\label{Acomponents}
\mathbb{P}(\xi_{j,l}=x_{j,l}, j \in [J] | \lambda) = e^{-\lambda\sum_{j=1}^{J}\psi_{j}\left(\sum_{i=1}^{M_{j}}\gamma_{i,j}\right)} \prod_{j=1}^{J} \frac{\left(\lambda\psi_{j}\left(\sum_{i=1}^{M_{j}}\gamma_{i,j}\right)\right)^{x_{j,l}}}{x_{j,l}!}
\end{equation}
This leads to the identification of the zero-set probability:
\begin{equation}
\label{Acomponents2}
\mathbb{P}(\xi_{j,l}=0, j \in [J] | \lambda) = e^{-\lambda\sum_{j=1}^{J}\psi_{j}\left(\sum_{i=1}^{M_{j}}\gamma_{i,j}\right)}
\end{equation}
with
$$
\int_{0}^{\infty}(1-e^{-\lambda\sum_{j=1}^{J}\psi_{j}(\sum_{i=1}^{M_{j}}\gamma_{i,j})})\tau_{0}(\lambda)\,d\lambda = \Psi_{0}(\sum_{j=1}^{J}\psi_{j}(\sum_{i=1}^{M_{j}}\gamma_{i,j})).
$$
This leads to the decomposition of $\tau_{0}$ as indicated and the representation
$$
B_{0}=B_{0,J}+\sum_{l=1}^{\varphi}H_{l}\delta_{Y_{l}}
$$
as indicated. The zero-truncated $\mathrm{tP}(\lambda\psi_{1}(\sum_{i=1}^{M_{1}}\gamma_{i,1}),\ldots, \lambda\psi_{J}(\sum_{i=1}^{M_{J}}\gamma_{i,J}))$ conditional distributions of the $\varphi$ iid vectors are obtained from \eqref{Acomponents} and \eqref{Acomponents2}. Giving the representation
   \begin{equation}
    \label{realizedAllocationprocess}
    \mathscr{A}_{J} \overset{d}{=} \left(\sum_{l=1}^{\varphi}X_{j,l}\delta_{\tilde{Y}_{l}}, j\in[J]\right)\overset{d}{=} \left( \sum_{l=1}^{\infty} \xi_{j,l} \delta_{Y_{l}}, \; j \in [J] \right),.
    \end{equation}
and the remaining results following from parametric Bayesian arguments, as indicated. The results in Proposition 3.3 follow directly or a direct application of results in~\citep[Section 4.2]{James2017}
\end{proof}
\begin{rem}The real contribution here is identifying the Allocation process $\mathscr{A}_{J}$ as a key, otherwise hidden, component in the various constructions. Its interpretation with respect to the Microbiome is a key insight. 
\end{rem}

We next treat the case of Theorem 3.3, while  as one can see from our further developments there are other representations for the processes 
$(Z^{(i)}_{j}, i\in [M_{j}]; j\in[J])$ our representations based on the identification of the Allocation process offer the most practically tractable Compound Poisson representations. 

\begin{proof}[Proof of Theorem 3.1] With our developments above, the results for Theorem 3.1 follow easily. Again Propostion 3.1 and basic properties of CRM using $B_{0}=\sum_{l=1}^{\infty}\lambda_{l}\delta_{Y_{l}}$ lead to the following representation, fro each $j\in[J]$
\[(Z^{(i)}_{j},i\in[M_{j}])\overset{d}=\left(\sum_{l=1}^{\infty}\left[\sum_{k=1}^{\xi_{j,l}}\tilde{C}^{(i)}_{j,k,l}\right]\delta_{Y_{l}},i\in[M_{j}]\right).
\]
The representation of the Allocation process in \ref{realizedAllocationprocess} clearly indicates how to obtain the resulting compound Poisson representations using $X_{j,l},$ $\varphi,$ $\tilde{Y}_{l}$ that reflect the non-zero entries of the processes $((Z^{(i)}_{j},i\in[M_{j}]), j\in[J]).$
\end{proof}

\subsection{Results in section 4} 
We now establish the results in Section 4. We apply a more direct approach exploiting the sum properties of Poisson variables, and directly the infinite divisibility of $\sigma_{j,l}(\lambda).$. 

\begin{proof}[Proof of Theorem 4.1]
We first summarize some properties/representations of the Poisson HIBP which showcase its mixed Poisson representations, and describe $(B_{j},j\in[J]),B_{0})$ as a multivariate CRM as in~\cite[Section 5]{James2017}, see also \citep{barndorff2001multivariate},. Using these representations, as we shall show,  a posterior analysis follows from \cite{James2017}[Section 5]. First as in that work,  it follows that with respect to the jumps of the processes $(B_{j},j\in[J]), B_{0}$ say 
$((\sigma_{j,l}(\lambda_{l}); j\in[J]),\lambda_{l})_{l\ge l}$ are the points of a Poisson random measures say $M(dt_{1},\ldots,dt_{J},d\lambda)$ with mean measure
\begin{equation}
\nu(d\mathbf{t},d\lambda):=\mathbb{E}[M(dt_{1},\ldots,dt_{J},d\lambda)]=\prod_{j=1}^{J}\mathbb{P}(\sigma_{j}(\lambda)\in dt_{j})\tau_{0}(\lambda)d\lambda
\end{equation}
This point is well known and can be read from \citep[Theorem 30.1 Chapter 6, pages 197-198]{Sato2013}.
Now we use the representation
 $\mathbf{Z}_{J}:=(Z^{(i)}_{j}, i\in[M_{j}],j\in [J])$ is equal in distribution to 
    \begin{equation}
    \label{poissonrep}
   (\sum_{l=1}^{\infty}\mathscr{P}^{(i)}_{j,l}(\gamma_{i,j}\sigma_{j,l}(\lambda_{l}))\delta_{Y_{l}}, i\in [M_{j}],j\in[J]),
    \end{equation}
which corresponds to a  mixed multivariate Poisson IBP, such that conditional on the jumps $((\sigma_{j,l}(\lambda_{l}); j\in[J]),\lambda_{l})_{l\ge l}$ each component has a multivariate Poisson distribution
\begin{equation}
    \label{jointPoisson}
\mathbb{P}(\mathscr{P}^{(i)}_{j}(\gamma_{i,j}t_{j})=n^{(i)}_{j}, i\in[M_{j}],j\in[J]|\mathbf{t},\lambda)=e^{-\sum_{j=1}^{J}t_{j}\sum_{i=1}^{M_{j}}\gamma_{i,j}} \prod_{j=1}^{J} \prod_{i=1}^{M_{j}}\frac{{\left(t_{j}\gamma_{i,j}\right)}^{n^{(i)}_{j}}}{n^{(i)}_{j}!}
\end{equation}

and the zero set is then determined by
\begin{equation}
    \label{multizeroset}
\mathbb{P}(\mathscr{P}^{(i)}_{j}(\gamma_{i,j}t_{j})=0, i\in[M_{j}],j\in[J]|\mathbf{t},\lambda)=e^{-\sum_{j=1}^{J}t_{j}\sum_{i=1}^{M_{j}}\gamma_{i,j}}.
\end{equation}
which gives the decomposition
$$
\nu(d\mathbf{t},d\lambda)=e^{-\sum_{j=1}^{J}t_{j}\sum_{i=1}^{M_{j}}\gamma_{i,j}}\nu(d\mathbf{t},d\lambda)+
(1-e^{-\sum_{j=1}^{J}t_{j}\sum_{i=1}^{M_{j}}\gamma_{i,j}})\nu(d\mathbf{t},d\lambda)
$$
with 
\begin{equation}
\label{eq:integral_identity}
\int_{\mathbb{R}_{+}^{J+1}} \left(1-e^{-\sum_{j=1}^{J}t_{j}\sum_{i=1}^{M_{j}}\gamma_{i,j}}\right) \nu(d\mathbf{t},d\lambda) = \Psi_{0}(\sum_{j=1}^{J}\psi_{j}(\sum_{i=1}^{M_{j}}\gamma_{i,j}))
\end{equation}
This decomposition of $\nu$ leads to the decomposition of \( (B_{j}, j \in [J], B_{0}) \) as:
    \begin{equation}
    \label{postdisint}
    \left( B_{j,M_{j}} + \sum_{l=1}^{\varphi} \tilde{\sigma}_{j,l}(H_{l}) \delta_{\tilde{Y}_{l}}, j \in J, B_{0,J} + \sum_{l=1}^{\varphi} H_{l} \delta_{\tilde{Y}_{l}} \right).
    \end{equation}
Where \( B_{0,J} \sim \mathrm{CRM}(\tau_{0,J}, F_{0}) \), \( B_{j,M_{j}} \sim \mathrm{CRM}(\tau^{(M_{j})}_{j}, B_{0,J}) \), and have the same distributions as in Theorem 4.1. The \( (\tilde{Y}_{l}) \) are the i.i.d. \( F_{0} \) and represent the \( \varphi \sim \mathrm{Poisson}(\Psi_{0}(\sum_{j=1}^{J}\psi_{j}(\sum_{i=1}^{M_{j}}\gamma_{i,j}))) \) number of distinct features that are extracted from the \( \sum_{j=1}^{J} M_{j} \) samples.  

The $\varphi$ iid vectors \( (\tilde{\sigma}_{j,l}(H_{l}), j \in [J], H_{l}) \)  correspond to the distribution of the jumps where at least one
occurrence of the species $\tilde{Y}_{l}$ is observed across the $J$ groups in the \( \sum_{j=1}^{J} M_{j} \) samples and has joint distribution indicated by,
  $$
\frac{\left(1-e^{-\sum_{j=1}^{J}t_{j}\sum_{i=1}^{M_{j}}\gamma_{i,j}}\right) \nu(d\mathbf{t},d\lambda)}{\Psi_{0}(\sum_{j=1}^{J}\psi_{j}(\sum_{i=1}^{M_{j}}\gamma_{i,j}))}
$$
with respect to the non-zero component vectors of the observed sum process there is the information $(N_{j,l},\tilde{Y}_{l}, l\in[r],\varphi=r, j\in [J]),$ where again $N_{j,l}=\sum_{k=1}^{X_{j,l}}\tilde{C}_{j,k,l}.$ It follows from \eqref{jointPoisson} and \eqref{multizeroset}
that for each $\tilde{Y}_{l},$ $(N_{j,l},j\in[J])$ given $(\tilde{\sigma}_{j,l}(H_{l})=t_j, j \in [J], H_{l}=\lambda)$ has a zero-truncated multivariate Posson distribution  $\mathrm{tP}((t_{j} \sum_{i=1}^{M_{j}}\gamma_{i,j}, j \in [J]))$.

What remains to establish Theorem 4.1 is to determine the marginal and posterior quantities to describe the overall posterior distribution of $(B_{j}, j\in [J], B_{0})|\mathbf{Z}_{J}$ based on the joint distribution of $((N_{j,l}=n_{j,l},\tilde{\sigma}_{j,l}(H_{l})=t_{j}, j\in[J]), H_{l}),$ for each $l\in[\varphi]$
expressed as 
\begin{equation}
\label{fulljoint}
\frac{\prod_{j=1}^{J} (\Mj)^{n_{j,l}} t_{j}^{n_{j,l}} e^{-t_{j} \Mj}}{(1 - e^{-\sum_{j=1}^{J} \Mj t_{j}}) \prod_{j=1}^{J} n_{j,l}!} \cdot \frac{(1 - e^{-t_{j} \sum_{j=1}^{J} \Mj}) \prod_{j=1}^{J} \eta_{j}(t_{j} | \lambda)}{(1 - e^{-\lambda \sum_{j=1}^{J} \psi_{j}(\Mj)})} f_{H_{l}}(\lambda),
\end{equation}
where $\eta_{j}(t_{j}|\lambda)=\mathbb{P}(\sigma_{j}(\lambda)\in dt_{j})/dt_{j}.$ One can then use Bayes rule applied to \eqref{postdisint} to obtain a description of the posterior distribution given $((N_{j,l}, j\in[J]),\tilde{Y}_{l},l\in[r],\varphi=r)$ However, for tractability and indeed a more informative result useful for recovering hidden information, which we use,  we exploit the relation $N_{j,l}=\sum_{k=1}^{X_{j,l}}\tilde{C}_{j,k,l}.$ Since in fact we know the marginal distributions of $(X_{j,l},C_{j,k,l})$ we can just use directly, with appropriate constraints, 
\begin{equation}
\label{eq:CgivenNdist}
\prod_{l=1}^{r}\frac{\mathbb{P}({X}_{j,l}=x_{j,l};j\in[J])\prod_{j=1}^{J}\mathbb{P}(C_{j,k,l}=c_{j,k,l}, k\in[x_{j,l}]|\,X_{j,l}=x_{j,l})}{\mathbb{P}\left(\sum_{k=1}^{X_{1,l}}C_{1,k,l}=n_{1,l},\ldots,\sum_{k=1}^{X_{J,l}}C_{J,k,l}=n_{J,l}\right)}
\end{equation}

which, since it is a discrete distribution, can be easily sampled as we have shown. The last piece of the description in Thoerem 4.1 requires the representation of $\tilde{\sigma}_{j,l}(\lambda_{l})$, which is established in Proposition 5.2. We will say a bit more about this but may conclude the result with this point.  
\end{proof}

\begin{proof}[Proof of Proposition 4.3]
It follows from \eqref{fulljoint} that the joint distribution of $(N_{j,l},j\in[J]), (\tilde{\sigma}_{j,l}(H_{l}), j\in[J])| H_{l}=\lambda,$  has the form
\begin{equation}
\label{jointmargsigmaN}
\frac{\prod_{j=1}^{J} (\Mj)^{n_{j,l}}}{ \prod_{j=1}^{J} n_{j,l}!} \cdot \frac{ e^{-t_{j} \Mj}\prod_{j=1}^{J} t_{j}^{n_{j,l}} \eta_{j}(t_{j} | \lambda)}{(1 - e^{-\lambda \sum_{j=1}^{J} \psi_{j}(\Mj)})},
\end{equation}
which is equal to
\begin{equation}
\label{idProp5.2den}
\prod_{j=1}^{J}\frac{t_{j}^{n_{j,l}}\eta^{[0]}_{j}(t_{j}|\lambda,\Mj)}{\Xi^{[n_{j,l}]}(\lambda\tau_{j},\Mj)}
\times\frac{\prod_{j=1}^{J} (\Mj)^{n_{j,l}}}{ \prod_{j=1}^{J} n_{j,l}!}\frac{\prod_{j=1}^{J}e^{-\lambda\psi_{j}(\Mj)} \Xi^{[n_{j,l}]}(\lambda\tau_{j}, \Mj)}{(1-e^{-\lambda\sum_{j=1}^{J}\psi_{j}(\Mj)})}
\end{equation}
where $\eta^{[0]}_{j}(t_{j}|\lambda,\Mj):=e^{-t_{j}\Mj}\eta_{j}(t_{j}|\lambda)e^{\lambda\psi_{j}(\Mj)}.$ The expression in \eqref{idProp5.2den} establishes (i) of Proposition 4.3.

\begin{rem}
\eqref{idProp5.2den} identifies the densities of $\tilde{\sigma}_{j,l}(\lambda)|N_{j,l}=n_{j,l},H_{l}=\lambda$ as $\eta^{[n_{j,l}]}_{j}(t_{j}|\lambda,\Mj)$ which agrees with the descriptions in Proposition 5.2. 
\end{rem}
There are the identities, as can be read from ~\cite[Chapter 1]{Pit06}, see also~\cite{JamesStick},
\begin{align}
    \mathbb{E}\left[ (\sigma_{j}(\lambda))^{n_{j,l}} e^{-\sigma_{j}(\lambda)\psi_{j}(\Mj)} \right] 
    &= \int_0^{\infty} t^{n_{j,l}} e^{-t \Mj} \eta_j(t|\lambda) \,dt \notag \\
    &= e^{-\lambda\psi_{j}(\Mj)} \Xi^{[n_{j,l}]}(\lambda\tau_{j}, \Mj), \label{eq:xi_moment_identity_brief}
\end{align}
where $\Xi^{[0]}(\lambda\tau_{j}, \Mj)=\Xi^{[0]}_{0}(\lambda\tau_{j}, \Mj)=1,$ and  $\sum_{x_{j,l}=1}^{n_{j,l}} \Xi^{[n_{j,l}]}_{x_{j,l}}(\lambda\tau_{j}, \Mj)$ if $n_{j,l}>0,$ and where
\begin{equation}
\label{simpBellsid}
 \Xi^{[n_{j,l}]}_{x_{j,l}}(\lambda\tau_{j}, \Mj)=\lambda^{x_{j,l}}\Xi^{[n_{j,l}]}_{x_{j,l}}(\tau_{j}, \Mj)=\frac{\lambda^{x_{j,l}}_{j,l}n_{j,l}!{(\psi_{j}(\Mj))}^{x_{j,l}}}{{(\Mj)}^{n_{j,l}}x_{j,l}!}
\mathbb{P}\left(\sum_{k=1}^{x_{j,l}}\tilde{C}_{j,k,l}=n_{j,l}\right), 
\end{equation}
where as stated in the main manuscript, this and other relationships can be read from~\cite{Pit06}[Exercise 1.5.2, p. 33]. But otherwise can be deduced from elementary probability working with the explicit distribution of $(\tilde{C}_{j,k,l})\overset{iid}\sim \mathrm{MtP}(\tau_{j},\Mj).$

The joint distribution of $(N_{j,l}=n_{j,l}; j\in[J]),H_{l}=\lambda$  can be expressed as 
\begin{equation}
\frac{\prod_{j=1}^{J} (\Mj)^{n_{j,l}}}{ \prod_{j=1}^{J} n_{j,l}!}\frac{\prod_{j=1}^{J}e^{-\lambda\psi_{j}(\Mj)} \Xi^{[n_{j,l}]}(\lambda\tau_{j}, \Mj)f_{H_{l}}(\lambda)}{(1-e^{-\lambda\sum_{j=1}^{J}\psi_{j}(\Mj)})}
\label{HNjoint}
\end{equation}
Integrating over \eqref{HNjoint} leads to the expression in (ii) of Proposition 4.3. Where with reference to ~\eqref{eq:integral_identity}
\begin{equation}
\label{eq:integralexpcumulatnts}
\left(\Psi_{0}\circ \sum_{j=1}^{J}\psi_{j}\right)^{(\vec{n}_{l})}(\vec{\gamma})=\int_{\mathbb{R}_{+}^{J+1}}
\left[\prod_{j=1}^{J}t^{n_{j,l}} e^{-t_{j}\sum_{i=1}^{M_{j}}\gamma_{i,j}}\right]\nu(d\mathbf{t},d\lambda)
\end{equation}
is a joint cumulant function. 

In addition, augmenting the sum representation $\sum_{x_{j,l}=1}^{n_{j,l}} \Xi^{[n_{j,l}]}_{x_{j,l}}(\lambda\tau_{j}, \Mj)$ and using the identity in~\eqref{simpBellsid} it follows that
\begin{equation}
\label{simpid}
\frac{\int_{0}^{\infty}\prod_{j=1}^{J}e^{-\lambda\psi_{j}(\Mj)} \Xi^{[n_{j,l}]}_{x_{j,l}}(\lambda\tau_{j}, \Mj)f_{H_{l}}(\lambda)d\lambda}{(1-e^{-\lambda\sum_{j=1}^{J}\psi_{j}(\Mj)})}
\end{equation}
is the same as 
\begin{equation}
\label{simpid2}
=\frac{e^{-\Psi_{0}(\psi_{j}(\Mj))}\Psi^{(x_{l})}_{0}(\psi_{j}(\Mj)) \prod_{j=1}^{J}\Xi^{[n_{j,l}]}_{x_{j,l}}(\tau_{j}, \Mj)}{\Psi_{0}(\psi_{j}(\Mj))}
\end{equation}
This gives (iii) of Proposition 4.3 or one could have proceeded directly using \eqref{eq:CgivenNdist}.
\end{proof}

\subsubsection{Comments on Proposition 4.1}
The distribution  $\tilde{\sigma}_{j,l}(\lambda)|N_{j,l}=n_{j,l},H_{l}=\lambda$  with density in \eqref{idProp5.2den}
is equivalent to the distribution of 
$\sigma_{j,l}(\lambda)|\mathscr{P}_{j,l}(\sigma_{j,l}(\lambda)\Mj)=n_{j,l},$ 
This sets up relations of the variables $\tilde{C}_{j,k,l},X_{j,l}$ to corresponding variables arising in~\cite{JLP2,FoFZhou}, which either by direct comparison or as informed in the works of~\cite{PitmanPoissonMix} and \cite{Pit97,Kolchin} leads to the  finite Gibbs partitions in Proposition 4.1. 

\subsubsection{Establishing Propositions 4.2, and the remainder of Theorem 4.1}

As mentioned in the main text, Proposition 4.2 then follows immediately otherwise from the existing quoted results in \citep{James2002,JamesStick,JLP2}, yielding
$$
\tilde{\sigma}_{j,l}(\lambda)\overset{d}=\hat{\sigma}_{j,l}(\lambda)+\sum_{k=1}^{X_{j,l}}S_{j,k,l}
$$
For more detailed specifics about it derivations and connections to mixture representations and  posterior distributions of Poisson random measures see \cite[Sections 5.1-5.2, p. 27-32]{James2002} in particular Theorem 5.1 and Corollary 5.1 of that work. 
\subsubsection{Results for the general prediction rule in Proposition 4.4}
Once the posterior distribution of \( (B_{j}, j \in [J], B_{0}) | \mathbf{Z}_{J} \) has been established as in Theorem 4.1 or the discussion above, it is straightforward to obtain the prediction rule in Proposition 4.4. We provide a few details. From \eqref{postdisint} 

\begin{enumerate}
\item given $B_{j,M_{j}} + \sum_{l=1}^{\varphi} \tilde{\sigma}_{j,l}(H_{l}) \delta_{\tilde{Y}_{l}}$ it follows that 
$Z^{(M_{j}+1)}_{j}|B_{j,M_{j}} + \sum_{l=1}^{\varphi} \tilde{\sigma}_{j,l}(H_{l})\sim \mathrm{PoiP}(\gamma_{M_{j}+1,j}[B_{j,M_{j}} + \sum_{l=1}^{\varphi} \tilde{\sigma}_{j,l}(H_{l}) \delta_{\tilde{Y}_{l}}])$
\item Hence there is a decomposition of $(Z^{(M_{j}+1)}_{j},j\in[J])|\mathbf{Z}_{J}$ as a vector 
$$
(\hat{Z}_{j}+\sum_{l=1}^{r}\mathscr{P}_{j,l}(\gamma_{M_{j}+1,j}\tilde{\sigma}_{j,l}(H_{l}))\delta_{\tilde{Y}_{l}}, j\in[J])
$$
\item where further from Proposition 4.2 or directly from Theorem 4.1, 
$$
\mathscr{P}_{j,l}(\gamma_{M_{j}+1,j}\tilde{\sigma}_{j,l}(H_{l}))\overset{d}=
\mathscr{P}^{(2)}_{j,l}(\gamma_{M_{j}+1,j}\hat{\sigma}_{j,l}(H_{l}))+\mathscr{P}_{j,l}(\gamma_{M_{j}+1,j}\sum_{k=1}^{X_{j,l}}S_{j,k,l})
$$
where $\mathscr{P}^{(2)}_{j,l}$ are independent Poisson variables with random intensities as indicated
\item $\hat{Z}_{j}|B_{j,M_{j}}\sim \mathrm{PoiP}(\gamma_{M_{j}+1,j}B_{j,M_{j}})$ and $B_{j,M_{j}}|B_{0,J}\sim \mathrm{CRM}(\tau^{(M_{j})}_{j},B_{0,J})$ with $B_{0,J}\sim CRM(\tau_{0,J},F_{0})$
\item hence the vector $(\hat{Z}_{j},j\in[J]),$ has a representation of the form ~\eqref{poissonrep} is sampled according to Theorem 3.1, based on the variable $C^{(*,1)}_{j,k,l},X^{*}_{j,l}$ as indicated in Proposition Proposition 4.4
    \item \( \sum_{l=1}^{r}\mathscr{P}_{j,l}(\gamma_{M_{j}+1,j}\hat{\sigma}_{j,l}(H_{l})) \delta_{\tilde{Y}_{l}} \) has distribution \( \mathrm{PoiP}\left( \gamma_{M_{j}+1,j}\sum_{l=1}^{r} \hat{\sigma}_{j,l}(H_{l}) \delta_{\tilde{Y}_{l}} \right) \). Hence, the vector is also generated by applying Theorem 3.1 based on \( C^{(*,2)}_{j,k,l}, \mathscr{P}^{*}_{j,l}(H_{l}) \), where $\mathscr{P}^{*}_{j,l}(H_{l})\sim \mathrm{Poisson}(H_{l}[\psi_{j}(\sum_{i=1}^{M_{j}+1}\gamma_{i,j})-\psi_{j}(\Mj)]).$
\end{enumerate}
%\newpage
\def\[#1\]{\begin{equation}\begin{aligned}#1\end{aligned}\end{equation}}
\newcommand{\given}{{\hspace{0.08em}|\hspace{0.08em}}}
\newcommand{\tC}{\widetilde{C}}
\newcommand{\tN}{\widetilde{N}}
\newcommand{\tY}{\widetilde{Y}}
\newcommand{\bigcdot}{{\boldsymbol{\cdot}}}
\newcommand{\bakappa}{\kappa_{\boldsymbol{\cdot}}}
\newcommand{\bX}{\mathbf{X}}
\newcommand{\bZ}{\mathbf{Z}}
\newcommand{\bn}{\mathbf{n}}
\newcommand{\bbP}{\mathbb{P}}
\newcommand{\1}[1]{\mathds{1}_{\{#1\}}}
\newcommand{\calN}{\mathcal{N}}

\section{Details on the inference procedures}
\subsection{MCMC for Posterior Inference}
In this section, we present an MCMC algorithm for posterior inference of the Poisson HIBP model with a GG or gamma prior. Specifically, we choose
\[
\tau_0(\lambda) = \frac{\theta_0}{\Gamma(1-\alpha_0)}\lambda^{-\alpha_0-1}e^{-\zeta_0\lambda}, \quad
\tau_j(s) = \frac{\theta_j}{\Gamma(1-\alpha_j)}s^{-\alpha_j-1}e^{-\zeta_j s} \text{ for } j \in [J],
\]
where $\alpha_0, (\alpha_j)_{j\in [J]} \in [0, 1)$, $\theta_0, (\theta_j)_{j\in[J]} > 0$. We fix $\zeta_0 = \{\zeta_j\}_{j\in [J]} = 1$ and $(\gamma_{i,j}=1)$ for simplicity, but inferring them does not add extra components to the algorithm. 

For the posterior inference, like in many real-world scenarios, we assume that the only aggregated counts $N_{j,l} := \sum_{k=1}^{X_{j,l}} \tilde{C}_{j,k,l}$ are observed. From the observed counts, we need to recover the latent variables $X_{j,l}$ with which we compute the joint probability as below:
\begin{equation}
    \begin{aligned}
    \lefteqn{\mathbb{P}(
    \varphi = r,
    ((N_{j,l}=n_{j,l}, X_{j,l}=x_{j,l}), l \in [r]), j\in[J]))
    } \\
    &\propto \frac{\theta_0^r e^{-\Psi_0(\kappa_\cdot)}}{r!(\zeta_0 + \kappa_\cdot)^{x_\cdot - \alpha_0 r}} \prod_{l=1}^r \frac{\mathds{1}_{\{x_l>0\}}\Gamma(x_l-\alpha_0)}{\Gamma(1-\alpha_0)}        
    \prod_{j=1}^J \frac{\mathds{1}_{\{n_{j,l}\geq x_{j,l}\}}\theta_j^{x_{j,l}}}{(\zeta_j + M_j)^{n_{j,l}-\alpha_jx_{j,l}}} \frac{S_{\alpha_j}(n_{j,l}, x_{j,l})}{n_{j,l}!},
    \end{aligned}
\end{equation}
where
\begin{equation}
    \begin{aligned}
&x_{l} := \sum_{j=1}^J x_{j,l}, \quad x_{\cdot} := \sum_{l=1}^r x_l, \\
&\kappa_j := \psi_j(M_j) = \left\{\begin{array}{ll}
\frac{\theta_j}{\alpha_j}((\zeta_j + M_j)^{\alpha_j}-\zeta_j^{\alpha_j})
& \alpha_j \in (0, 1), \\
\theta_j \log(1 + M_j/\zeta_j) & \alpha_j = 0
\end{array}\right., 
\quad \kappa_\cdot := \sum_{j=1}^J \kappa_j, \\
&\Psi_0(\kappa_\cdot) = 
\left\{\begin{array}{ll}
\frac{\theta_0}{\alpha_0}((\zeta_0 + \kappa_\cdot)^{\alpha_0}-\zeta_0^{\alpha_0})     &  \alpha_0 \in (0, 1)\\
 \theta_0 \log(1 + \kappa_\cdot/\zeta_0)    &  \alpha_0 = 0 
\end{array}\right.
    \end{aligned}
\end{equation}
Note that the formulation above includes Gamma case ($\alpha_0 = 0$ or $\alpha_j = 0$). Given the observed count data $\mathbf{n} := ((n_{j,l}, l \in [r]), j\in [J])$, we simulate the joint posterior of the latent variables $\mathbf{X} := ((X_{j,l}, l\in[r]), j\in[J])$ and the parameters $\phi := \{\alpha_0, (\alpha_j)_{j\in [J]}, \theta_0, (\theta_j)_{j\in [J]}\}$. For the parameters, we introduce the following priors:
\begin{equation}
    \begin{aligned}
    &\pi_0(\alpha_0) = \mathrm{logit}\,\mathcal{N}(\alpha_0|0, 1) \propto \alpha_0^{-1}(1-\alpha_0)^{-1}\exp(-(\log(\alpha_0/(1-\alpha_0))^2/2), \\
    &\pi_0(\theta_0) = \log\,\mathcal{N}(\theta_0|0, 1) \propto \theta_0^{-1}\exp(-(\log\theta_0)^2), \\
    &\pi_0(\alpha_j) = \mathrm{logit}\,\mathcal{N}(\alpha_j|0, 1), \,\, \pi_0(\theta_j) = \log\,\mathcal{N}(\theta_j|0, 1) \text{ for } j \in [J].
    \end{aligned}
\end{equation}
The target posterior distribution to simulate is then defined as,
\[
\pi(\phi, \mathbf{X} | \mathbf{n}) \propto \mathbb{P}(   \varphi = r,
    ((N_{j,l}=n_{j,l}, X_{j,l}=x_{j,l}))) \pi_0(\alpha_0)\pi_0(\theta_0)\prod_{j=1}^J \pi_0(\alpha_j)\pi_0(\theta_j),
\]
where we are omitting the ranges of the indices for notational simplicity. From below, we present a MCMC algorithm to simulate the posterior.

\paragraph{Sampling $\mathbf{X}$} The conditional distribution for each $X_{j,l}$ is given as a discrete distribution, so we can employ Gibbs sampling.
\[
\mathbb{P}(X_{j,l}=x|\mathrm{others}) \propto \frac{\Gamma(x + x_l^{-j} - \alpha_0)\theta_j^x(\zeta_j + M_j)^{\alpha_j x}S_{\alpha_j}(n_{j,l}, x)}{(\zeta_0 + \kappa_\cdot)^x}, \quad x \in [n_{j,l}],
\]
where $x_l^{-j} := \sum_{j'\neq j} x_{j', l}$.
\paragraph{Sampling $\phi$} The parameters are sampled from via random-walk Metropolis-Hastings with the following proposal distributions:
\[
q(\alpha_j'|\alpha_j) = \mathrm{logit}\,\mathcal{N}(\alpha_j'| \log(\alpha_j/(1-\alpha_j)), \delta^2),\,\,
q(\theta_j'|\theta_j) = \log\,\mathcal{N}(\theta_j'|\log\theta_j, \delta^2) \text{ for } j \in \{0\} \cup [J].
\]

\subsection{Computing Predictive Likelihoods}
Proposition 4.4 outlines how to generate new populations conditioned on previously observed ones. Building on this, we now describe a procedure for computing the predictive likelihood of a set of test documents $((Z_j^{(M_j+i)},\ i \in [m_j]),\ j \in [J])$ for Poisson HIBP models, where $m_j$ denotes the number of novel documents (or customers) in group $j$. While Proposition 4.4 is stated for the special case $m_j = 1$ for all $j \in [J]$, the procedure naturally extends to arbitrary $m_j$. Let us first define some variables related to the counts,
\begin{equation}
    \begin{aligned}
&N_{j,v}^{(*,1)} \overset{d}{=} \sum_{k=1}^{X_{j,v}^*} \tilde{C}_{j,k,v}^{(*,1)}, \quad v \in [\varphi^*],\quad
 X_{j,l}^{(*,2)} \overset{d}{=} \mathscr{P}_{j,l}^*(H_l), \quad N_{j,l}^{(*,2)} \overset{d}{=} \sum_{k=1}^{X_{j,l}^{(*,2)}} \tilde{C}^{(*,2)}_{j,k,l}, \\
&\tilde{C}^{(*,3)}_{j,k,l} \overset{d}{=}
\mathscr{P}_{j,k,l}^{(M_j + m_j)}(S_{j,k,l}), \quad N_{j,l}^{(*,3)} \overset{d}{=} \sum_{k=1}^{X_{j,l}} \tilde{C}^{(*,3)}_{j,k,l}, \quad N_{j,l}^{(*,4)} \overset{d}{=} N_{j,l}^{(*,2)} + N_{j,l}^{(*,3)}.
    \end{aligned}
\end{equation}
From the test documents, we only observe the aggregated counts for the novel species $(N_{j,v}^{(*,1)})$,
and the aggregated counts for the existing species $(N_{j,l}^{(*,4)})$. However, these counts are not enough to compute predictive likelihoods as they are missing latent information, for instance,
the latent counts $X_{j,v}^*$ and $X_{j,l}^{(*,2)}$. From Proposition 4.4, we can compute the joint predictive likelihood of the observed latent counts and the latent counts. Firstly, the predictive likelihood for the novel species is computed as,
\begin{equation}
    \begin{aligned}
        \mathbb{P}(\varphi^*=r^*, (N_{j,v}^{(*,1)}=n_{j,v}^{(1)})\,|\, \mathbf{n})         
                &\propto \frac{\theta_0^{r^*}e^{-\Lambda_0(\kappa_\cdot, \kappa_\cdot^*)}}{r^*!(\kappa_\cdot^*+\kappa_\cdot+\zeta_0)^{x_\cdot^{(1)}-\alpha_0r^*}}
        \prod_{v=1}^{r^*}\frac{\mathds{1}_{\{x_v^{(1)}>0\}}\Gamma(x_v^{(1)}-\alpha_0)}{\Gamma(1-\alpha_0)} \\
        & \times \prod_{j=1}^J \frac{\mathds{1}_{\{n_{j,v}^{(1)} \geq x_{j,v}^{(1)}\}}\theta_j^{x_{j,v}^{(1)}}m_j^{n_{j,v}^{(1)}}}{(m_j+M_j+\zeta_j)^{n_{j,v}^{(1)}-\alpha_jx_{j,v}^{(1)}}} \frac{S_{\alpha_j}(n_{j,v}^{(1)}, x_{j,v}^{(1)}))}{n_{j,v}^{(1)}!},
    \end{aligned}
\end{equation}
where $x_v^{(1)} := \sum_{j=1}^J x_{j,v}^{(1)}$ and $x_\cdot^{(1)} := \sum_{v=1}^{r^*} x_v^{(1)}$ and,
\[
\kappa_j^* := \psi_j(M_j+m_j)-\psi_j(M_j), \quad \kappa_\cdot^* := \sum_{j=1}^J \kappa_j^*, \quad
\Lambda_0(\kappa_\cdot, \kappa_\cdot^*) := \Psi_0(\kappa_\cdot+\kappa_\cdot^*)-\Psi_0(\kappa_\cdot).
\]
Secondly, the predictive likelihood for observed species is computed as,
\begin{equation}
    \begin{aligned}
        \lefteqn{\mathbb{P}((N_{j,l}^{(*,4)}=n_{j,l}^{(4)})\,|\, \mathbf{n}, (X_{j,l}=x_{j,l}))} \\
        &  \propto \frac{(\kappa_\cdot + \zeta_0)^{x_\cdot-\alpha_0r}}{(\kappa_\cdot^* + \kappa_\cdot + \zeta_0)^{x_\cdot^{(2)}+x_\cdot - \alpha_0r}} \prod_{l=1}^r \frac{\mathds{1}_{\{x_l^{(2)}>0\}}\Gamma(x_l^{(2)}+x_l-\alpha_0)}{\Gamma(x_l-\alpha_0)} \\
        & \times  \prod_{j=1}^J  \frac{\mathds{1}_{\{n_{j,l}^{(4)}\geq x_{j,l}^{(2)}\}}\theta_j^{x_{j,l}^{(2)}} m_j^{n_{j,l}^{(4)}}}{(m_j+M_j+\zeta_j)^{n_{j,l}^{(4)}+n_{j,l}-\alpha_j (x_{j,l}^{(2)}+x_{j,l})}} \\
        &\times \left[\sum_{ n_{j,l}^{(2)}=x_{j,l}^{(2)}}^{n_{j,l}^{(4)}} 
\frac{\Gamma(n_{j,l}^{(4)}- n_{j,l}^{(2)} + n_{j,l}-\alpha_j x_{j,l})}{(n_{j,l}^{(4)}-n)!\Gamma(n_{j,l}-\alpha_jx_{j,l})}\frac{S_{\alpha_j}( n_{j,l}^{(2)}, x_{j,l}^{(2)})}{ n_{j,l}^{(2)}!}\right]^{\mathds{1}_{\{x_{j,l}>0\}}\mathds{1}_{\{x_{j,l}^{(2)} > 0\}}} \\
&\times \left[ \frac{S_{\alpha_j}(n_{j,l}^{(4)}, x_{j,l}^{(2)})}{n_{j,l}^{(4)}!} \right]^{\mathds{1}_{\{x_{j,l}=0\}}}
\left[ \frac{\Gamma(n_{j,l}^{(4)} + n_{j,l}-\alpha_j x_{j,l})}{n_{j,l}^{(4)}!\Gamma(n_{j,l}-\alpha_j x_{j,l})} \right]^{\mathds{1}_{\{x_{j,l}^{(2)}=0\}}}.
    \end{aligned}
\end{equation}
Note here that we are summing over all possible decompositions of $n_{j,l}^{(4)}$ into $n_{j,l}^{(2)} + n_{j,l}^{(3)}$, where $x_{j,l}^{(2)} \leq n_{j,l}^{(2)} \leq n_{j,l}^{(4)}$. Using the predictive likelihood, we can quantify how well the model predicts the test documents.

\section{Additional Results}
\subsection{Simulated Data}
\begin{figure}
    \centering
    \includegraphics[width=0.24\linewidth]{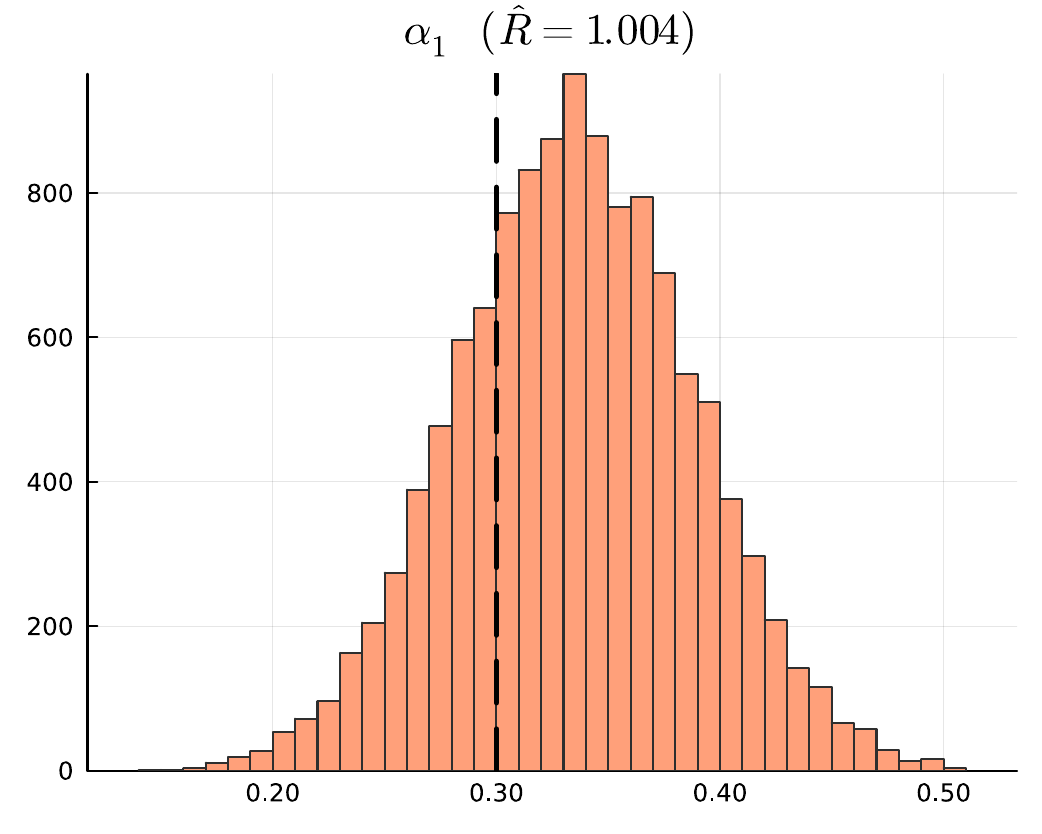}
    \includegraphics[width=0.24\linewidth]{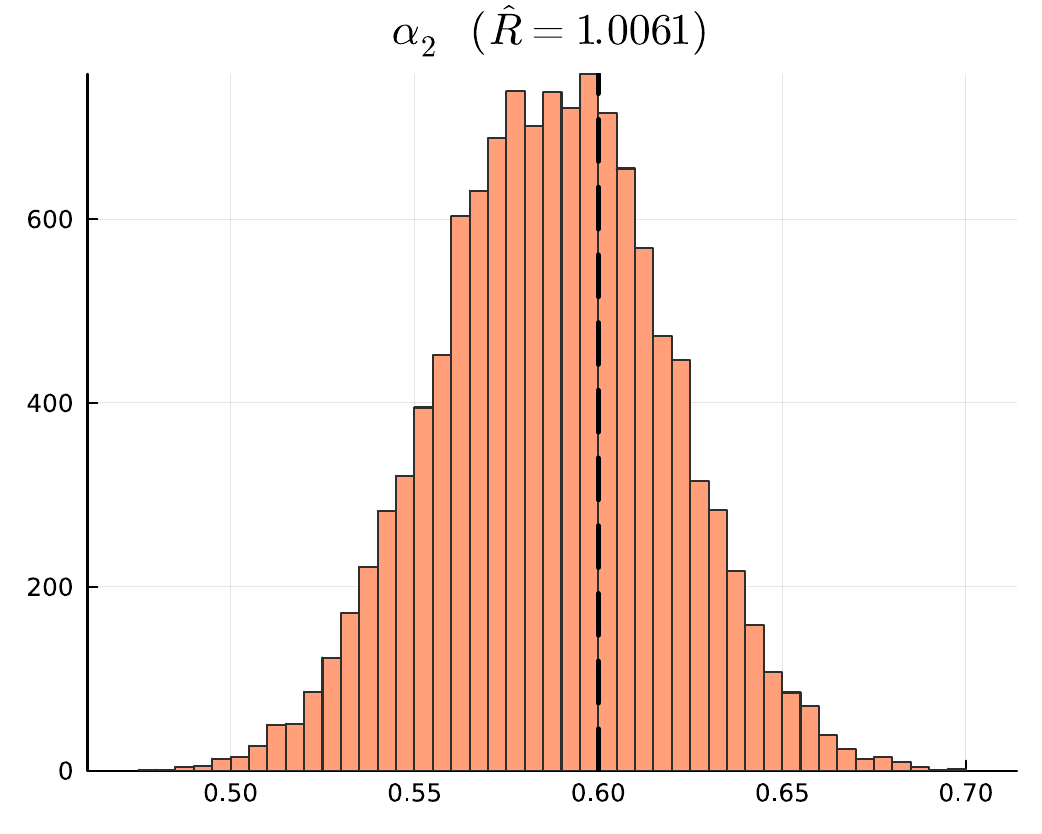}
    \includegraphics[width=0.24\linewidth]{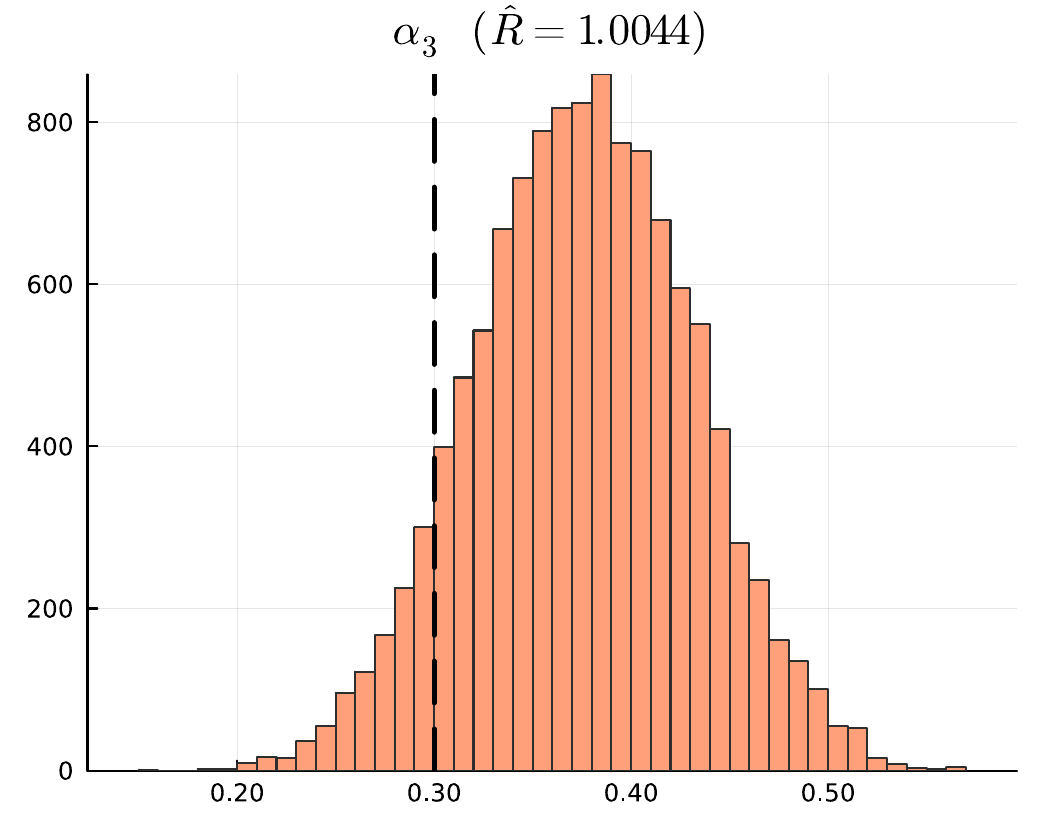}
    \includegraphics[width=0.24\linewidth]{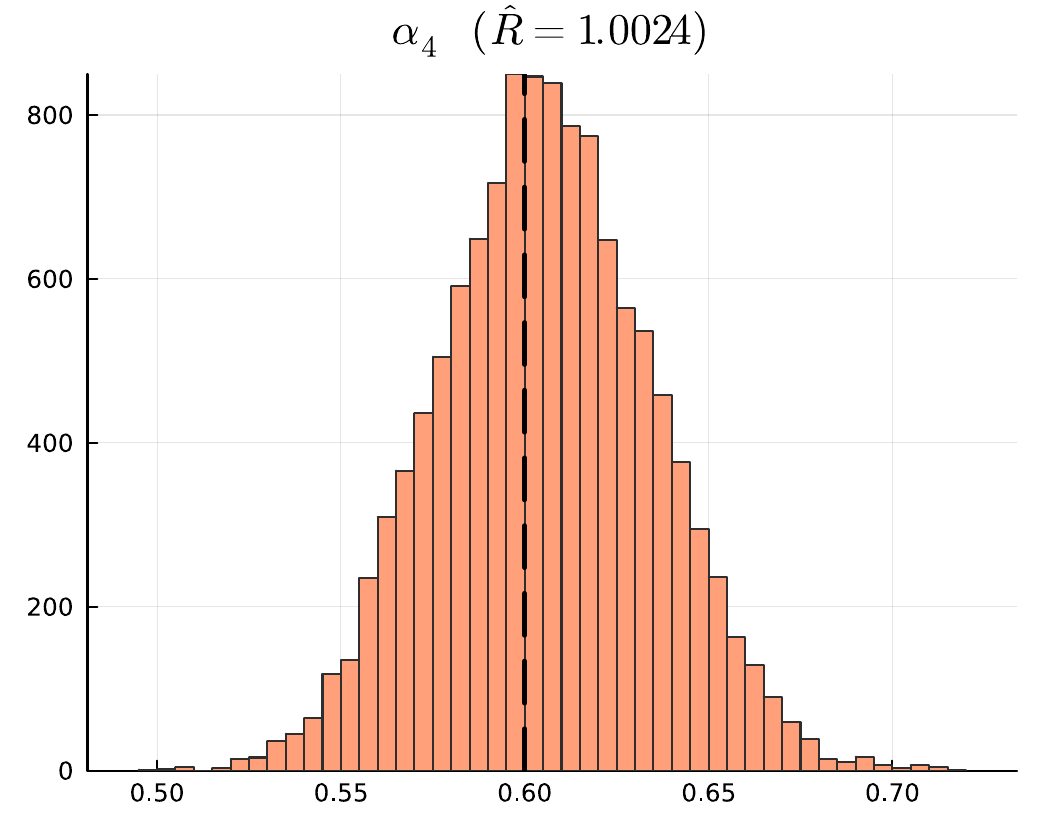}
    \includegraphics[width=0.24\linewidth]{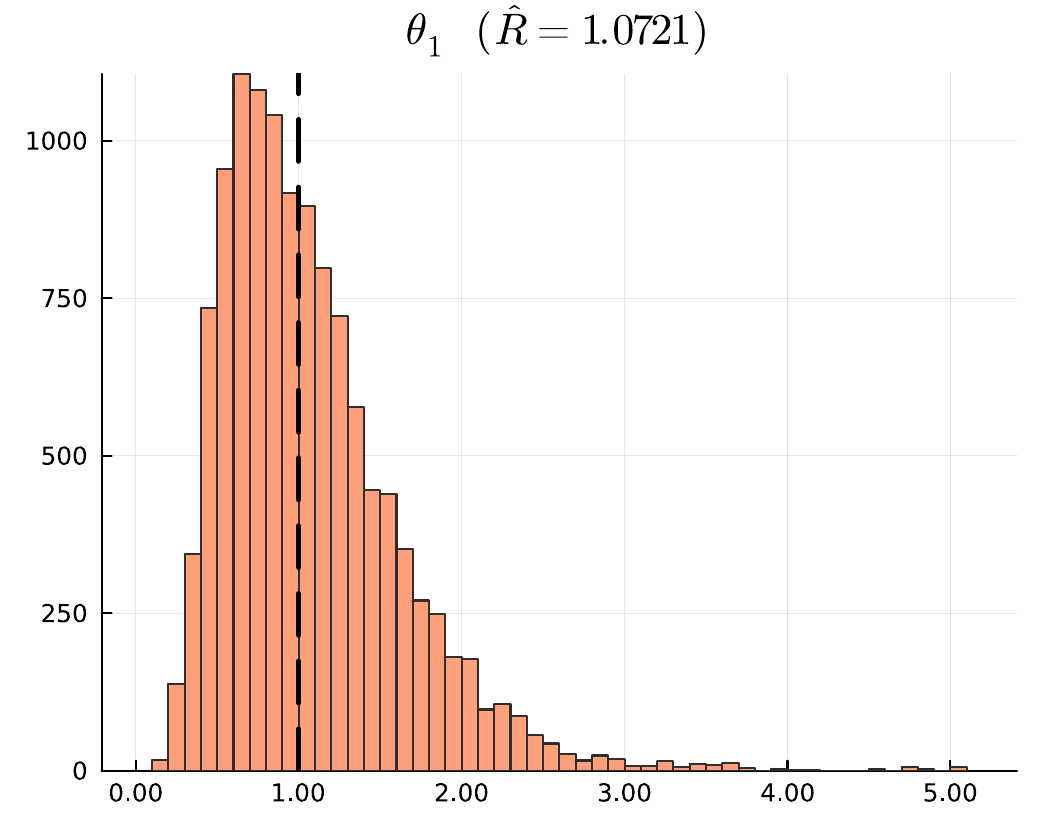}
    \includegraphics[width=0.24\linewidth]{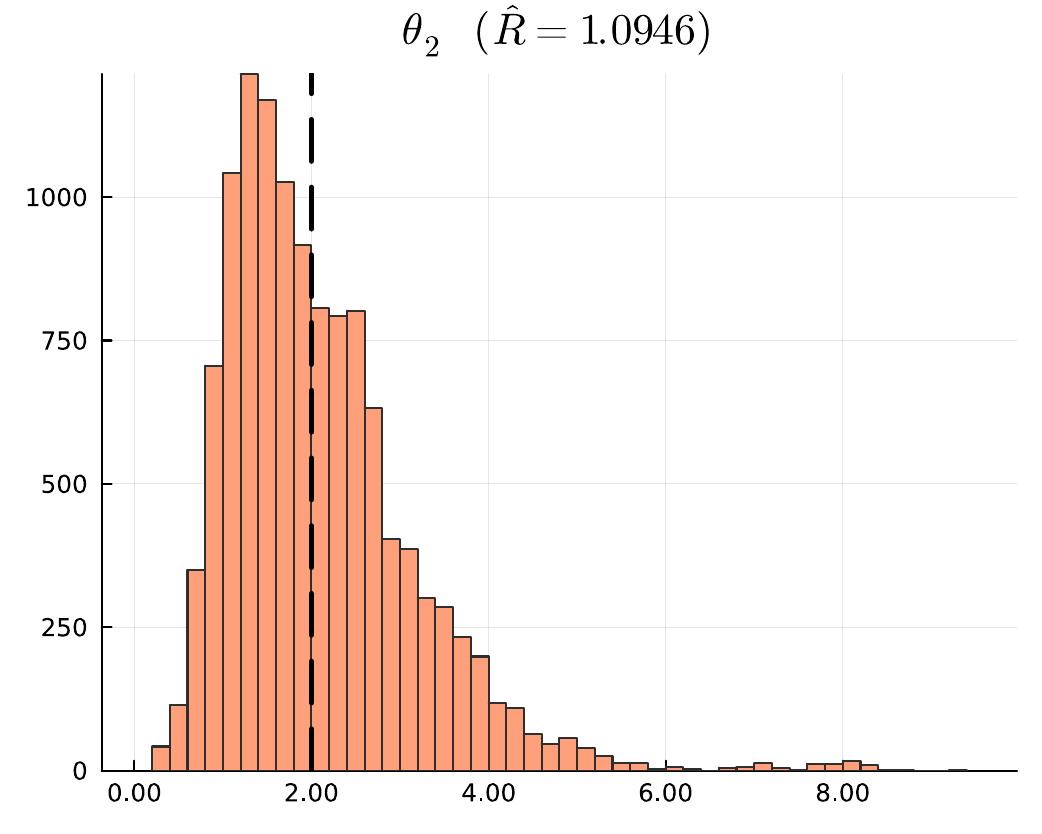}
    \includegraphics[width=0.24\linewidth]{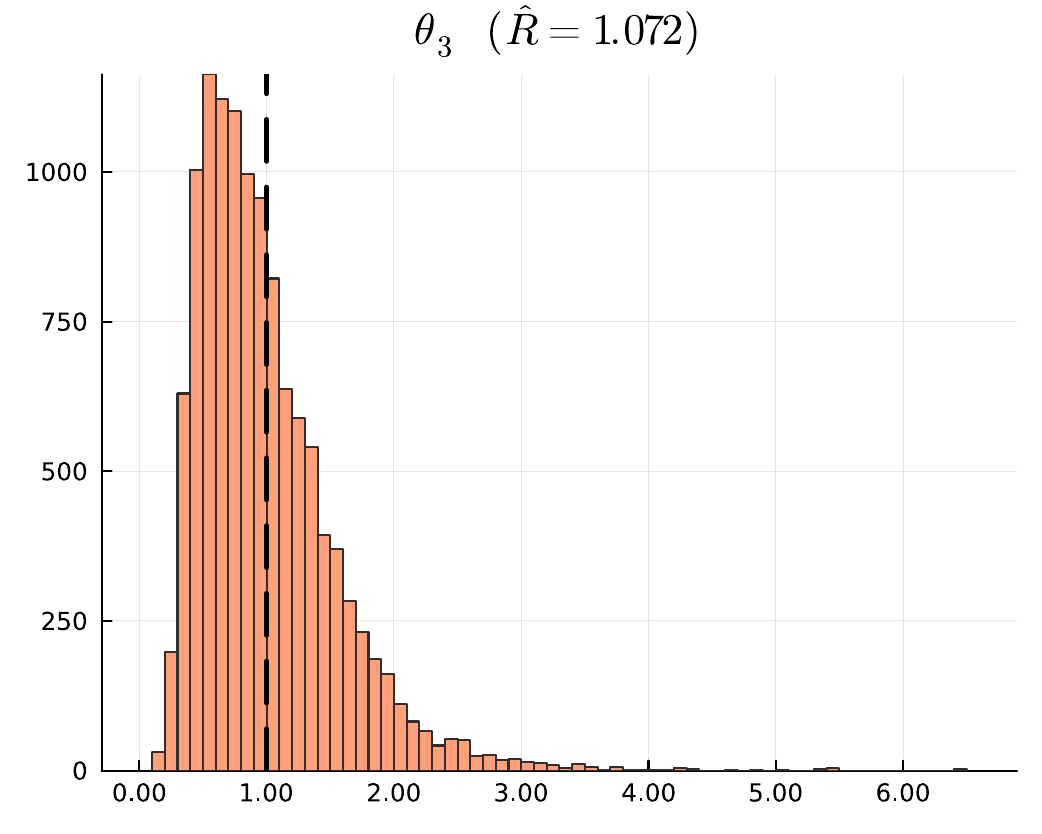}
    \includegraphics[width=0.24\linewidth]{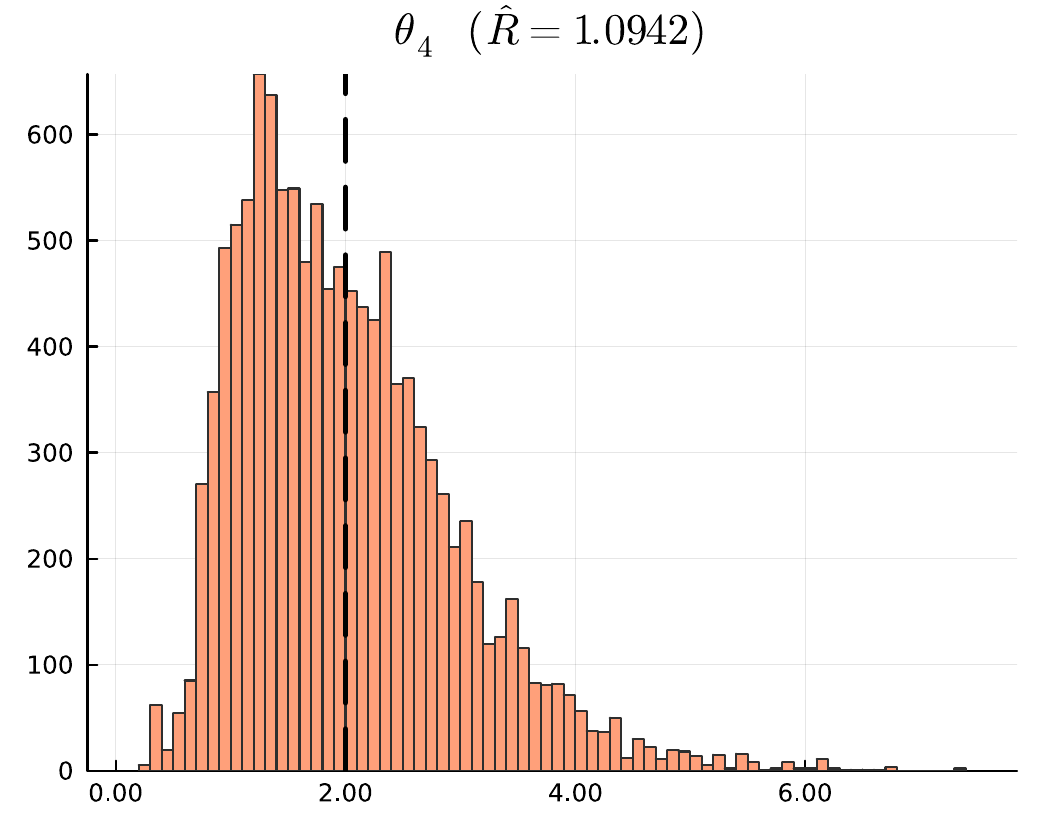}
    \caption{Posterior distributions of the GG HIBP model parameters inferred from the simulated data, along with corresponding $\hat{R}$ statistics. Black dashed lines denote the true values.}\label{fig:simulated_posterior_gg}    
\end{figure}

\begin{figure}
    \centering
    \includegraphics[width=0.32\linewidth]{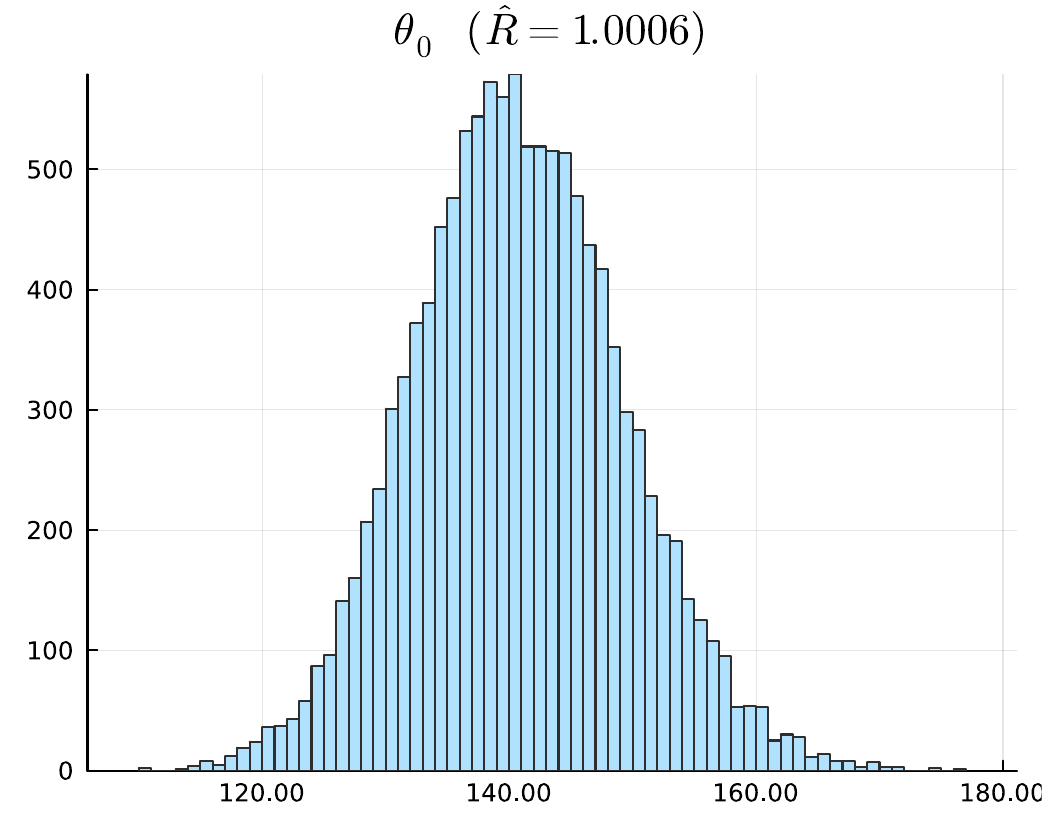}
    \includegraphics[width=0.32\linewidth]{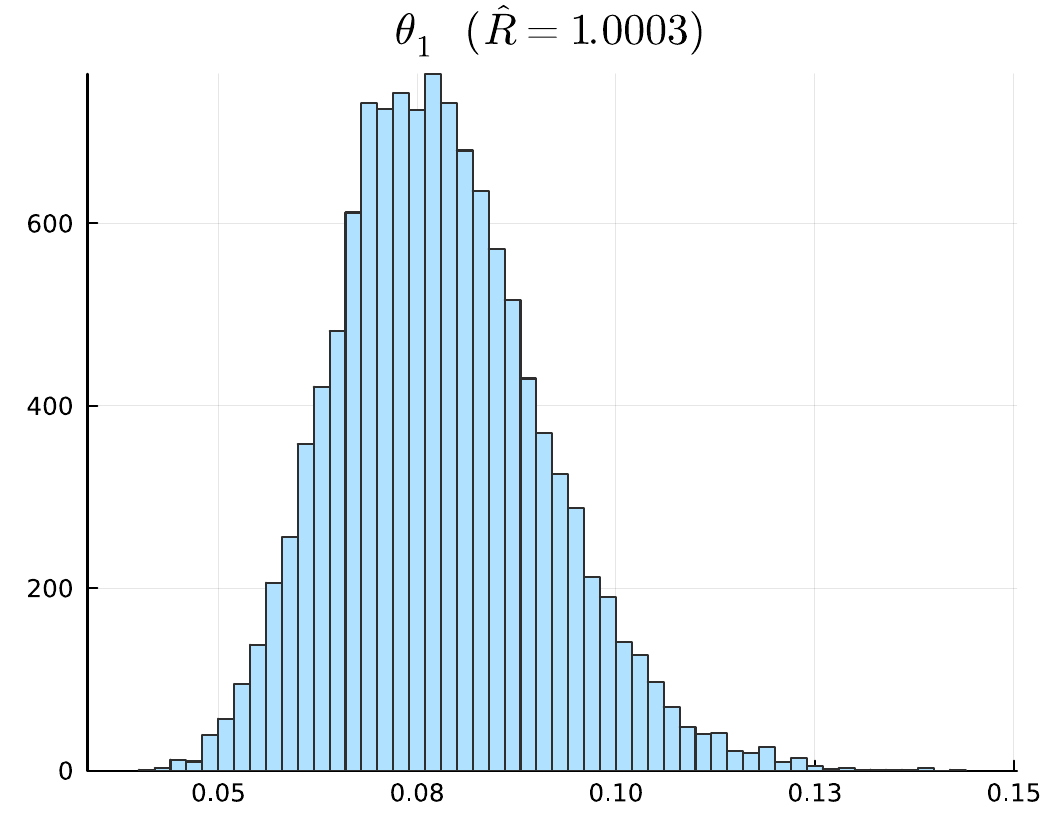}
    \includegraphics[width=0.32\linewidth]{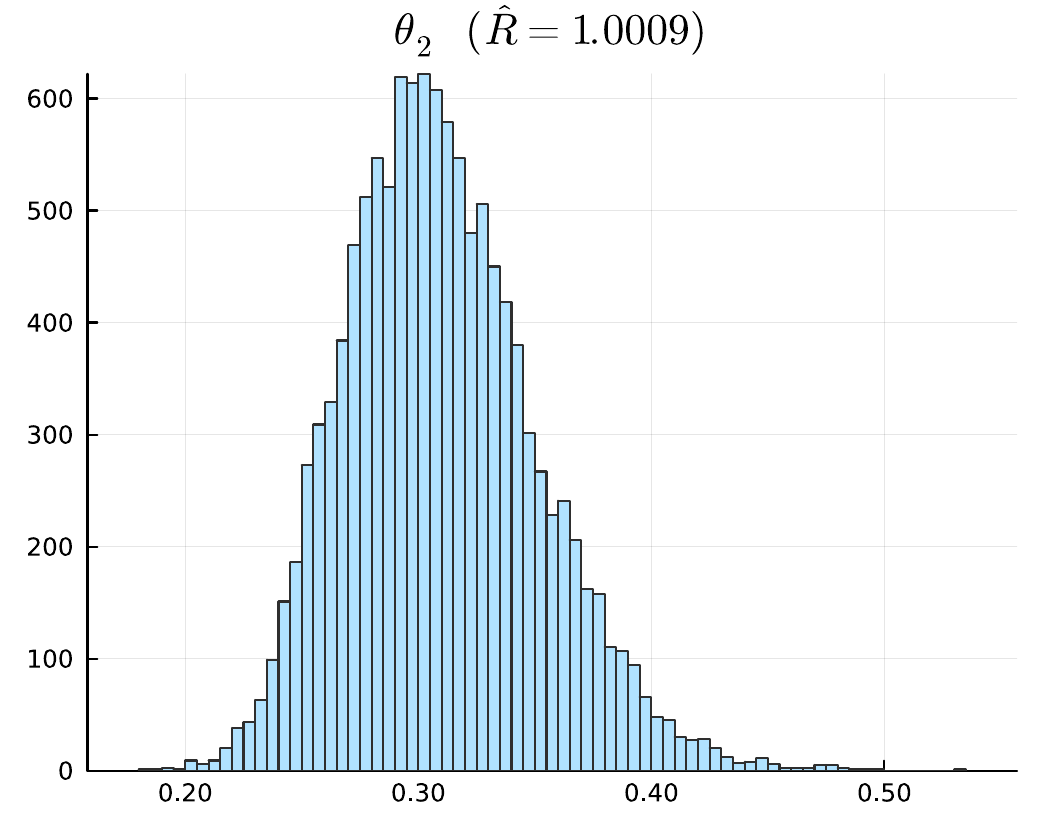}
    \includegraphics[width=0.32\linewidth]{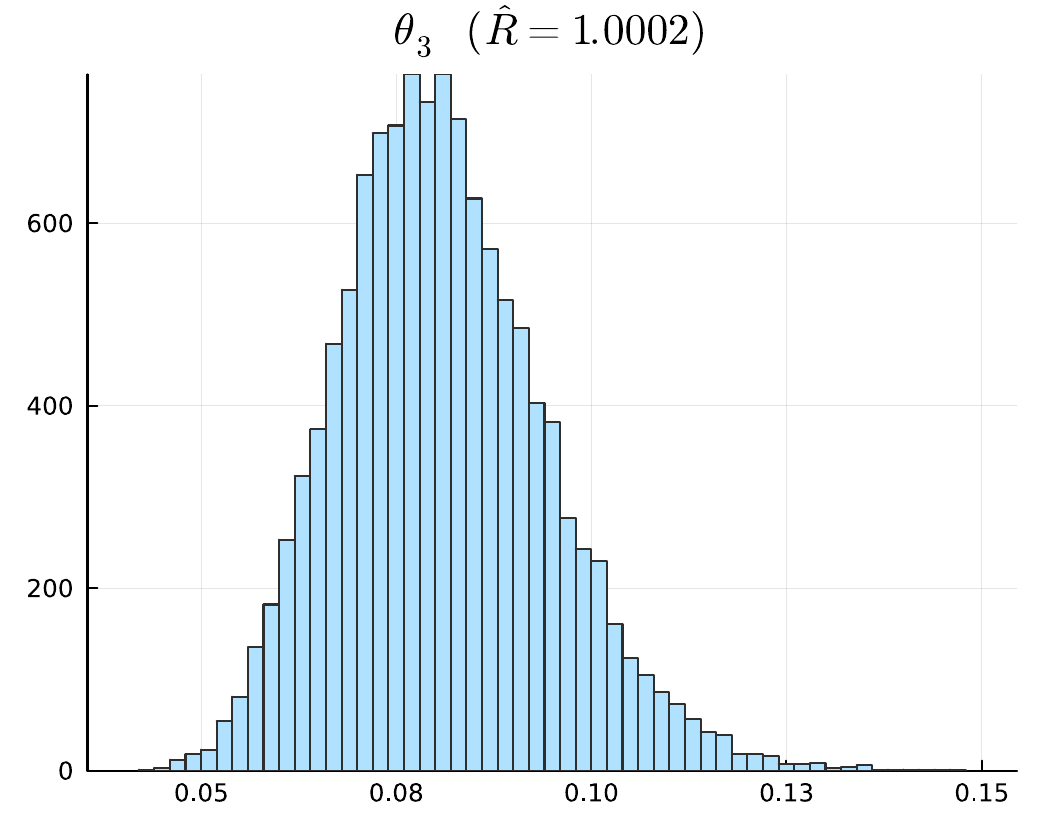}
    \includegraphics[width=0.32\linewidth]{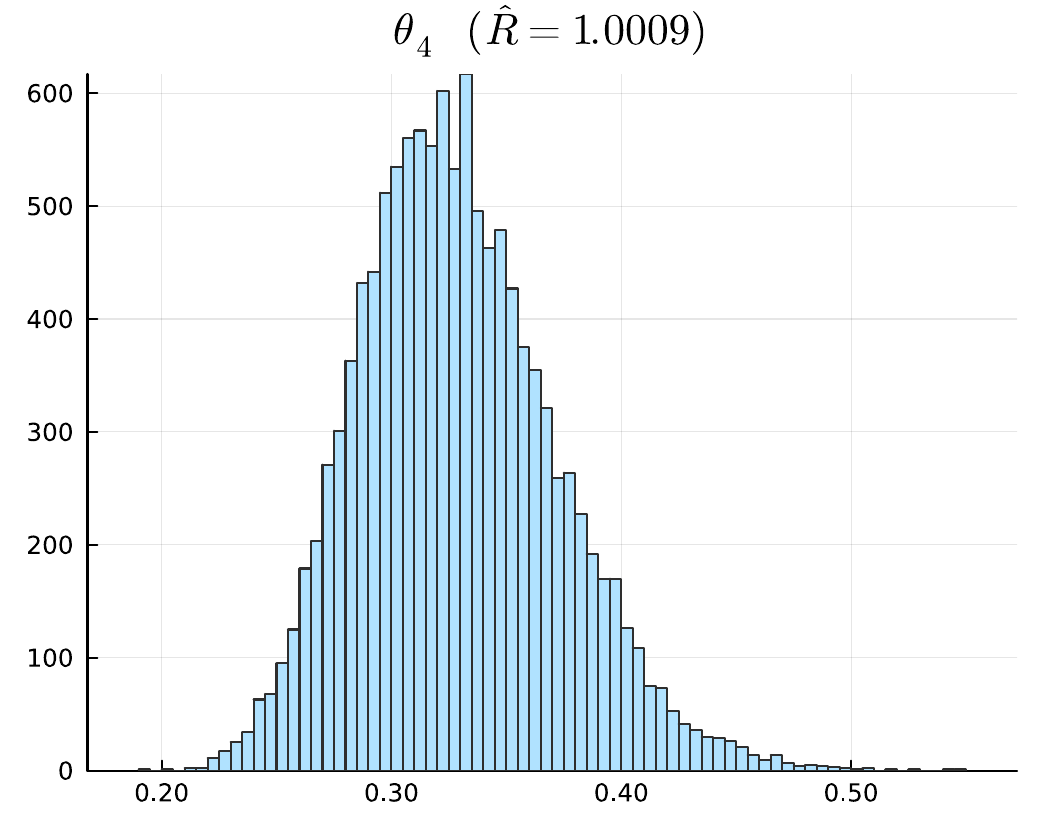}
    \caption{Posterior distributions of the Gamma HIBP model parameters inferred from the simulated data, along with corresponding $\hat{R}$ statistics. Black dashed lines denote the true values.}\label{fig:simulated_posterior_gamma}    
\end{figure}

Figure~\ref{fig:simulated_posterior_gg} presents the posterior distributions of the parameters other than $(\alpha_0, \theta_0)$. For all parameters, the sampler successfully concentrates around the true values.
We also present the posterior distributions of the parameters for Gamma HIBP model in Figure~\ref{fig:simulated_posterior_gamma}.

\subsection{Microbiome data}
\begin{figure}
    \centering
    \includegraphics[width=0.5\linewidth]{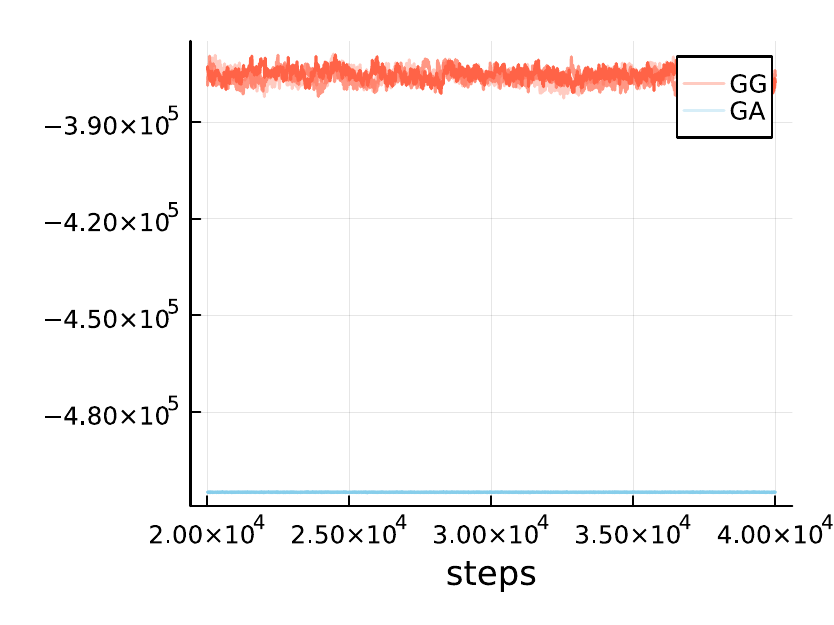}
    \caption{Test log-likelihood values of GG and Gamma HIBP models for microbiome data.}
    \label{fig:microbiome_test_ll}
\end{figure}

\begin{figure}
    \centering
    \includegraphics[width=0.19\linewidth]{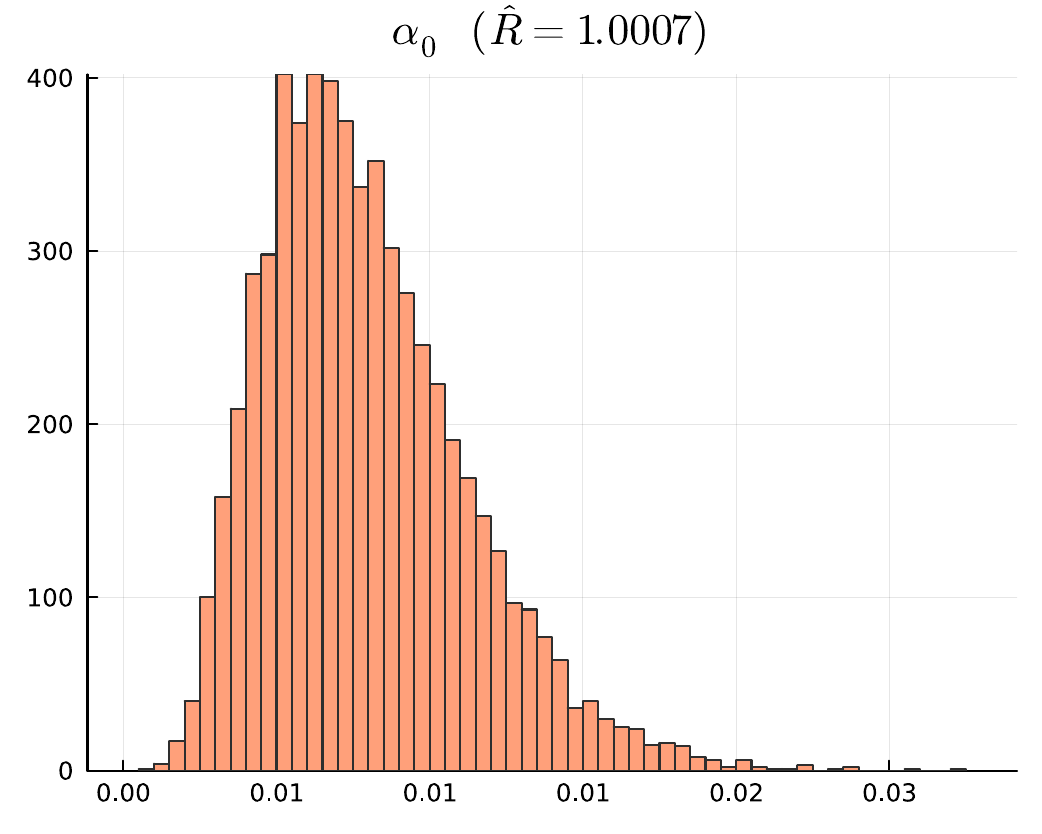}
    \includegraphics[width=0.19\linewidth]{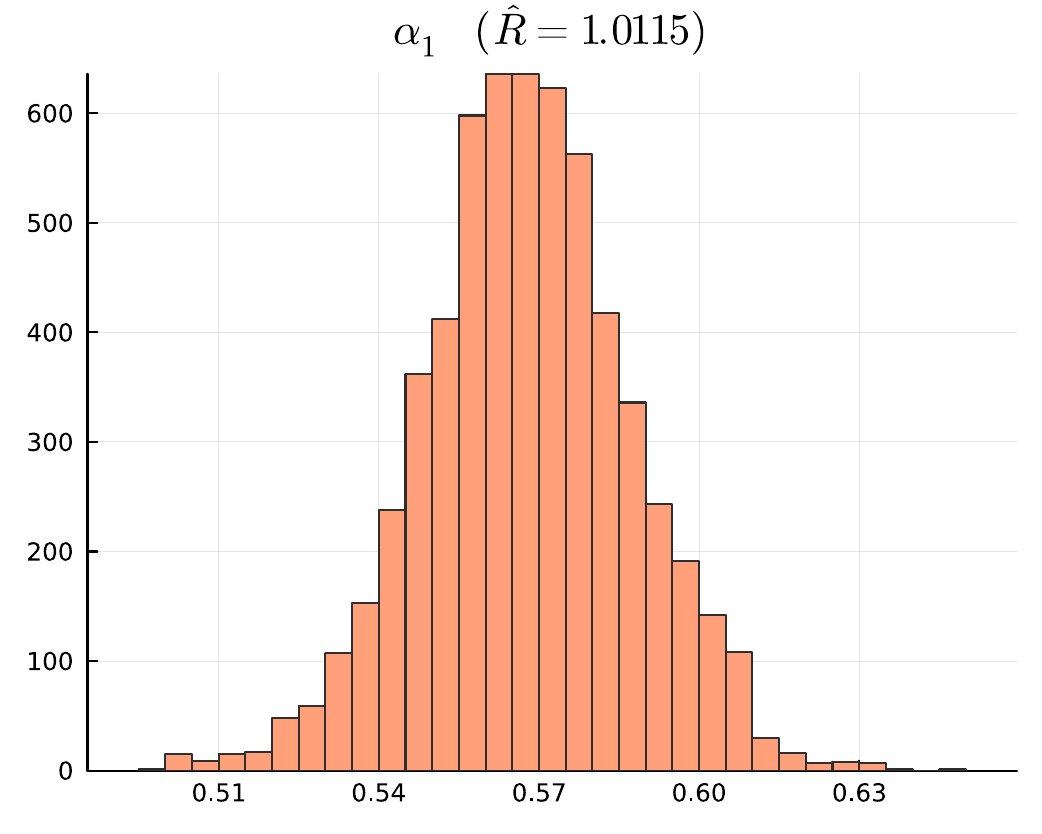}
    \includegraphics[width=0.19\linewidth]{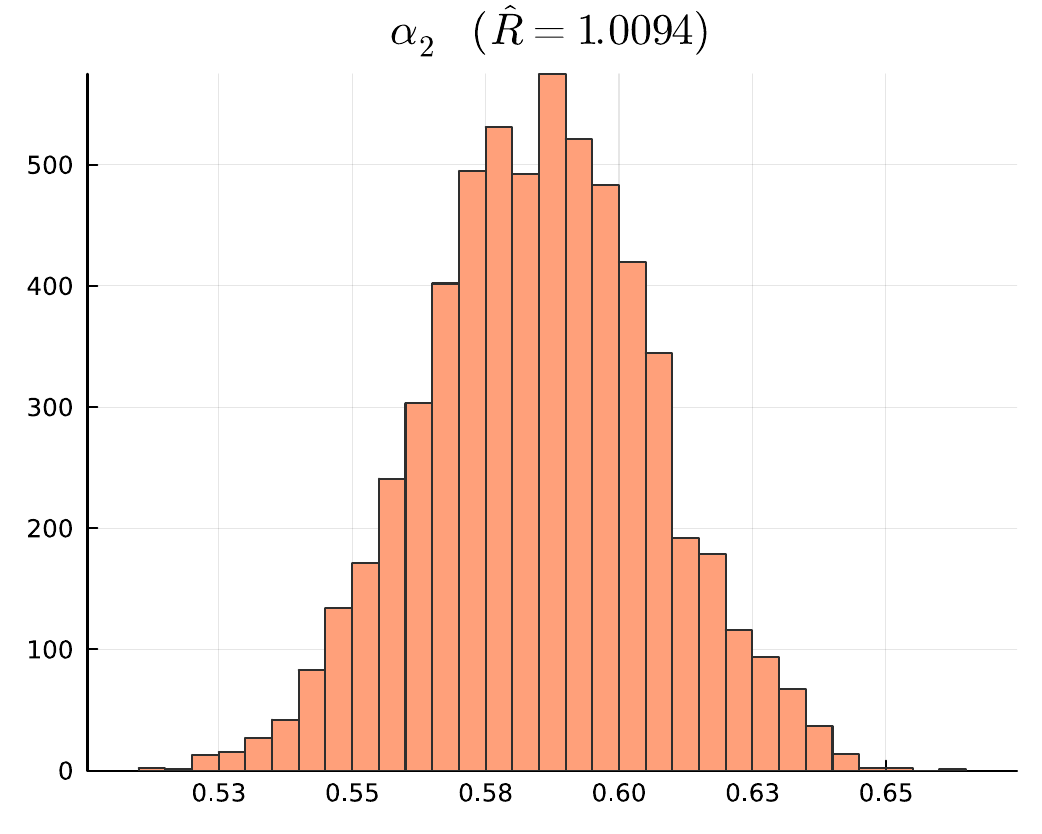}
    \includegraphics[width=0.19\linewidth]{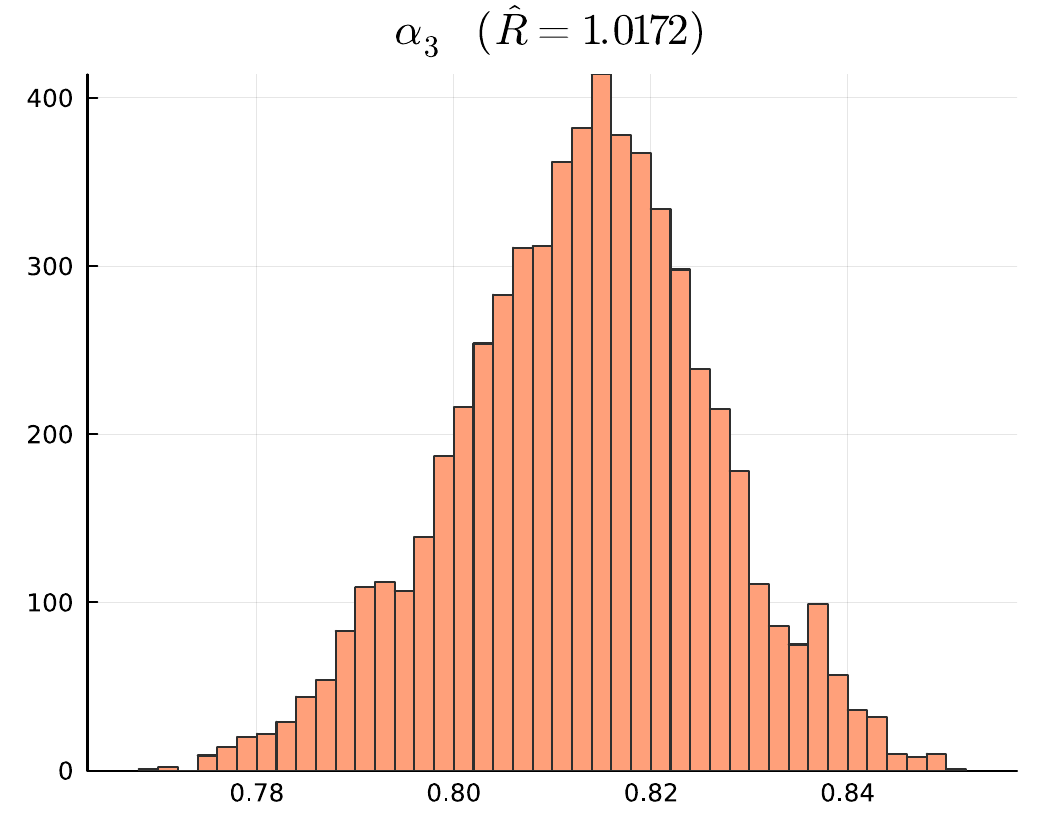}
    \includegraphics[width=0.19\linewidth]{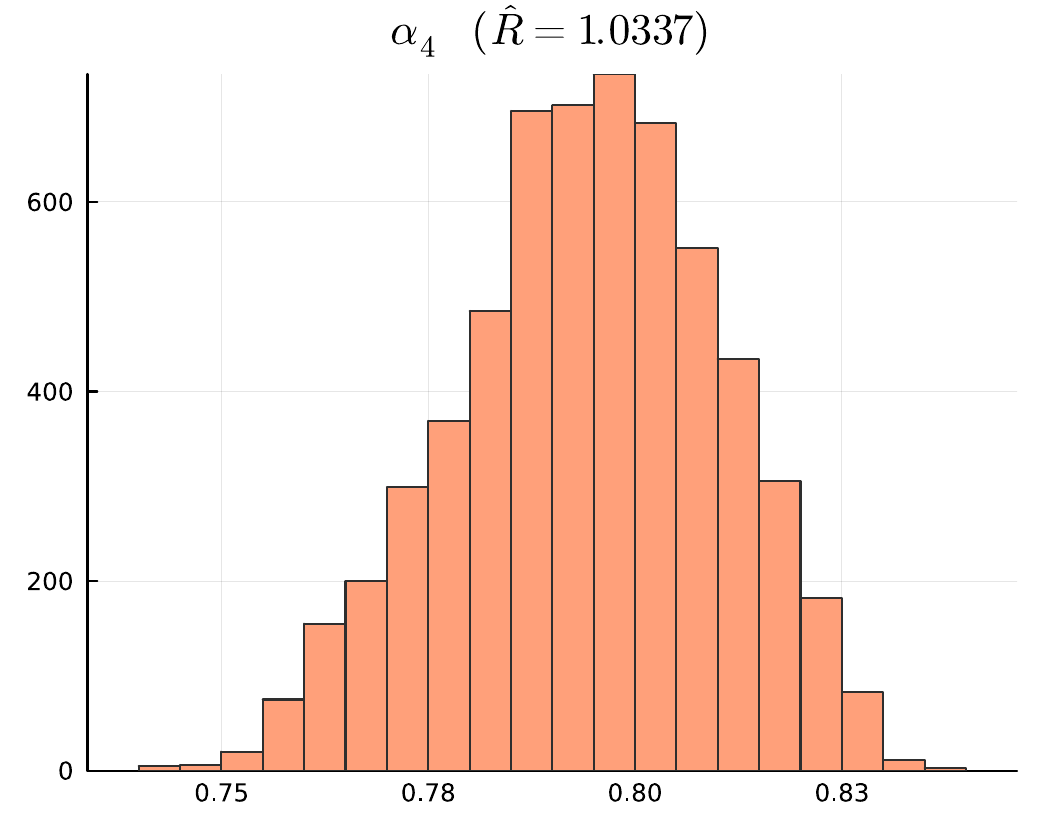}
    \includegraphics[width=0.19\linewidth]{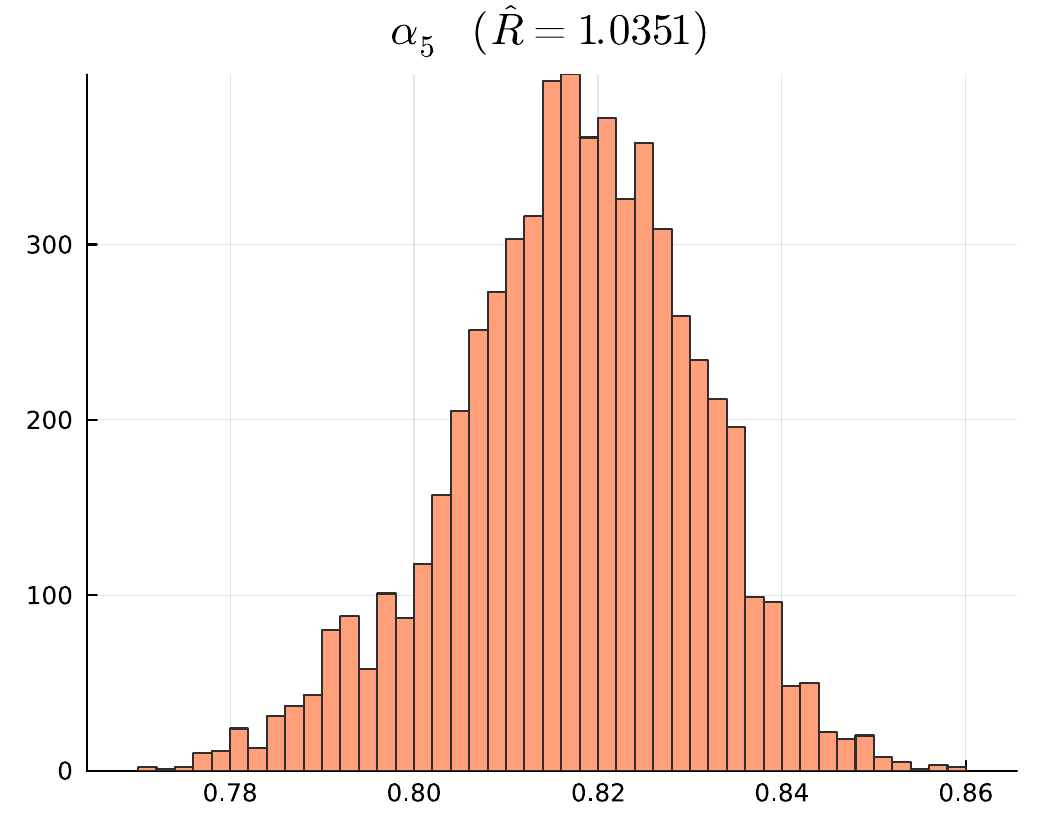}
    \includegraphics[width=0.19\linewidth]{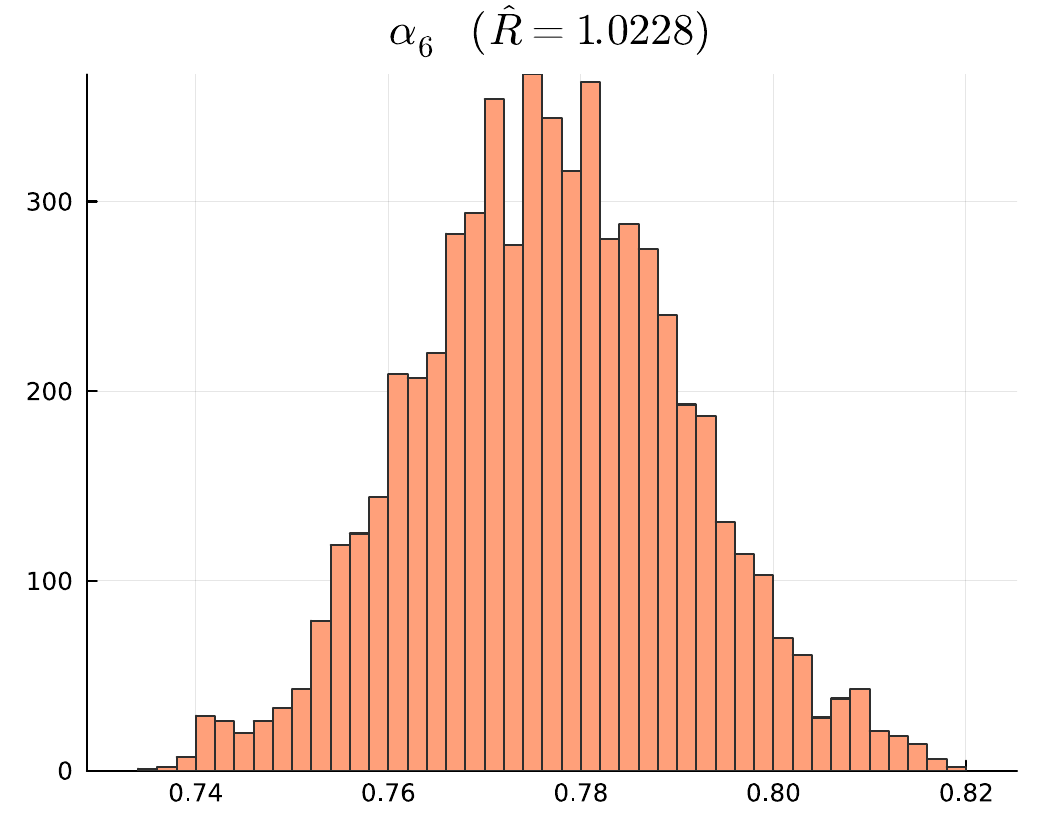}
    \includegraphics[width=0.19\linewidth]{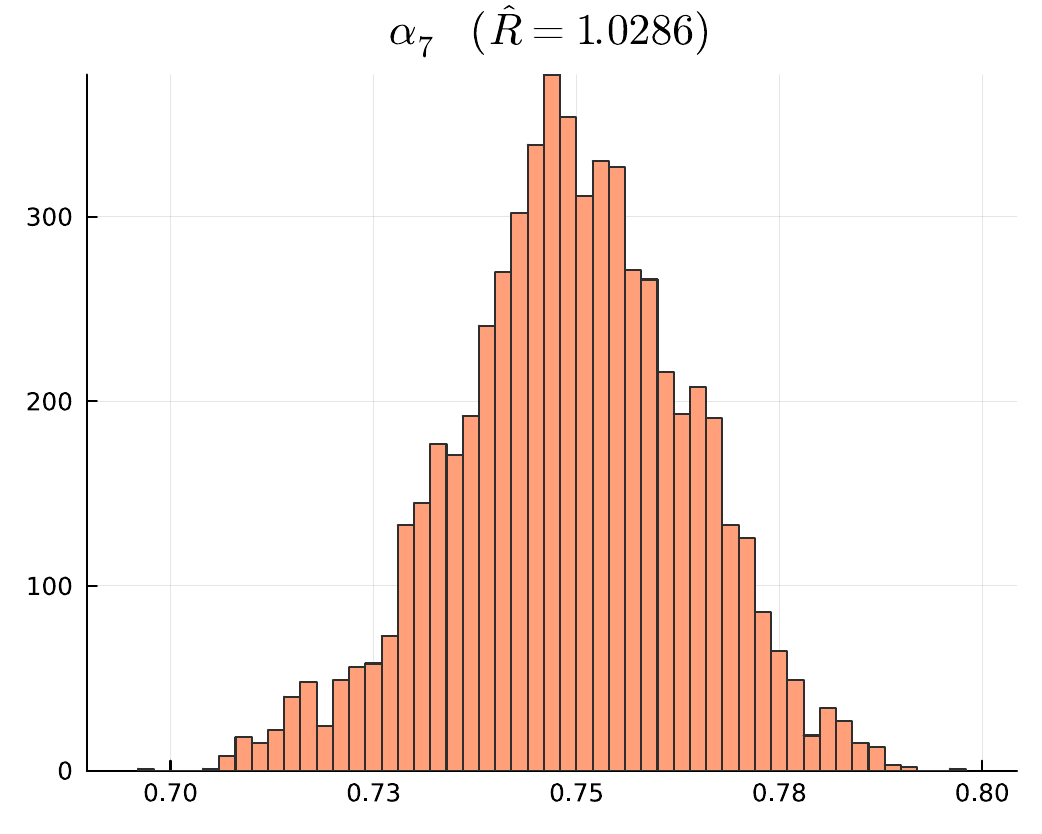}
    \includegraphics[width=0.19\linewidth]{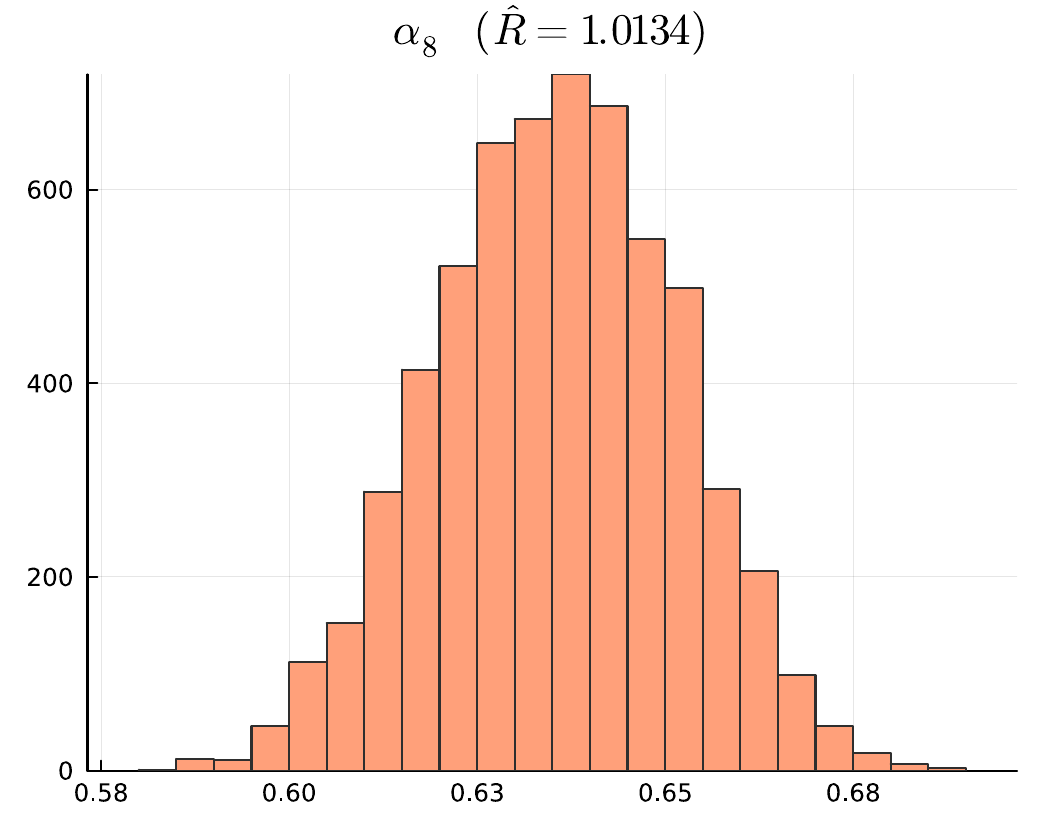}
    \includegraphics[width=0.19\linewidth]{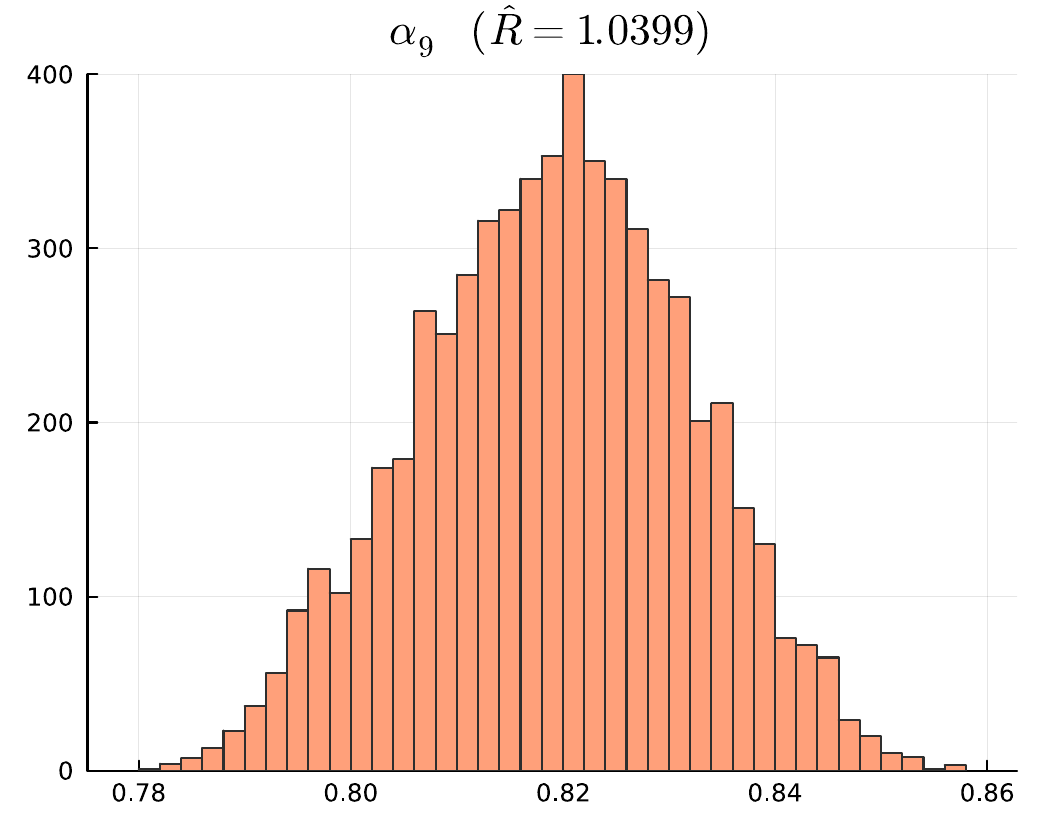}
    \includegraphics[width=0.19\linewidth]{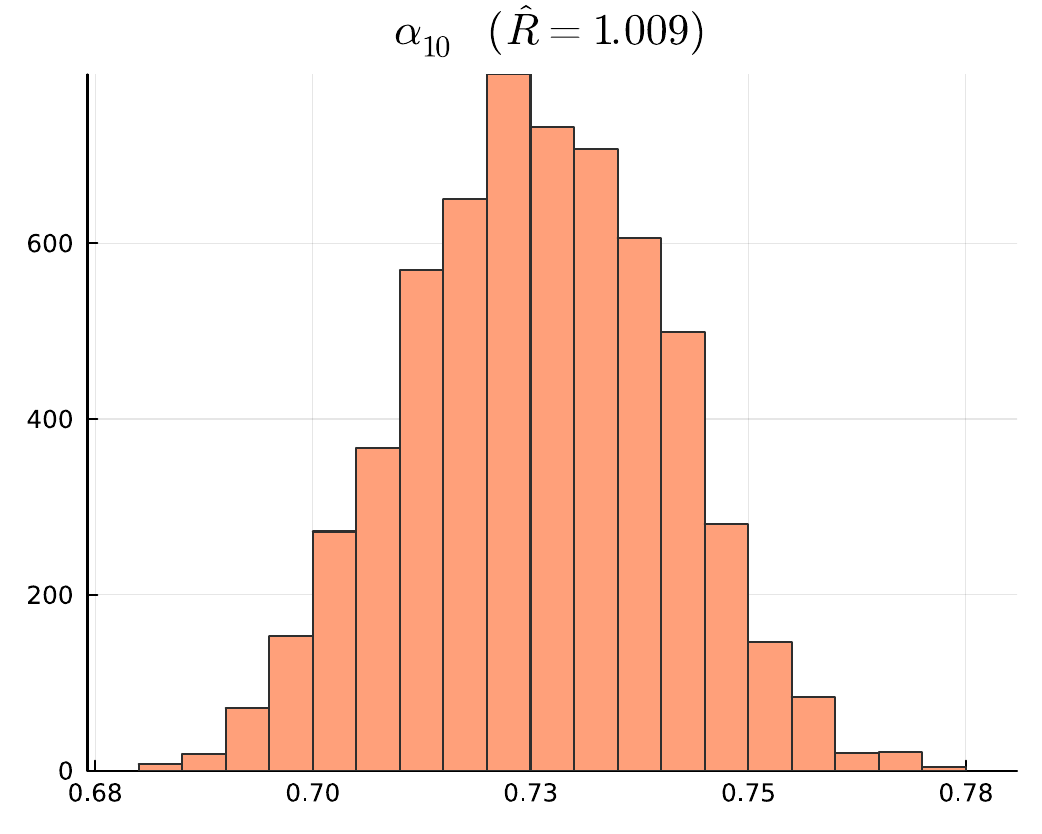}
    \includegraphics[width=0.19\linewidth]{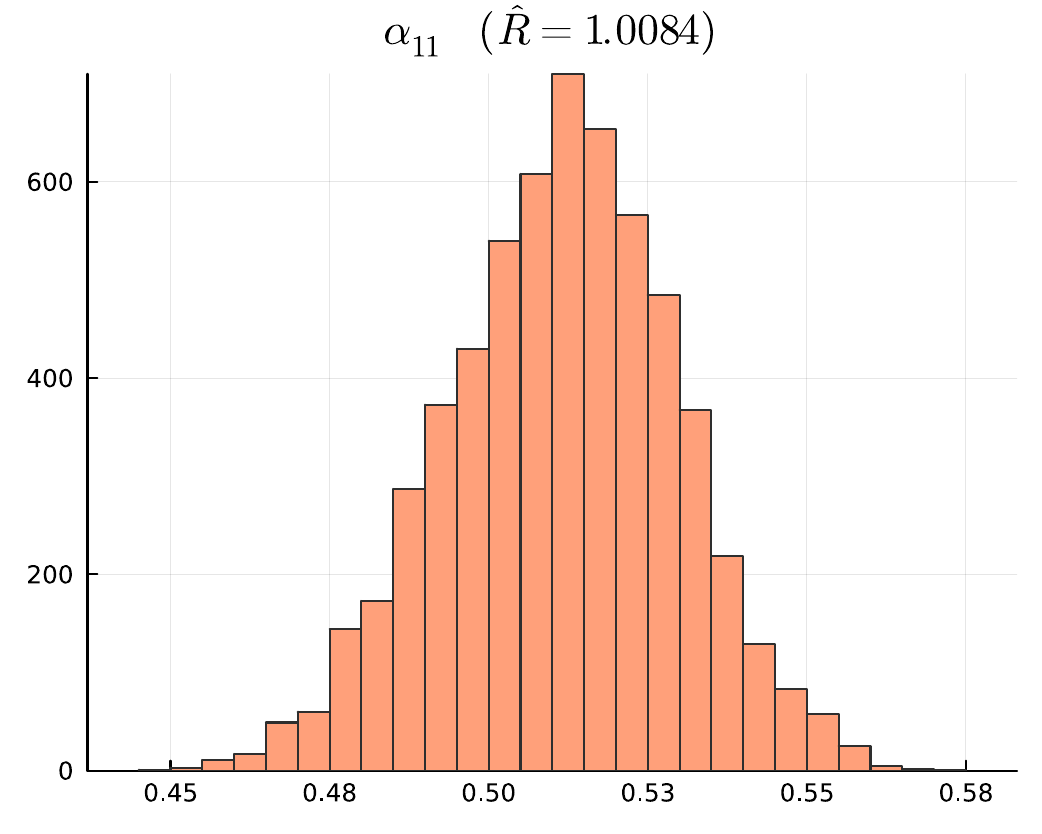}
    \includegraphics[width=0.19\linewidth]{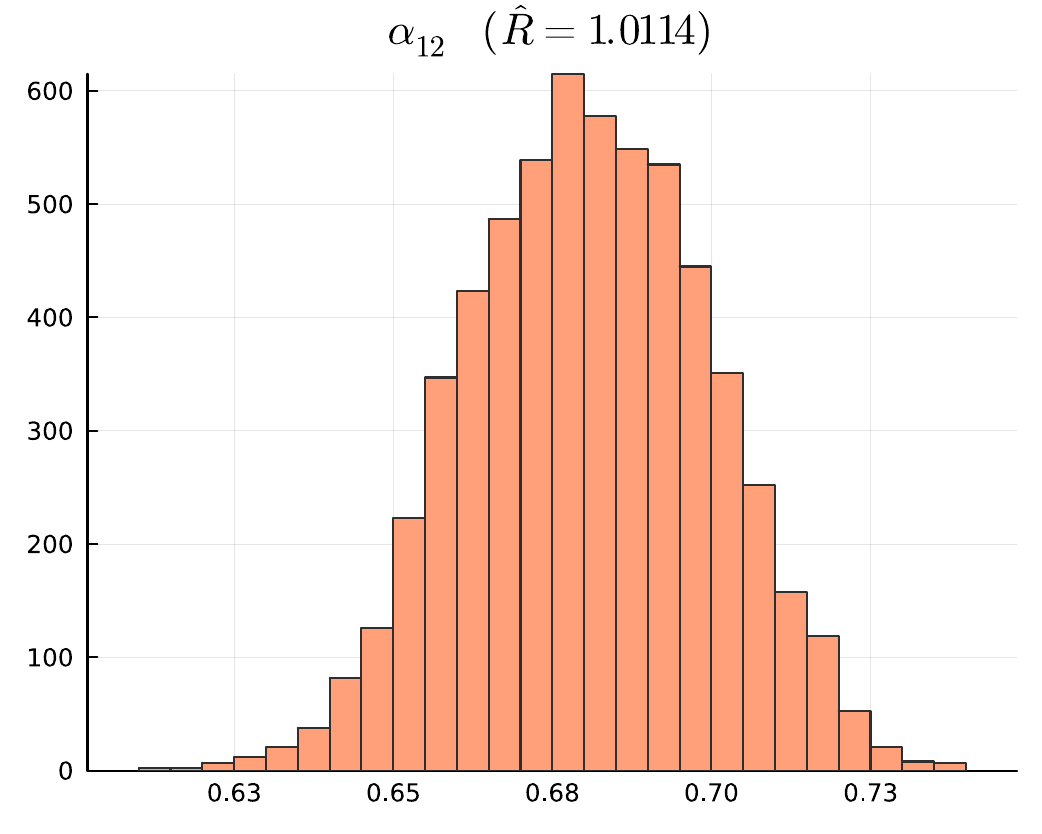}\\
    \includegraphics[width=0.19\linewidth]{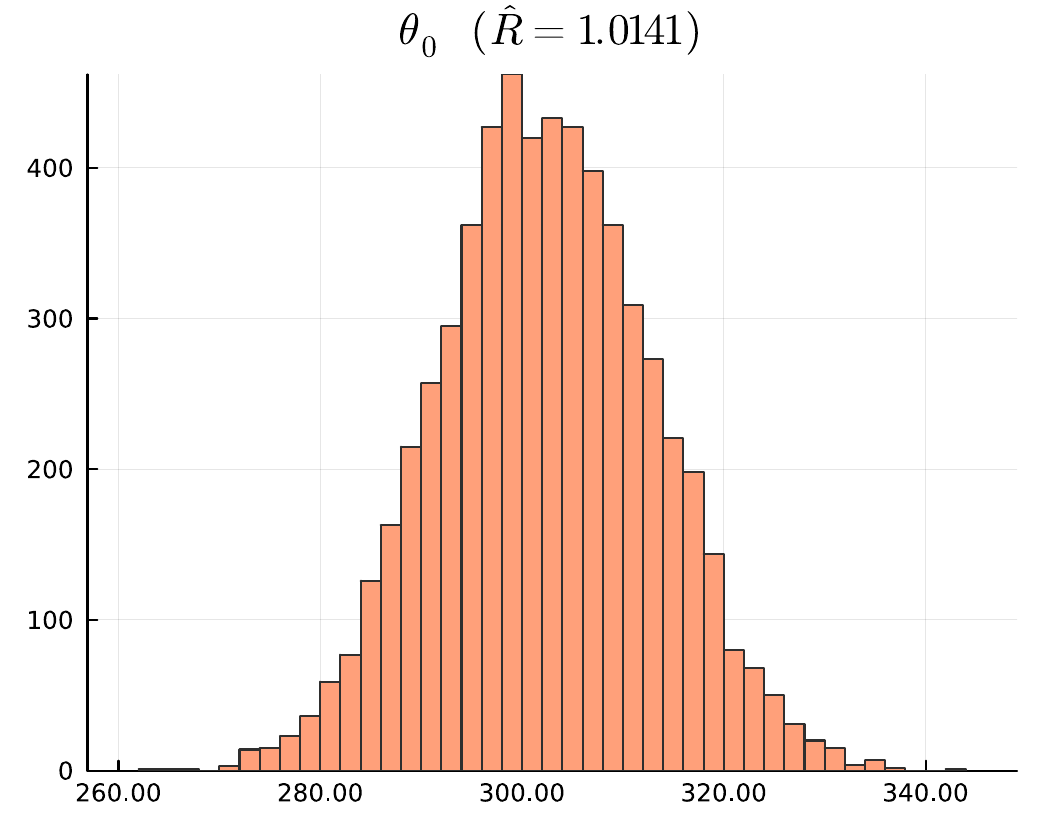}
    \includegraphics[width=0.19\linewidth]{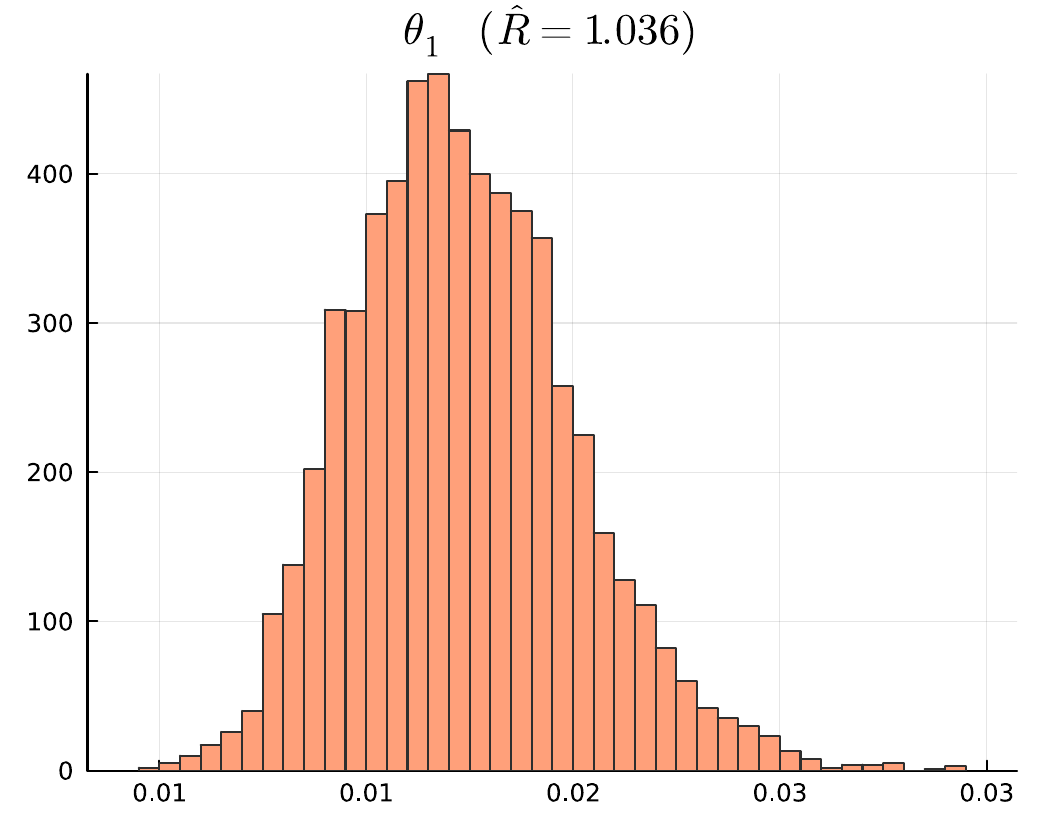}
    \includegraphics[width=0.19\linewidth]{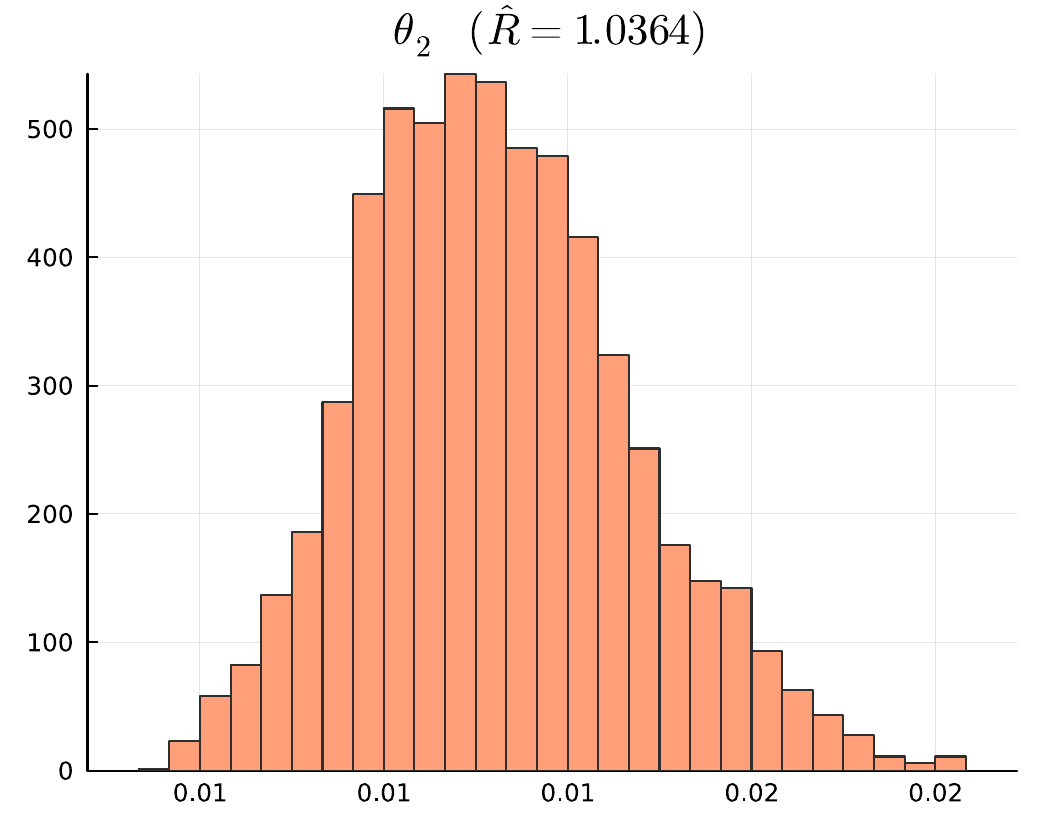}
    \includegraphics[width=0.19\linewidth]{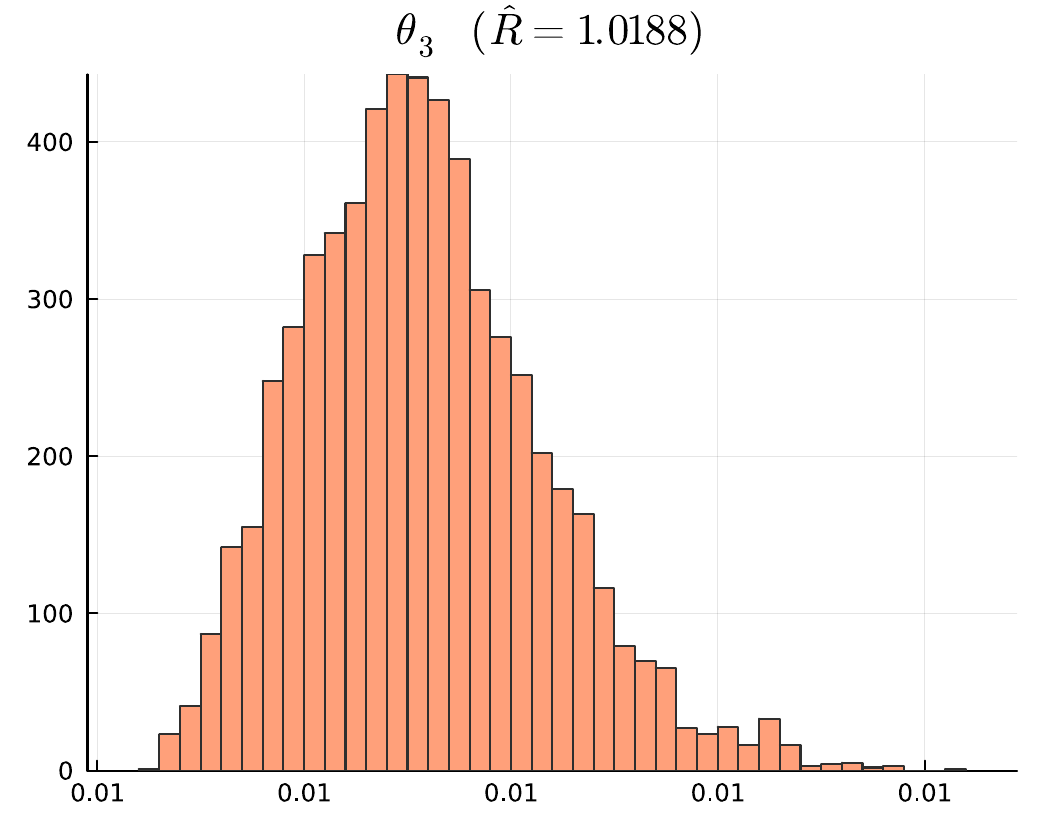}
    \includegraphics[width=0.19\linewidth]{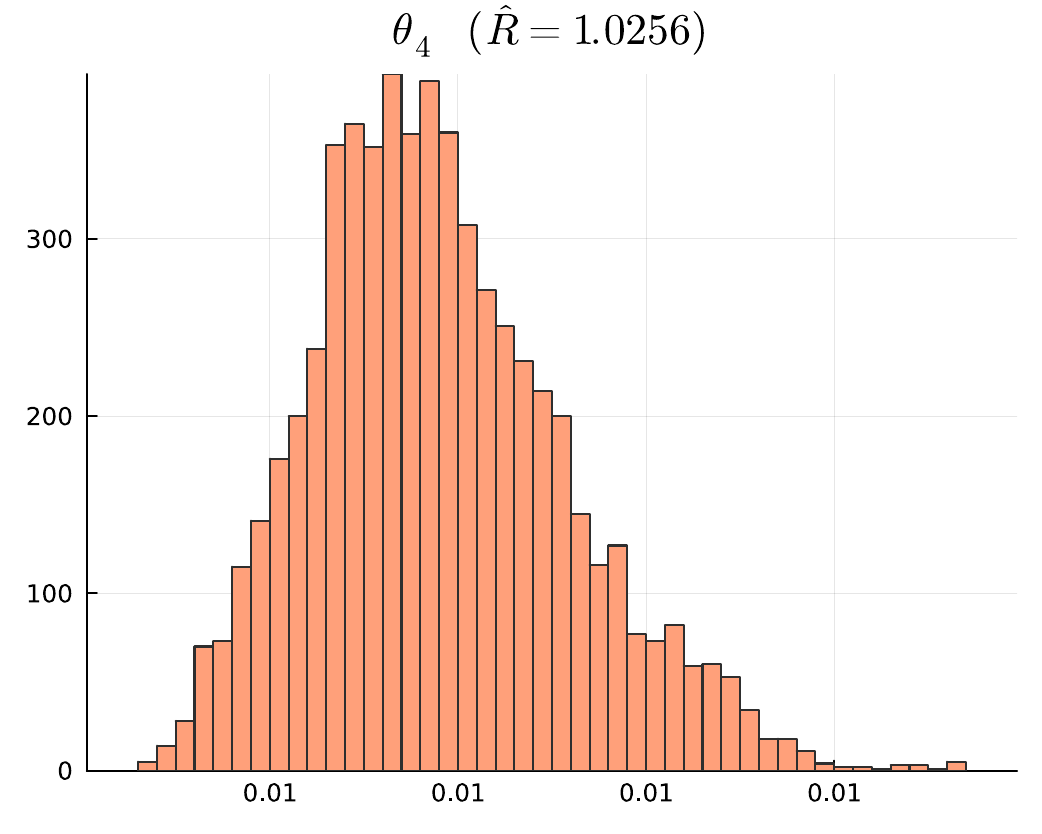}
    \includegraphics[width=0.19\linewidth]{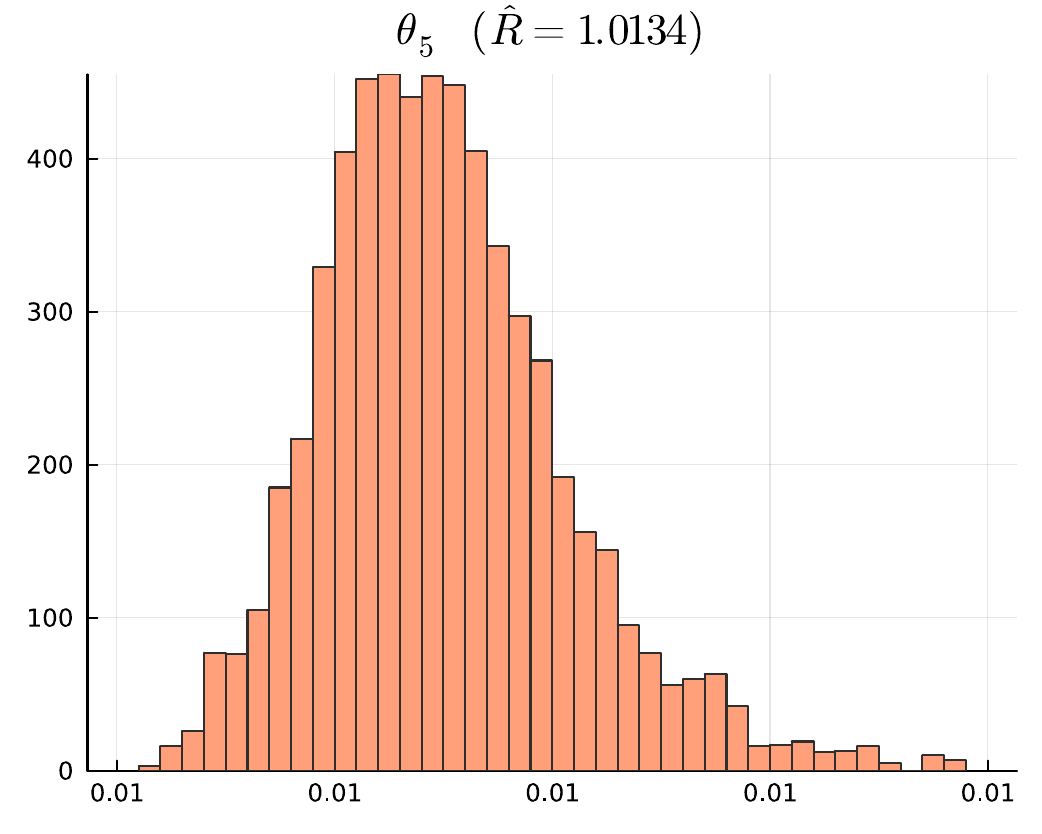}
    \includegraphics[width=0.19\linewidth]{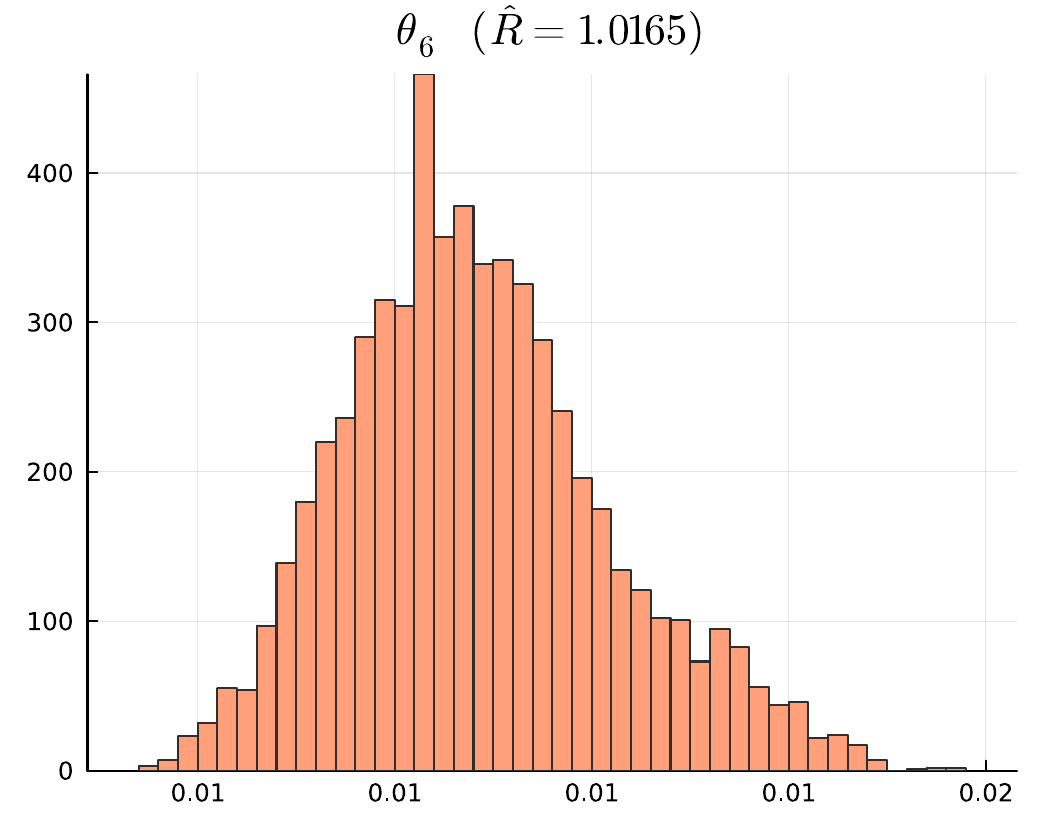}
    \includegraphics[width=0.19\linewidth]{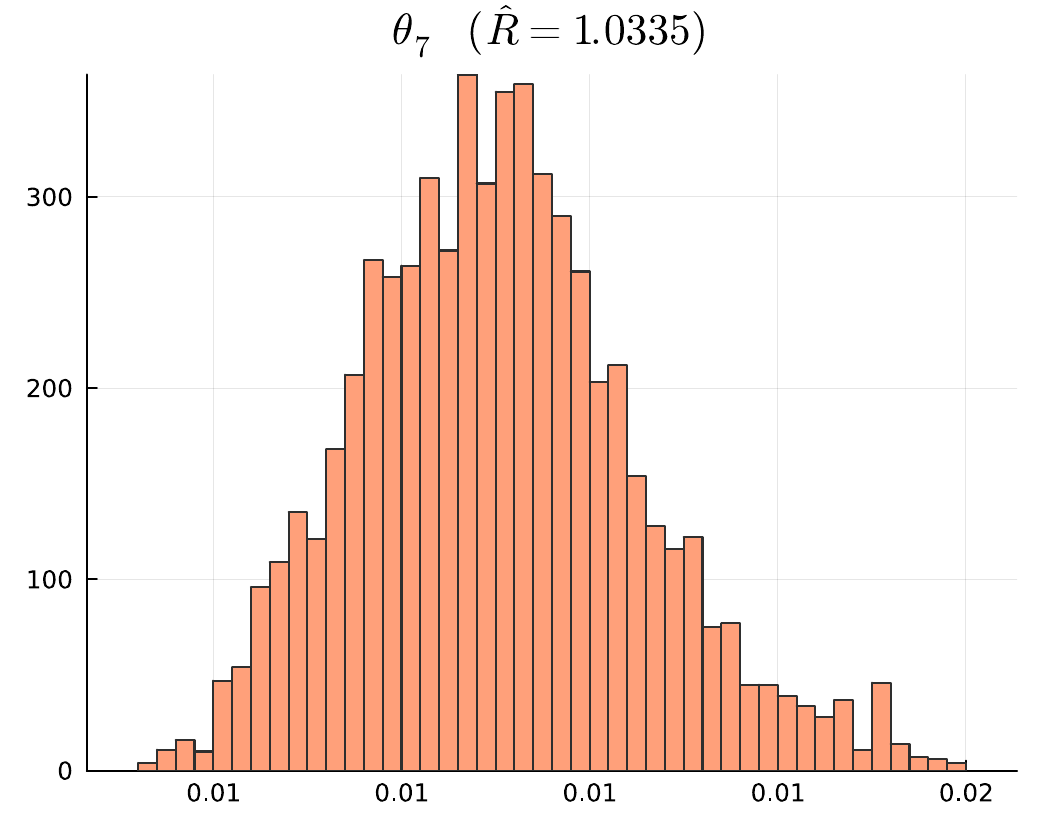}
    \includegraphics[width=0.19\linewidth]{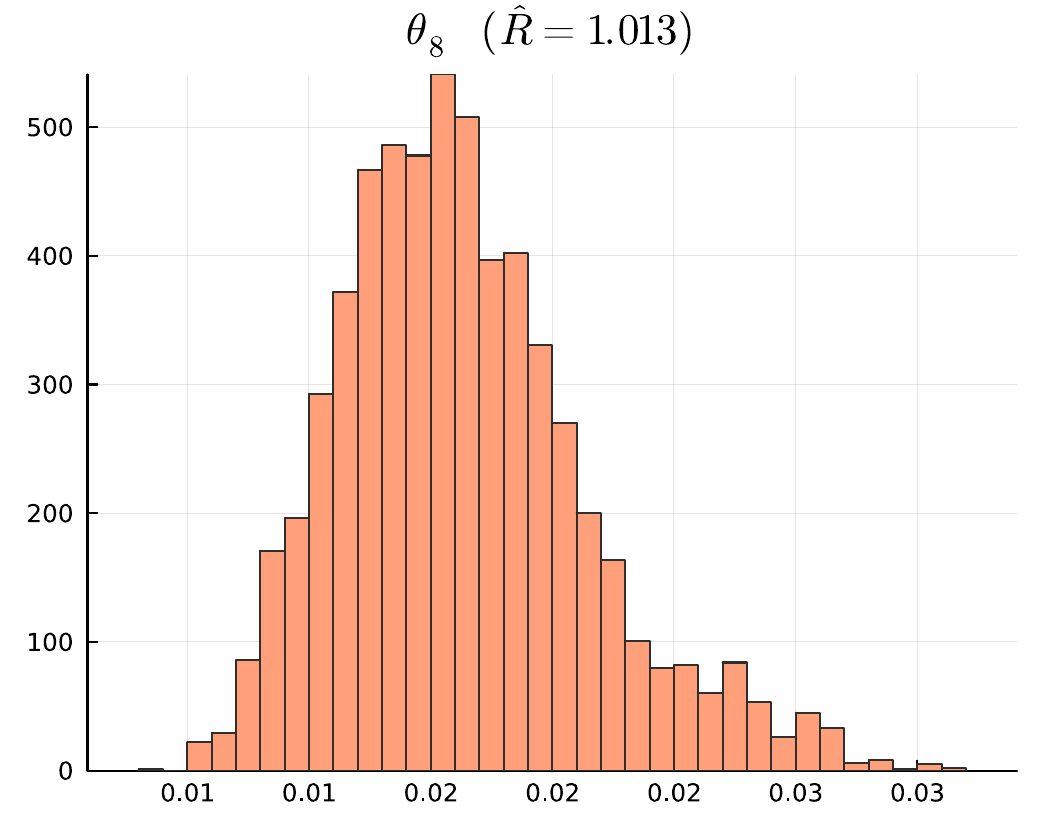}
    \includegraphics[width=0.19\linewidth]{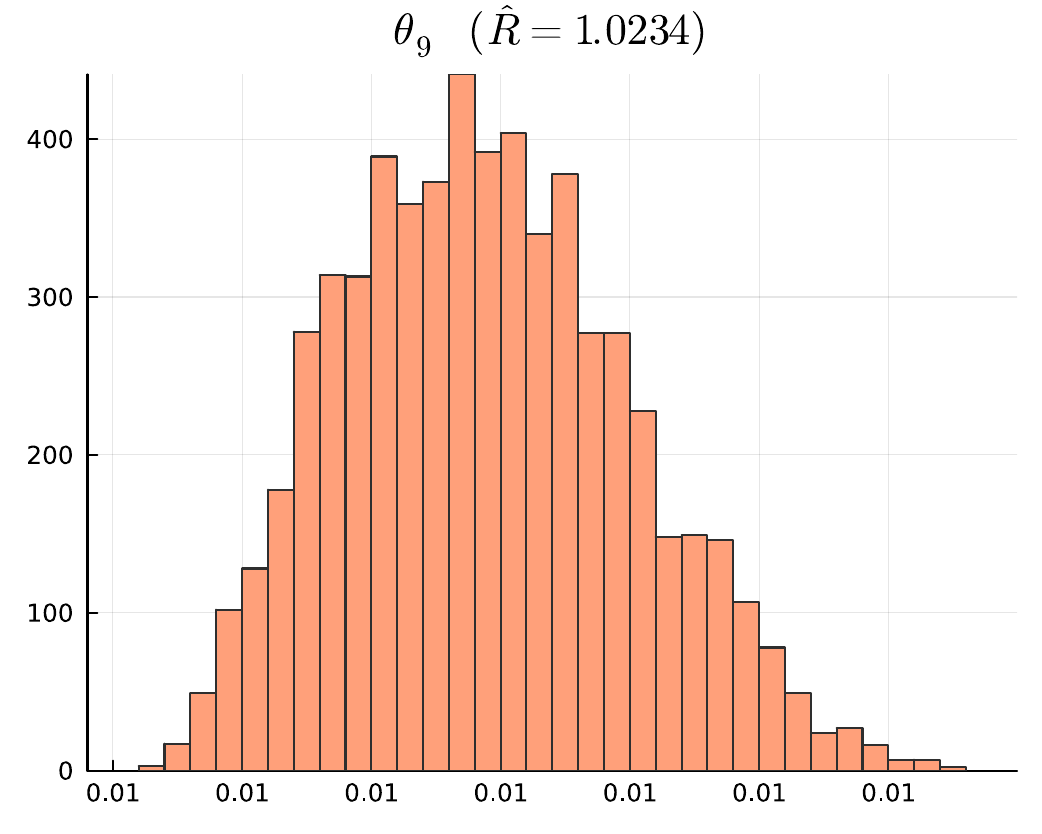}
    \includegraphics[width=0.19\linewidth]{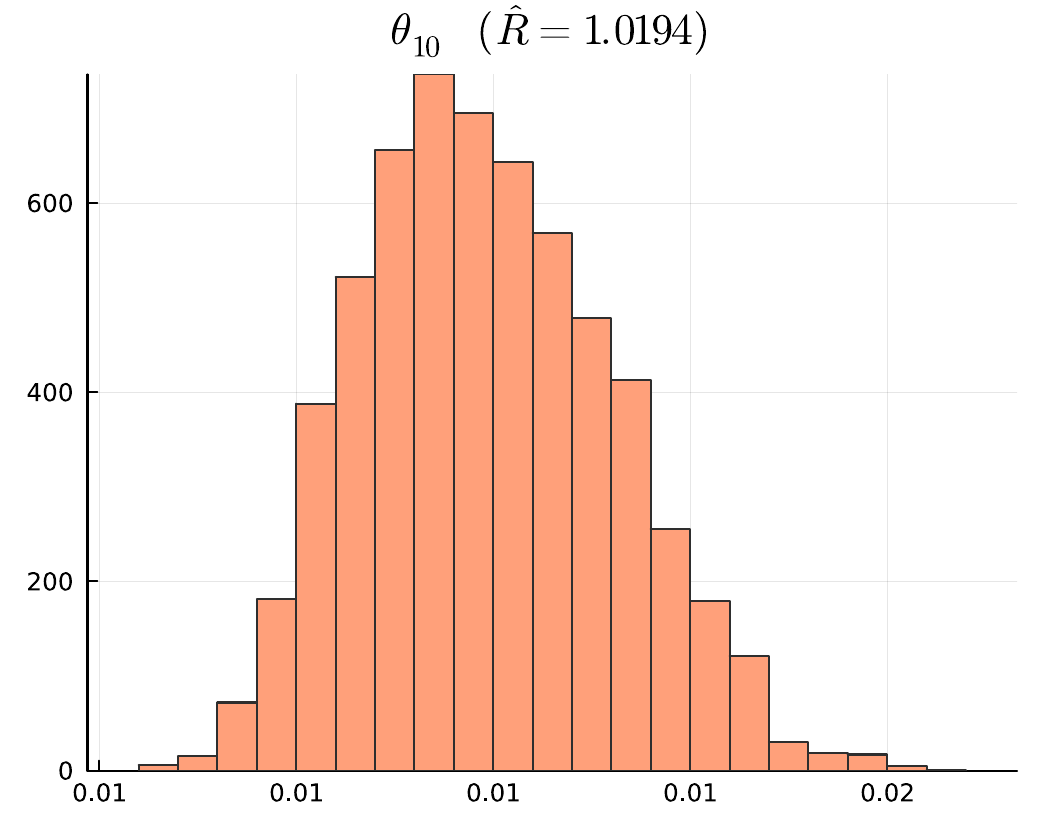}
    \includegraphics[width=0.19\linewidth]{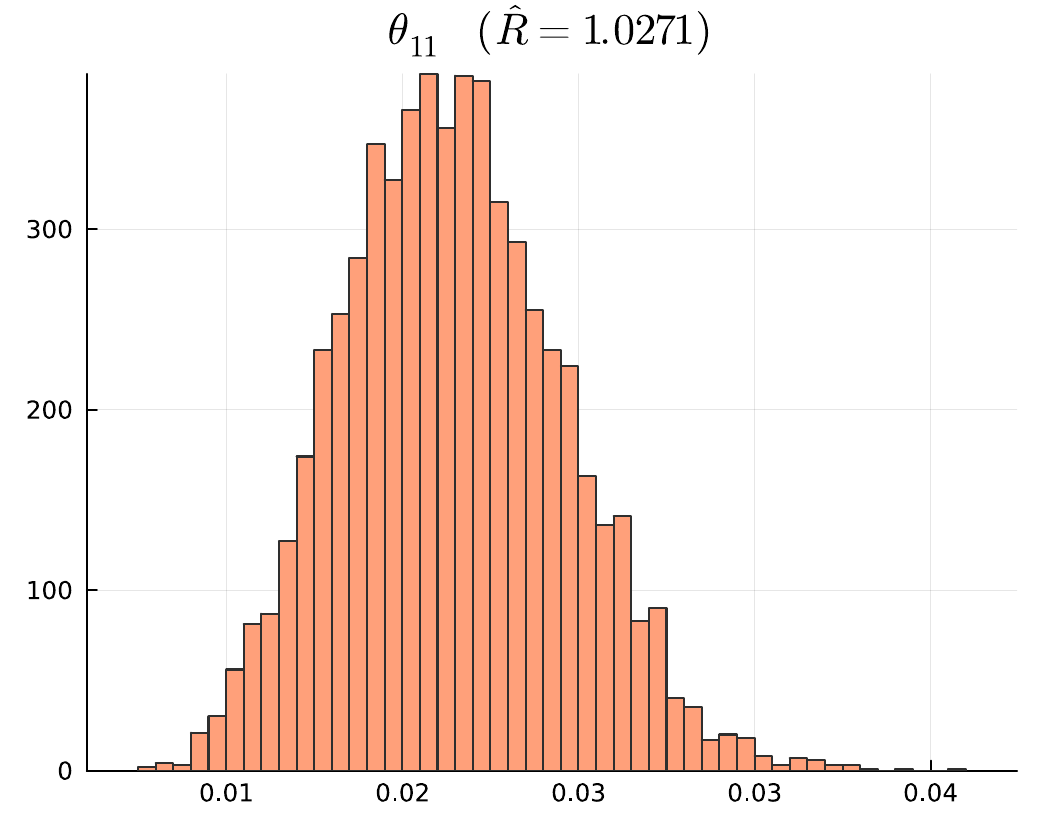}
    \includegraphics[width=0.19\linewidth]{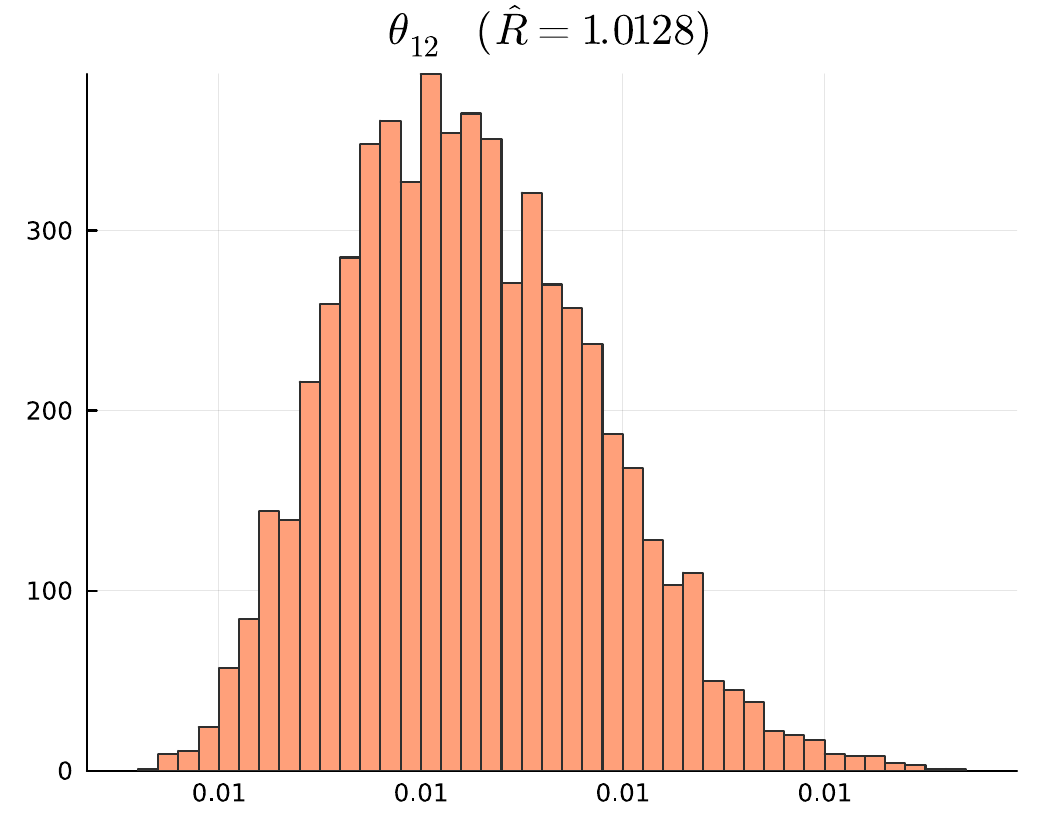}
    \caption{Posterior distributions of the GG HIBP model parameters inferred from the microbiome data, along with corresponding $\hat{R}$ statistics.}
    \label{fig:microbiome_posterior_gg}
\end{figure}

\begin{figure}
    \centering
    \includegraphics[width=0.19\linewidth]{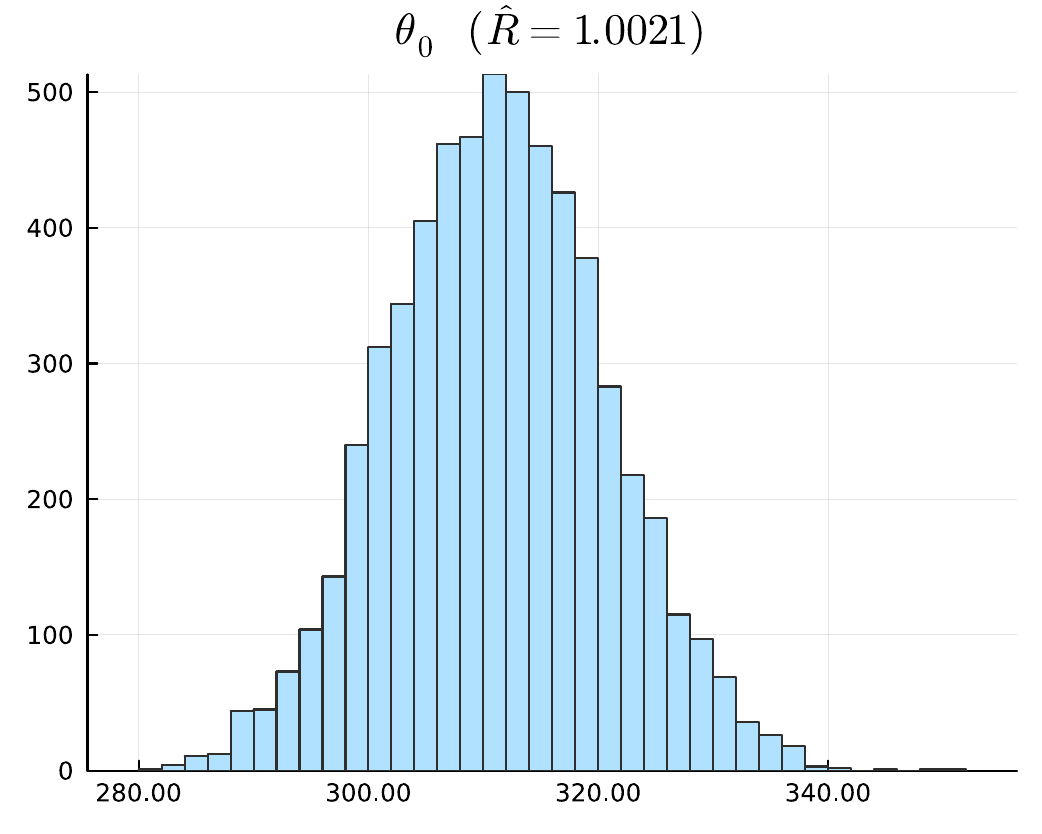}
    \includegraphics[width=0.19\linewidth]{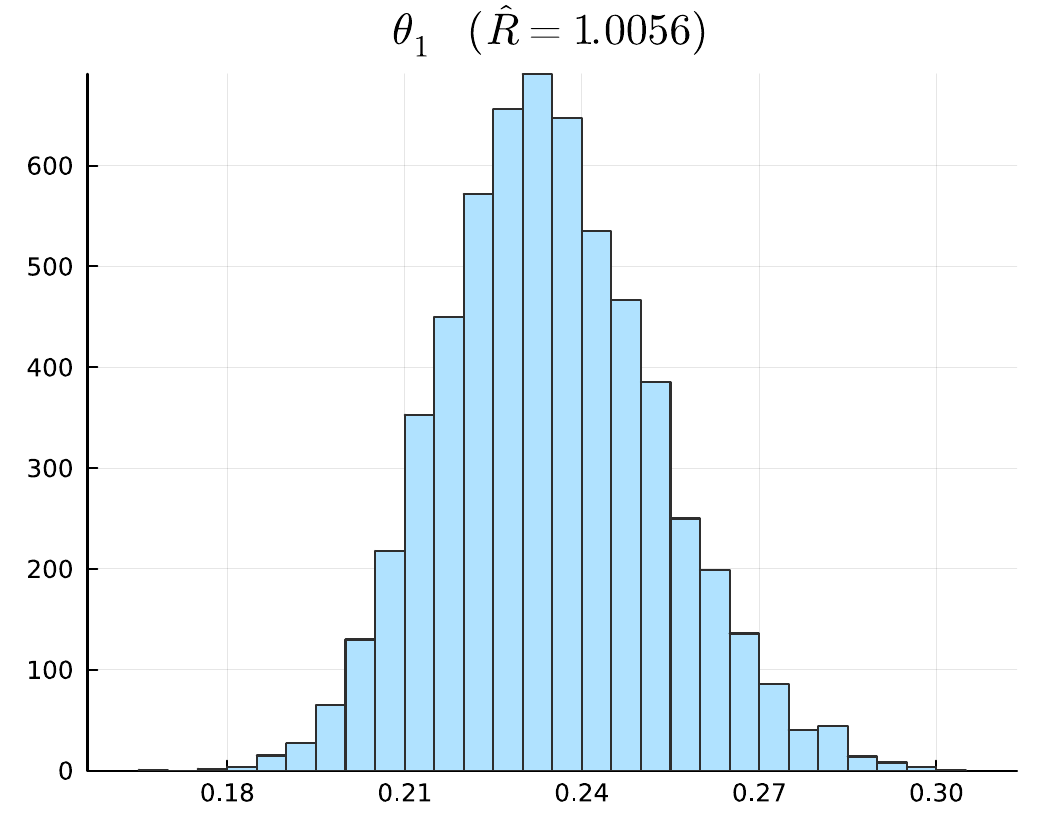}
    \includegraphics[width=0.19\linewidth]{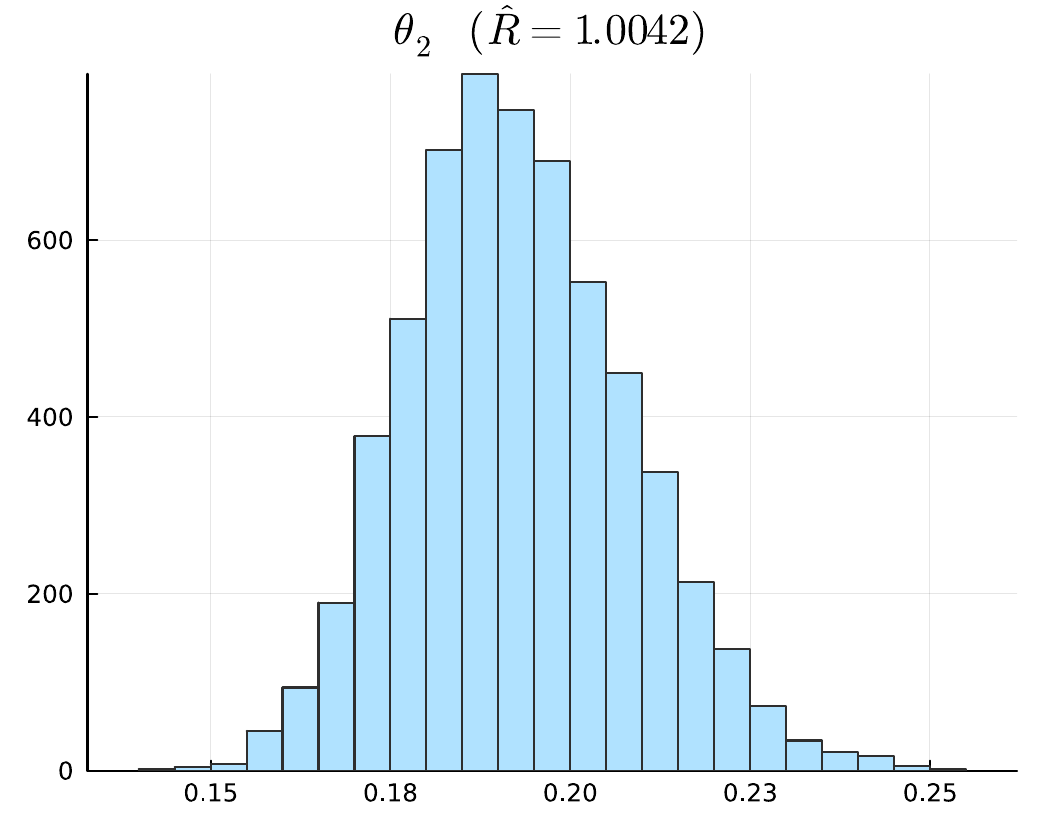}
    \includegraphics[width=0.19\linewidth]{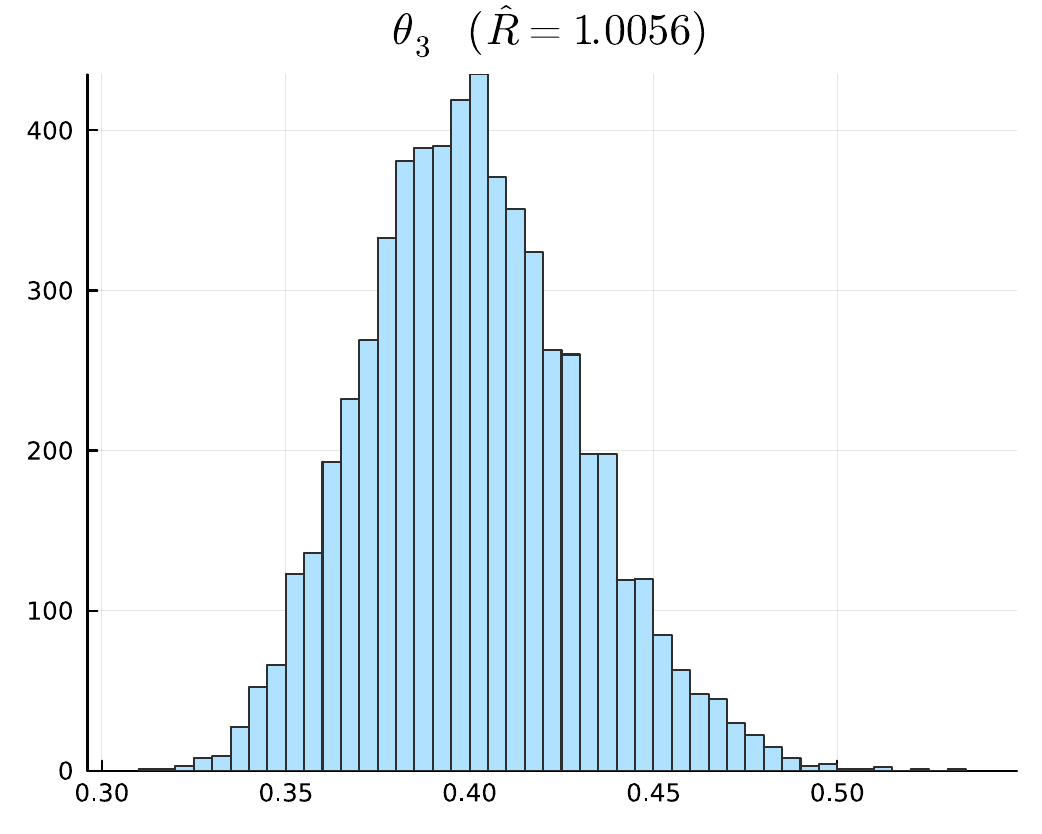}
    \includegraphics[width=0.19\linewidth]{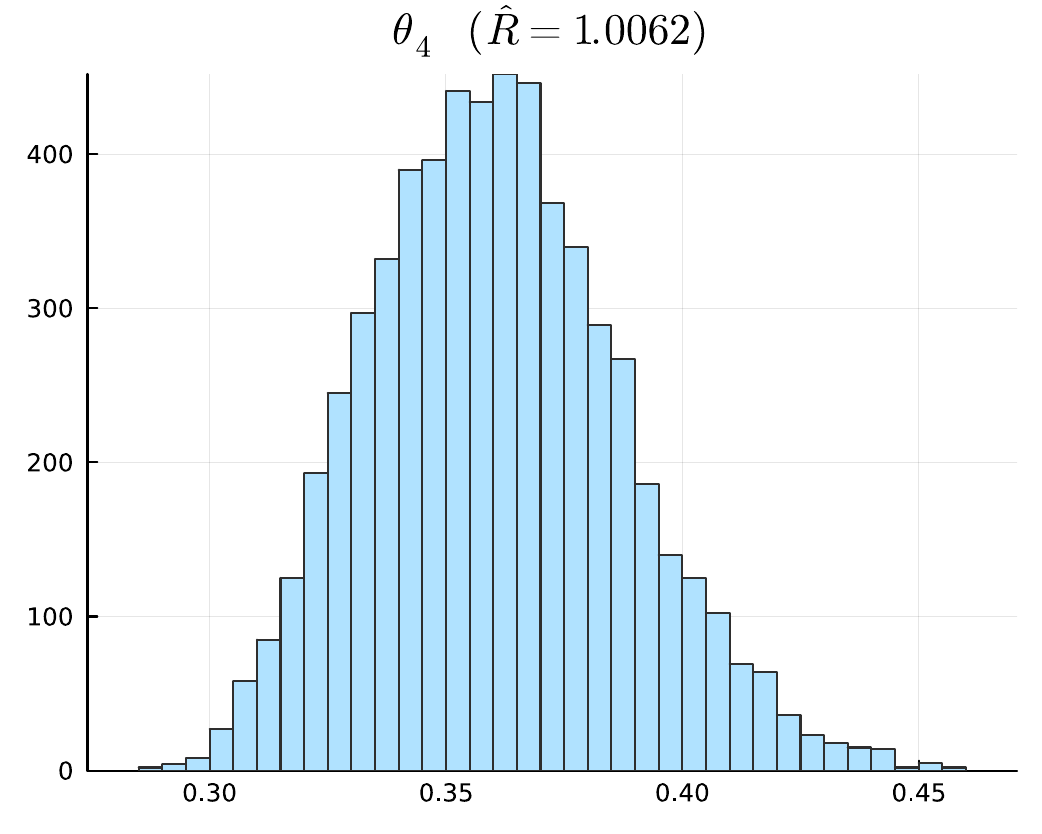}
    \includegraphics[width=0.19\linewidth]{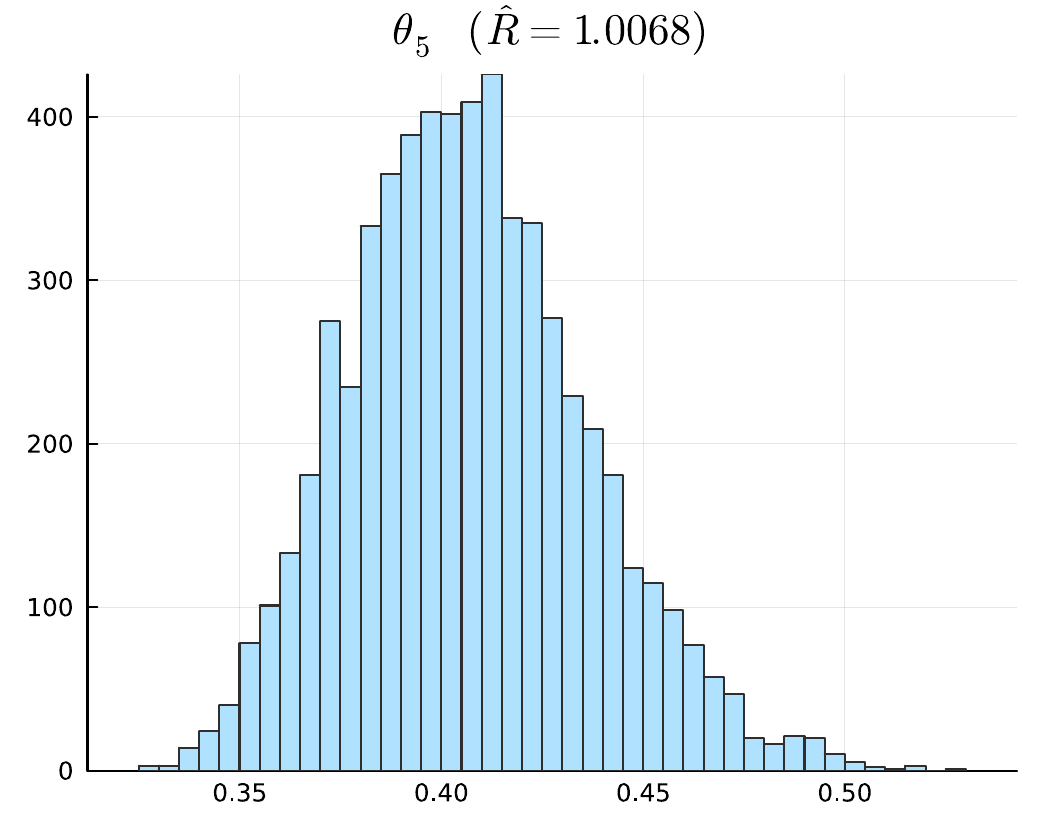}
    \includegraphics[width=0.19\linewidth]{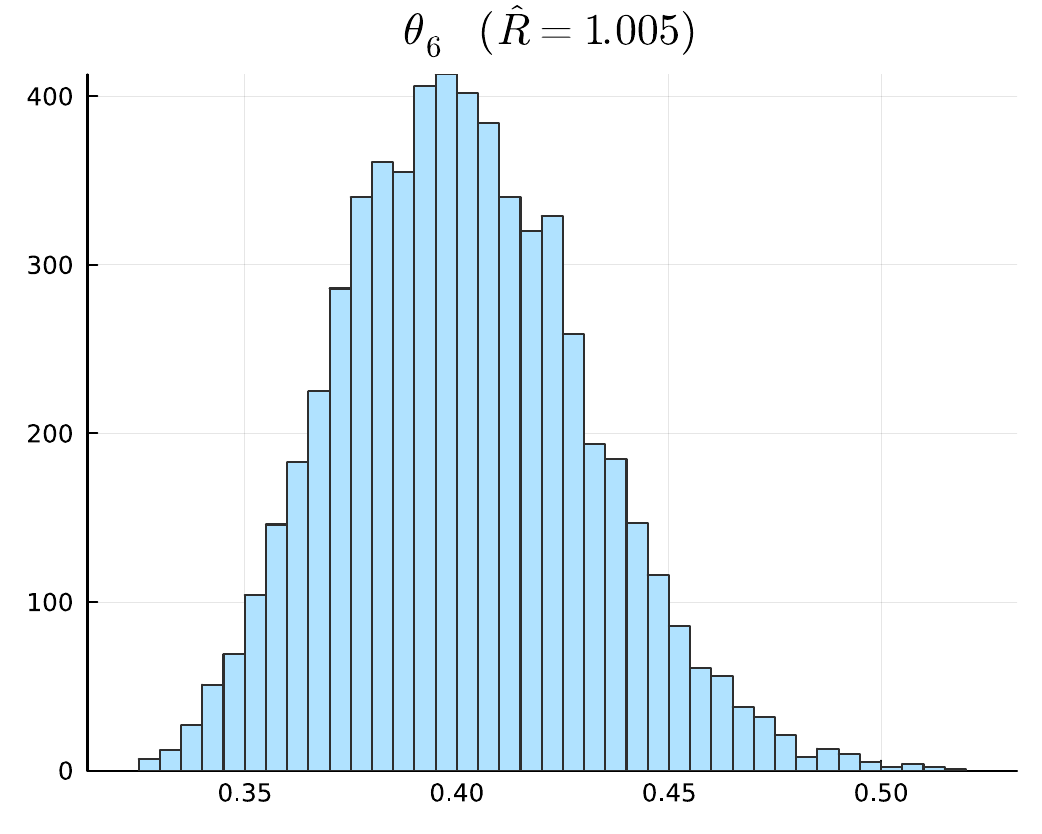}
    \includegraphics[width=0.19\linewidth]{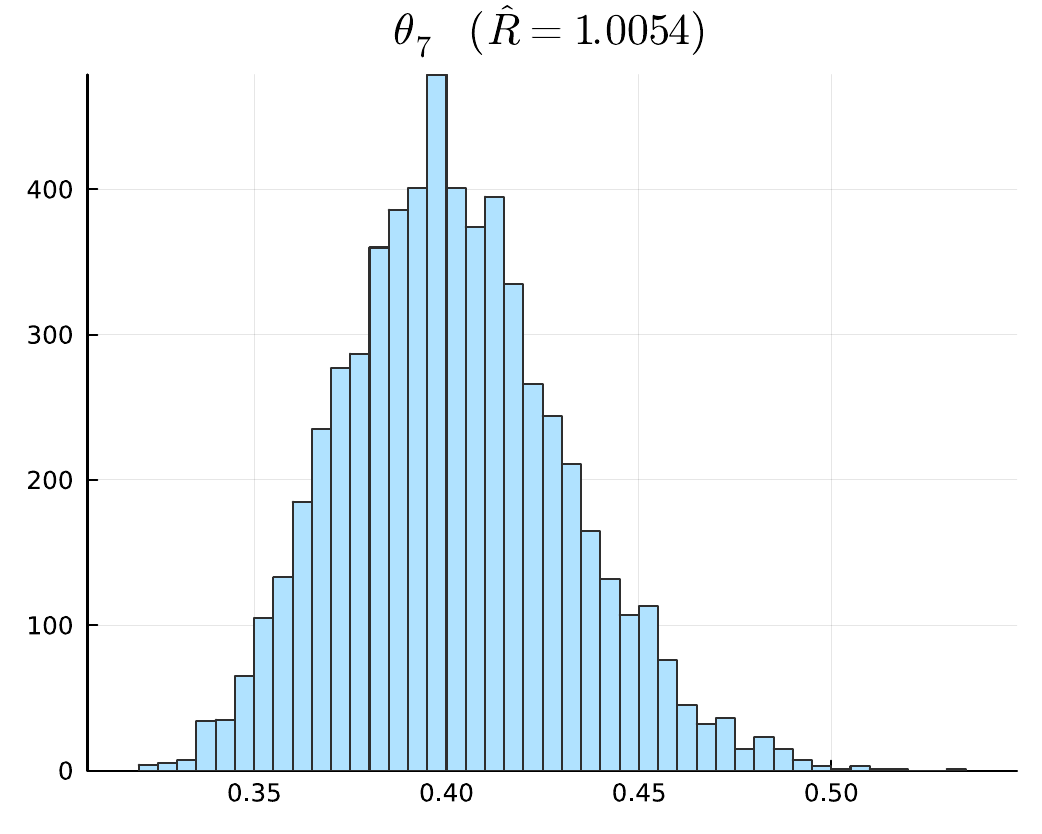}
    \includegraphics[width=0.19\linewidth]{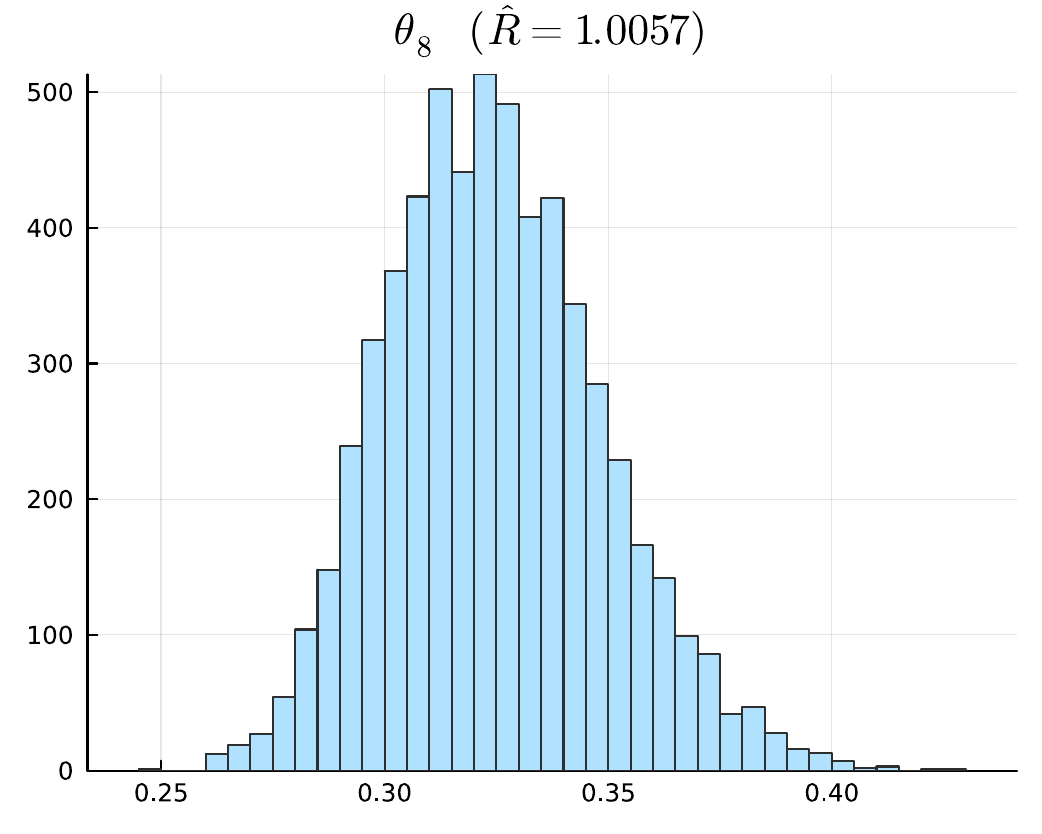}
    \includegraphics[width=0.19\linewidth]{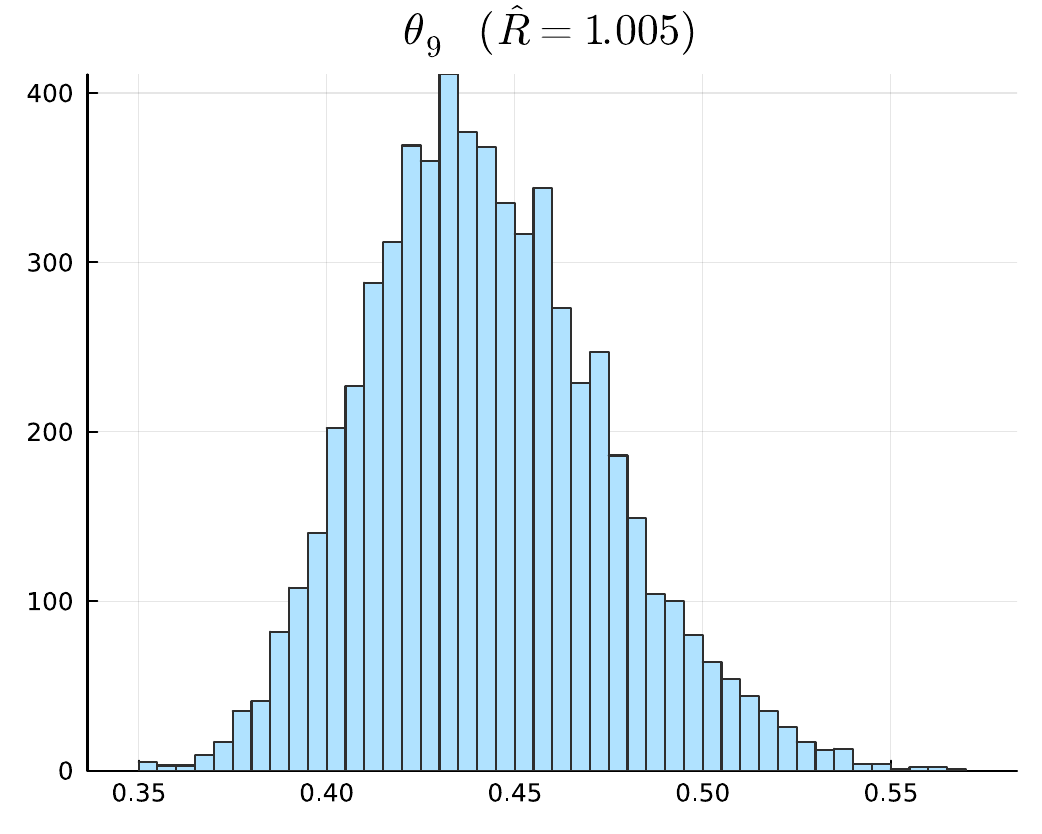}
    \includegraphics[width=0.19\linewidth]{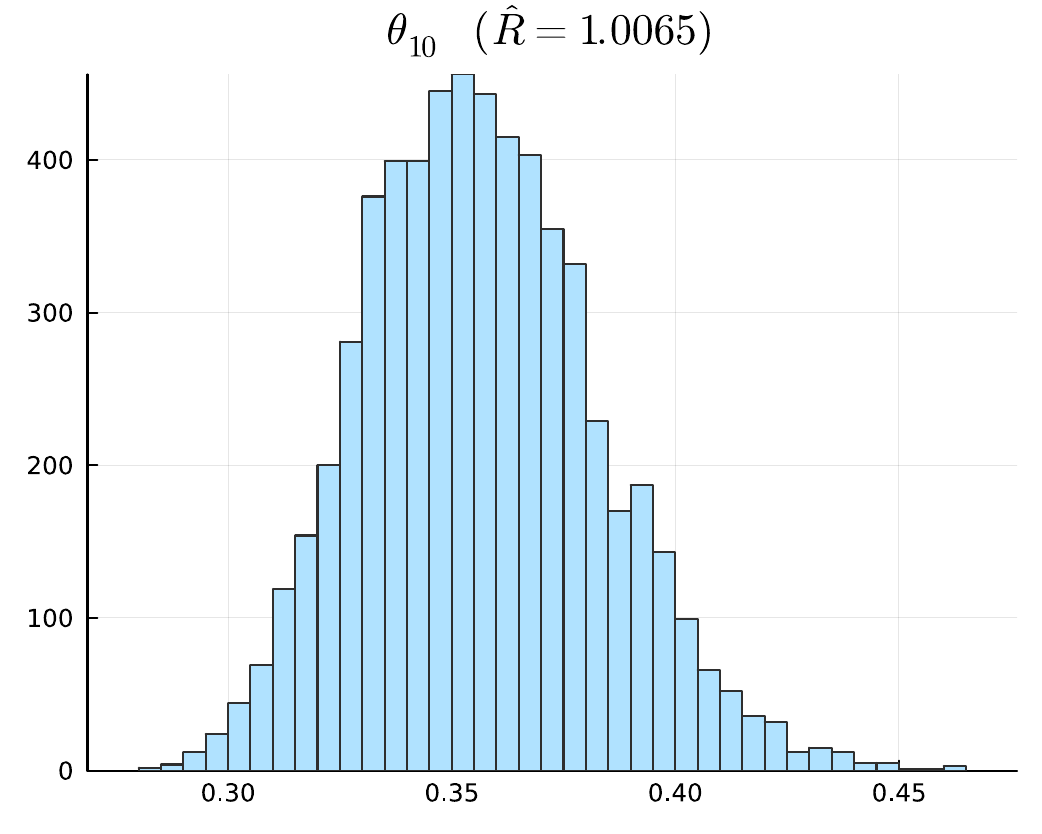}
    \includegraphics[width=0.19\linewidth]{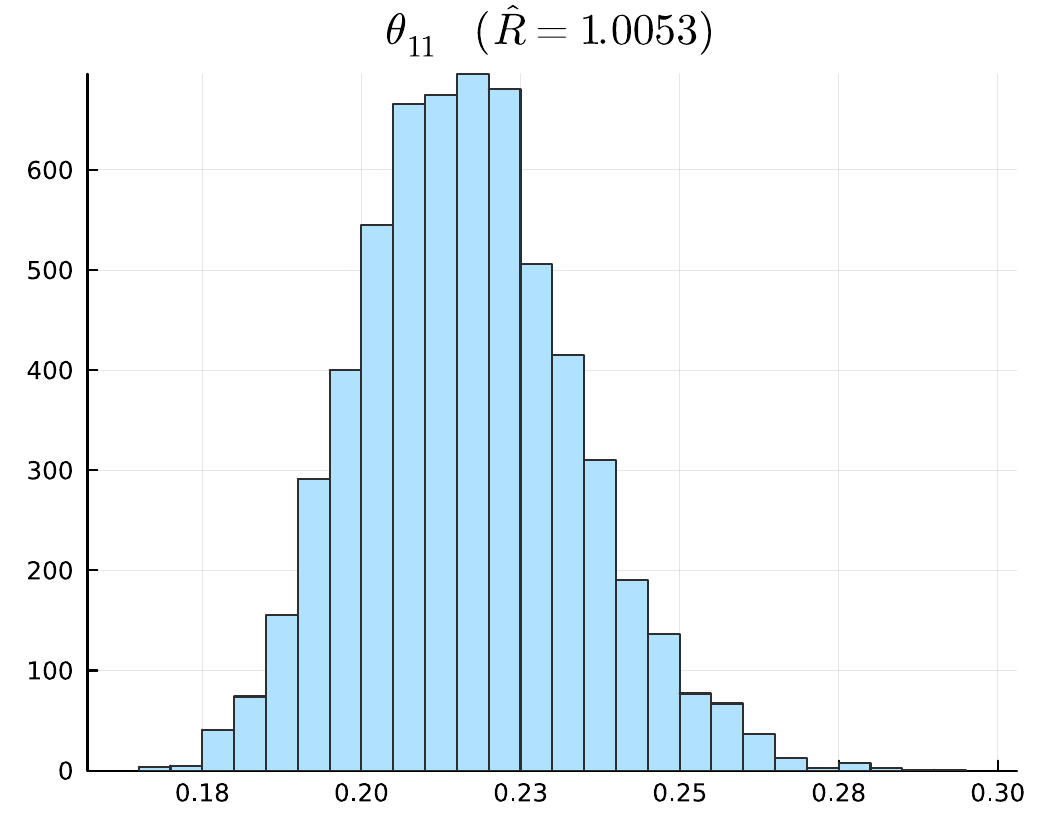}
    \includegraphics[width=0.19\linewidth]{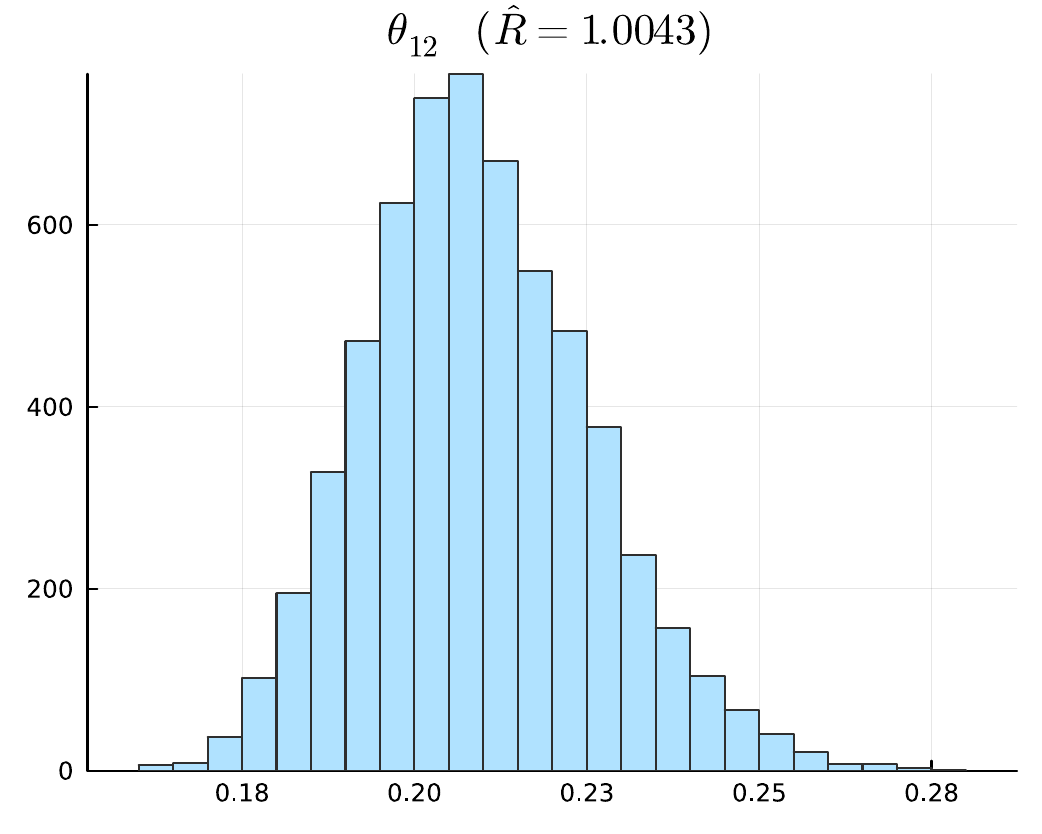}
    \caption{Posterior distributions of the Gamma HIBP model parameters inferred from the microbiome data, along with corresponding $\hat{R}$ statistics.}
        \label{fig:microbiome_posterior_gamma}
\end{figure}

\begin{figure}
    \centering
    \includegraphics[width=0.24\linewidth]{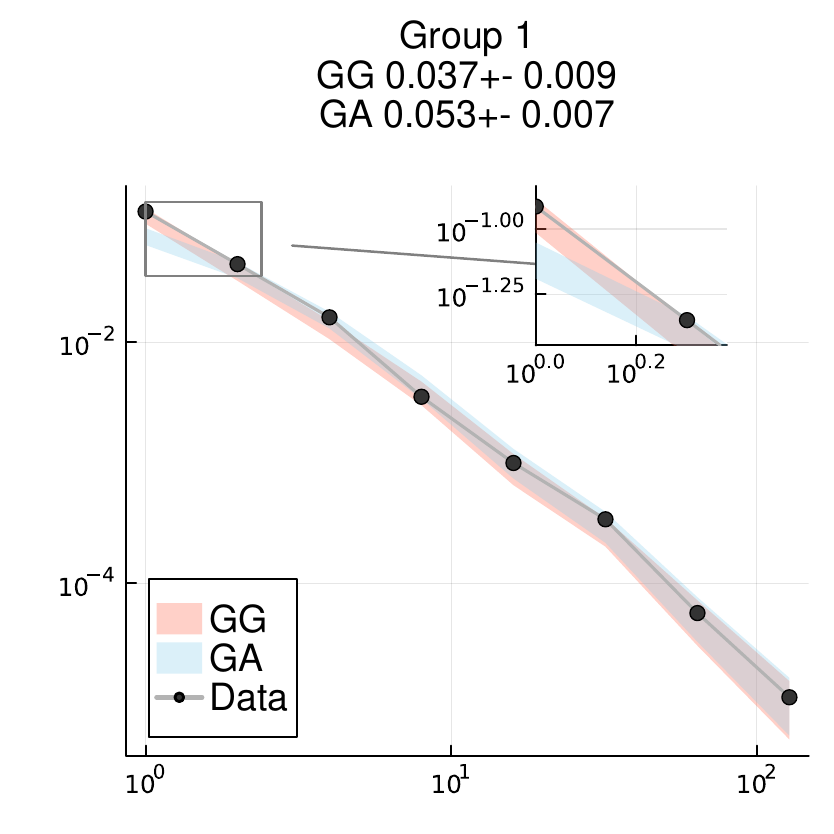}
    \includegraphics[width=0.24\linewidth]{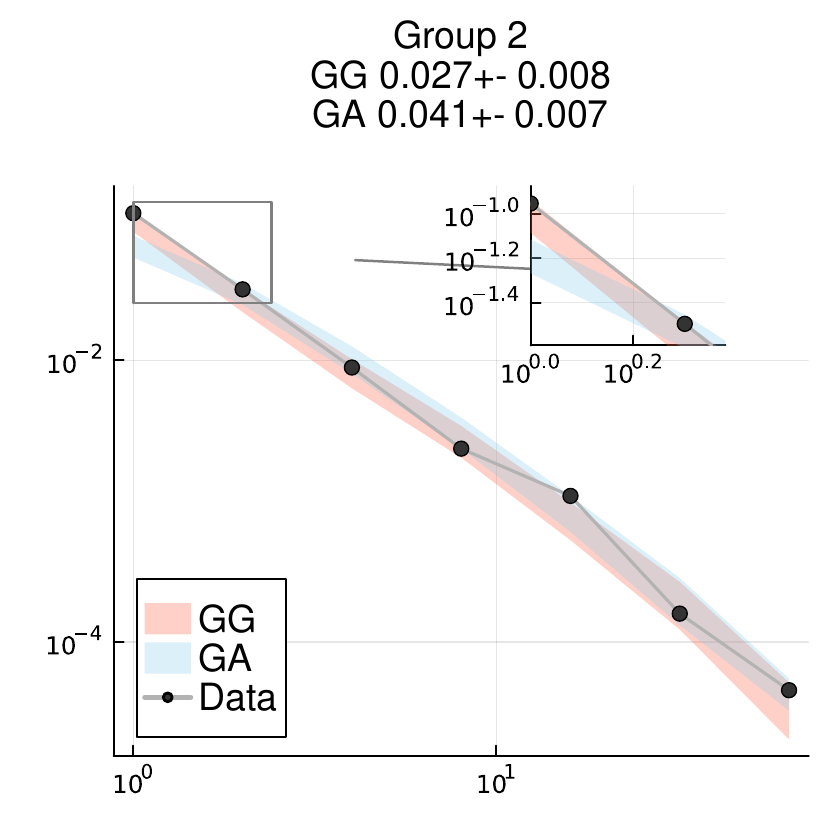}
    \includegraphics[width=0.24\linewidth]{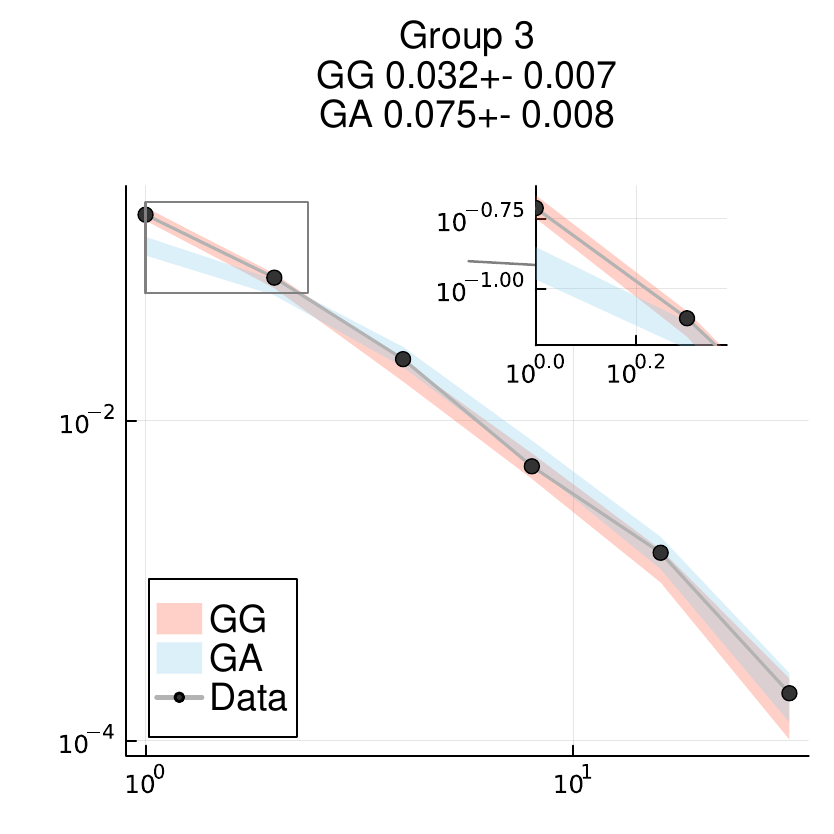}
    \includegraphics[width=0.24\linewidth]{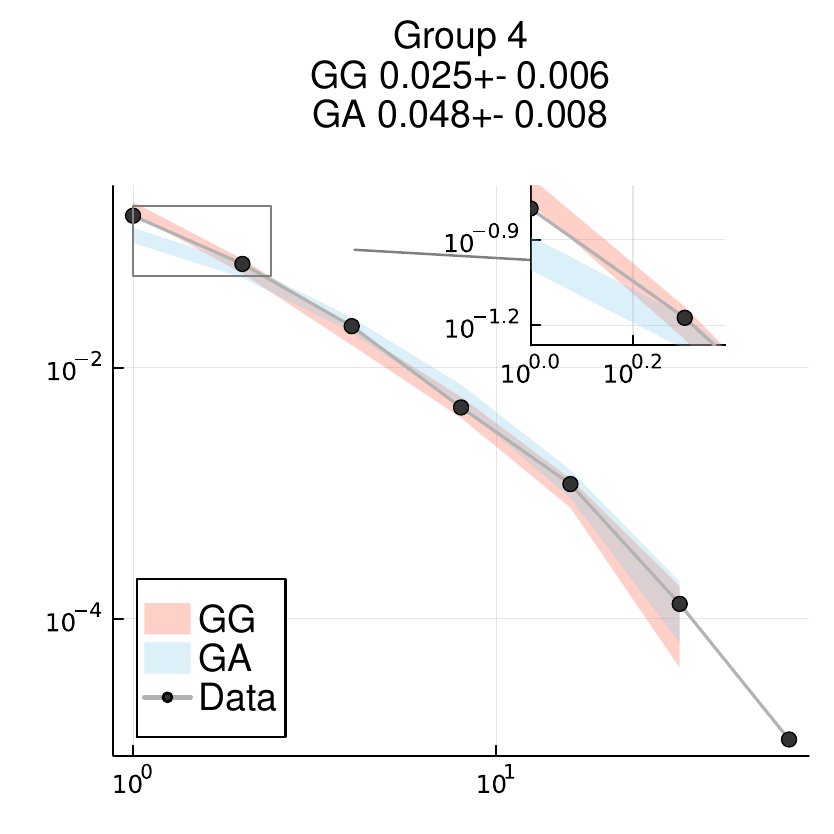}
    \includegraphics[width=0.24\linewidth]{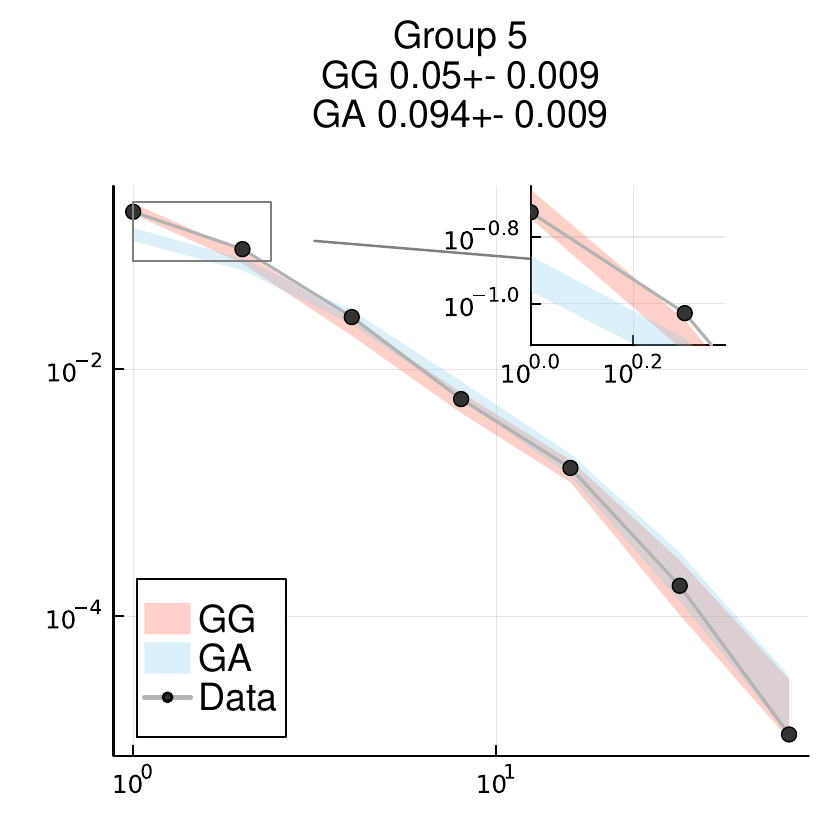}
    \includegraphics[width=0.24\linewidth]{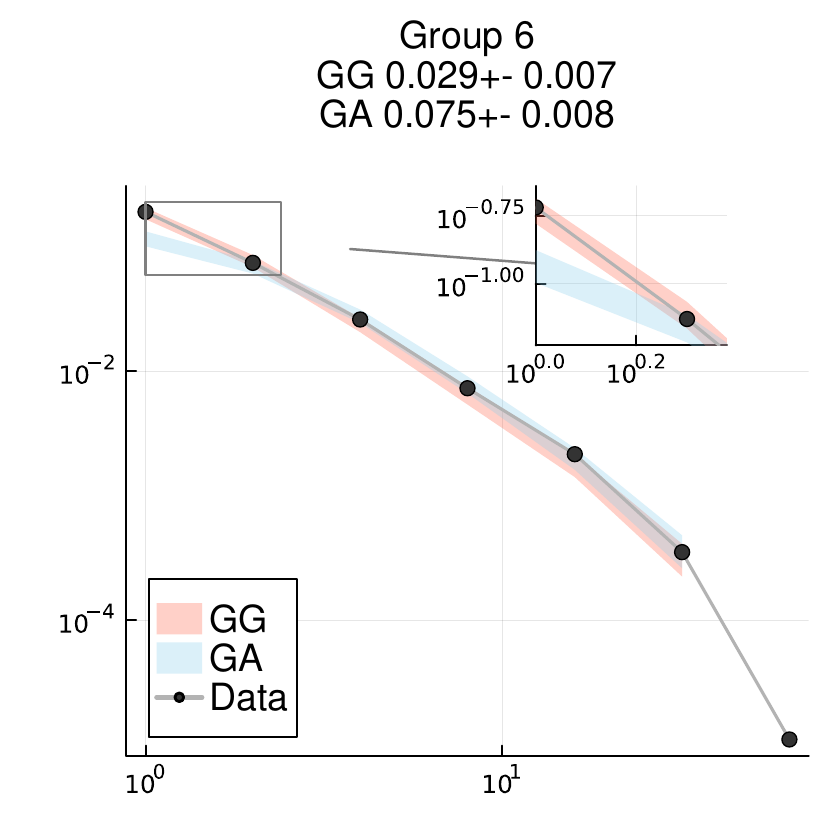}
    \includegraphics[width=0.24\linewidth]{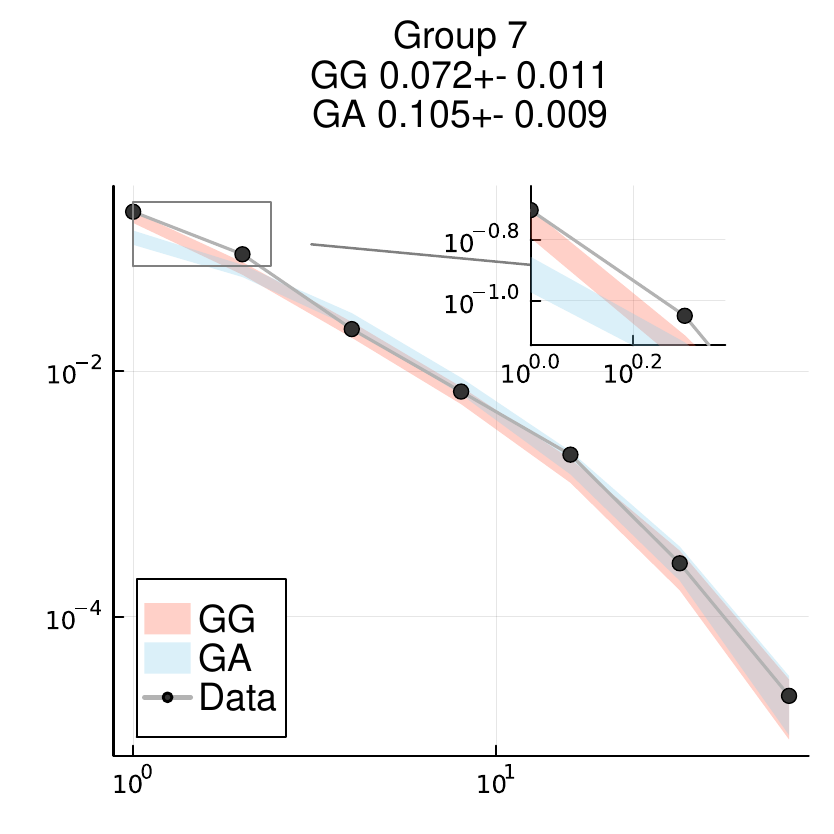}
    \includegraphics[width=0.24\linewidth]{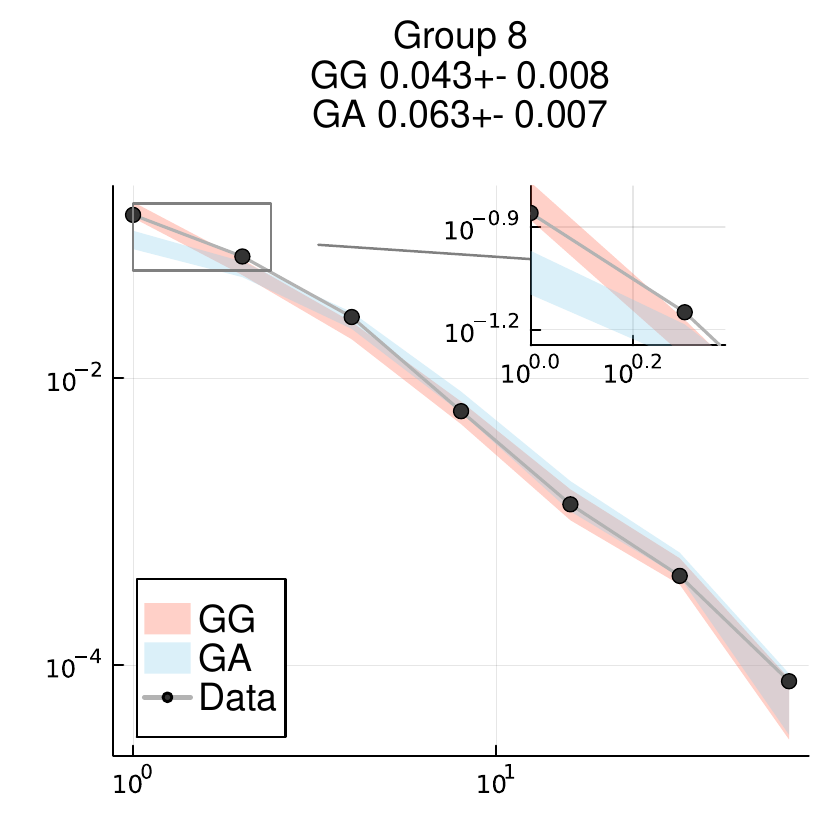}
    \includegraphics[width=0.24\linewidth]{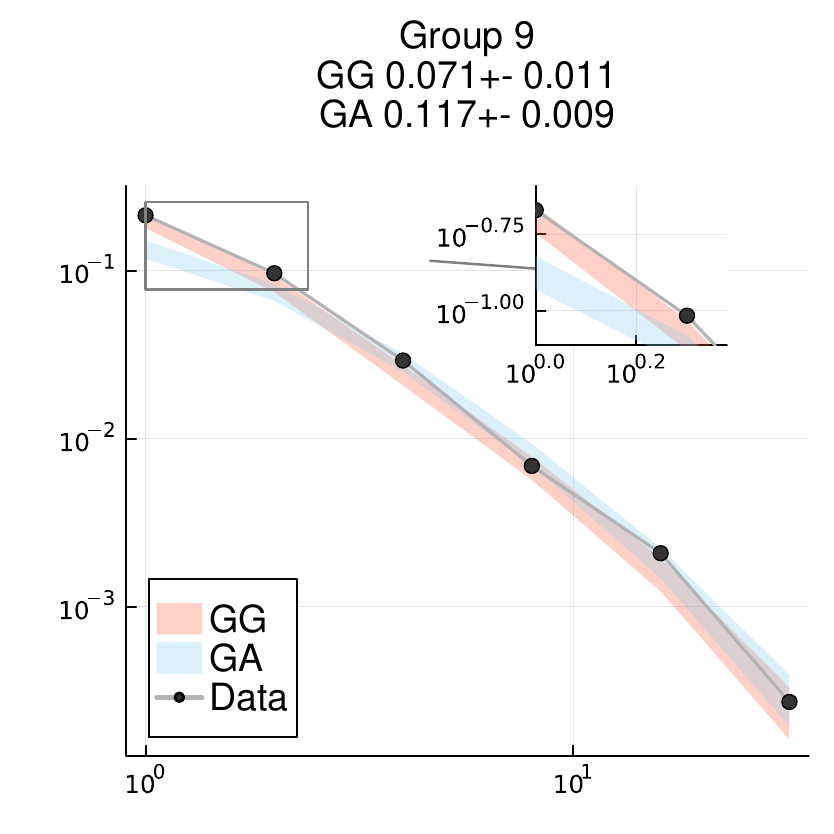}
    \includegraphics[width=0.24\linewidth]{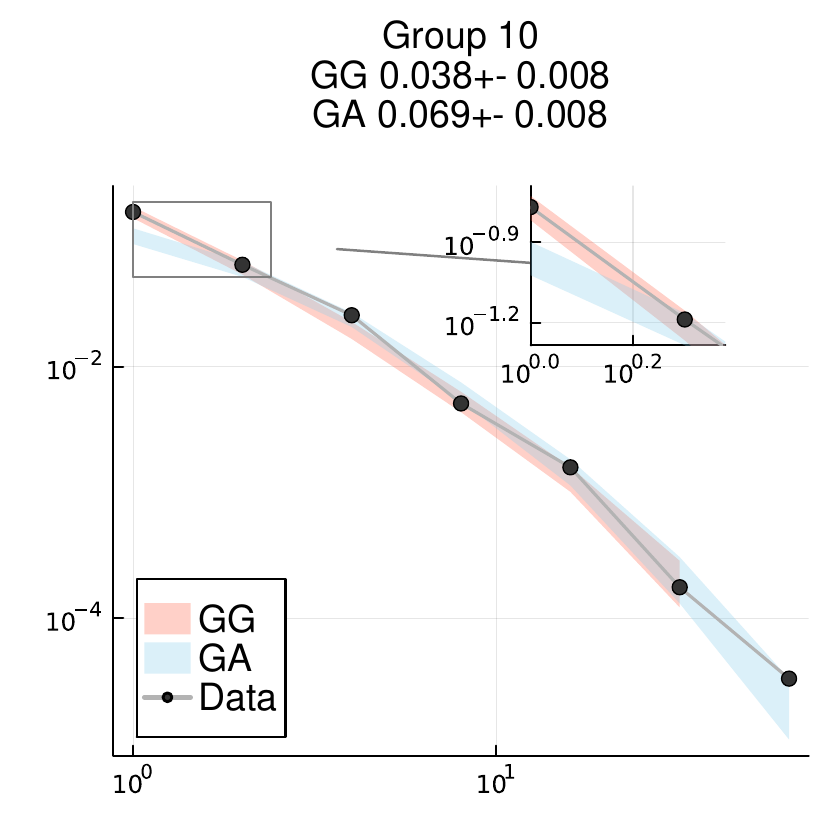}
    \includegraphics[width=0.24\linewidth]{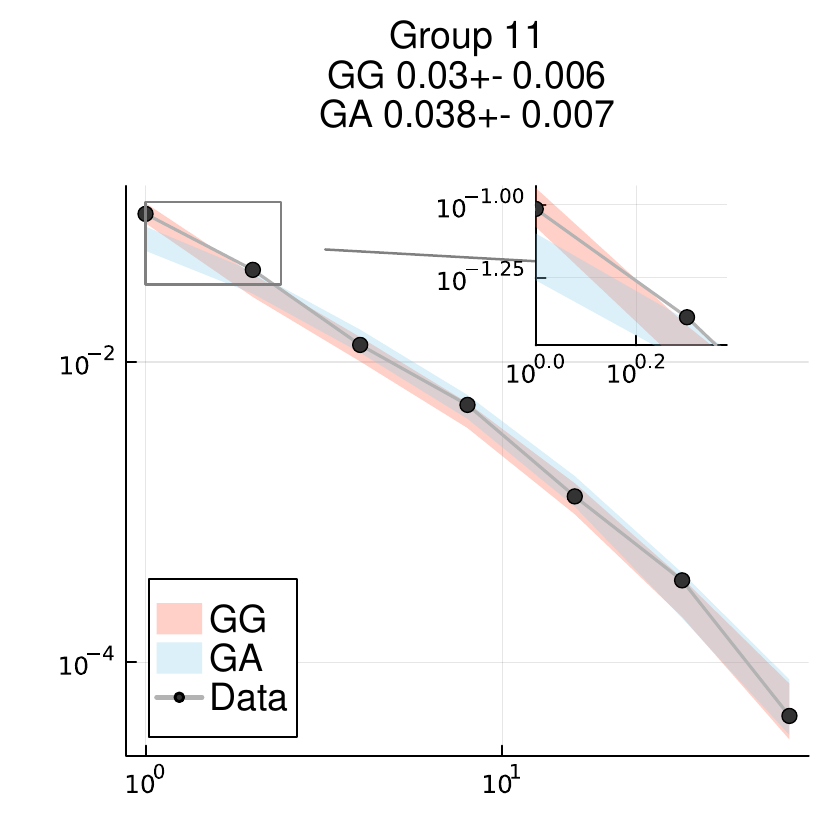}
    \includegraphics[width=0.24\linewidth]{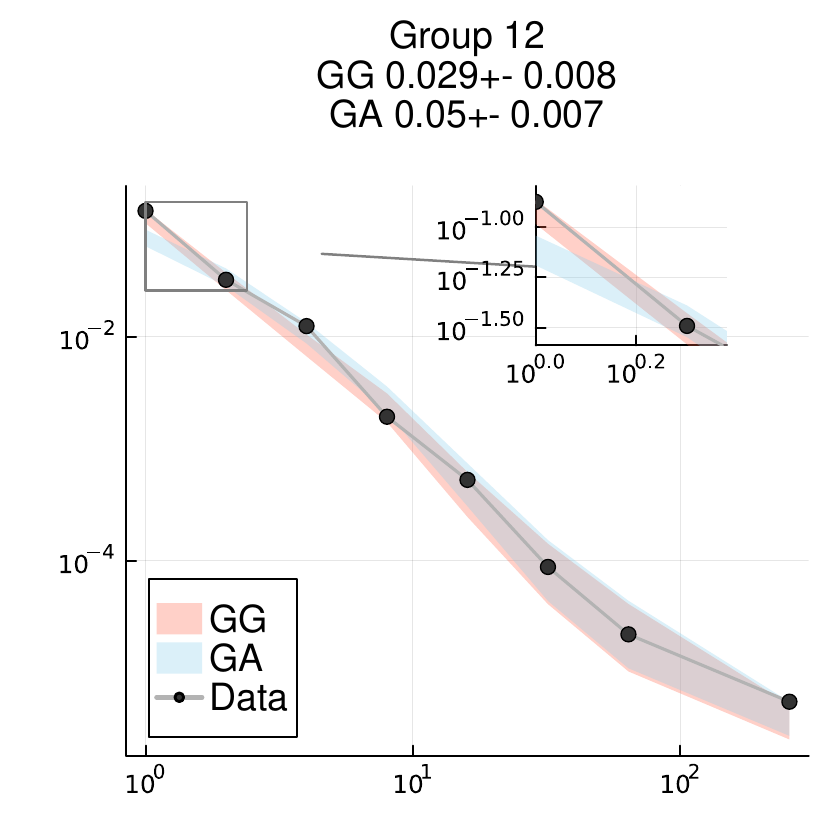}
    \caption{KS statistics between the FoFs of the predicted data and the observed test data.}
    \label{fig:microbiome_cnts_ks}
\end{figure}

\begin{figure}
    \centering
    \includegraphics[width=0.24\linewidth]{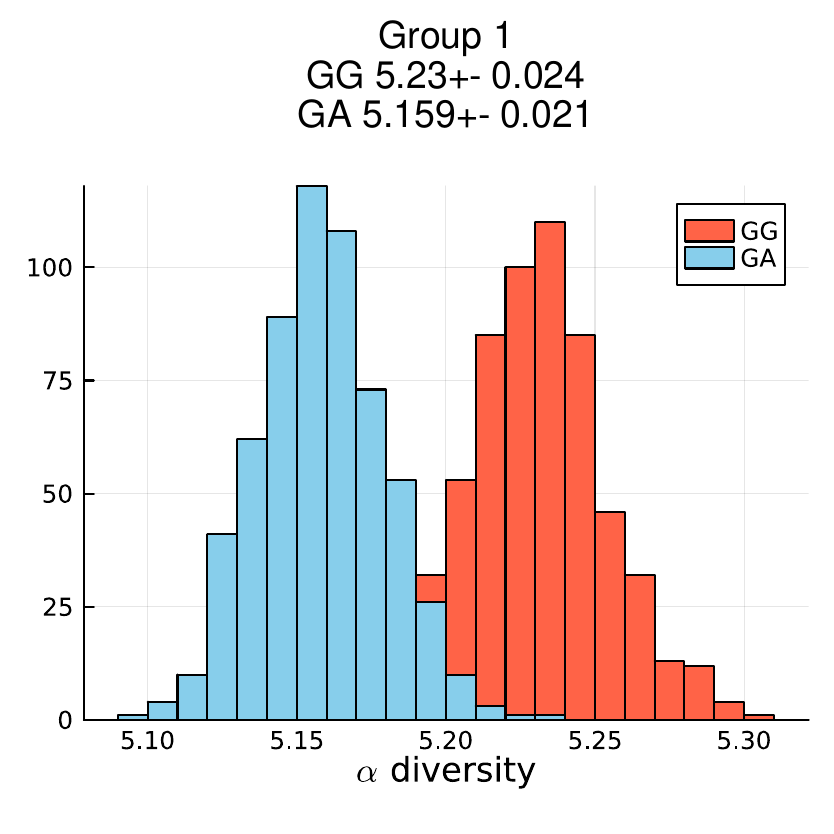}
    \includegraphics[width=0.24\linewidth]{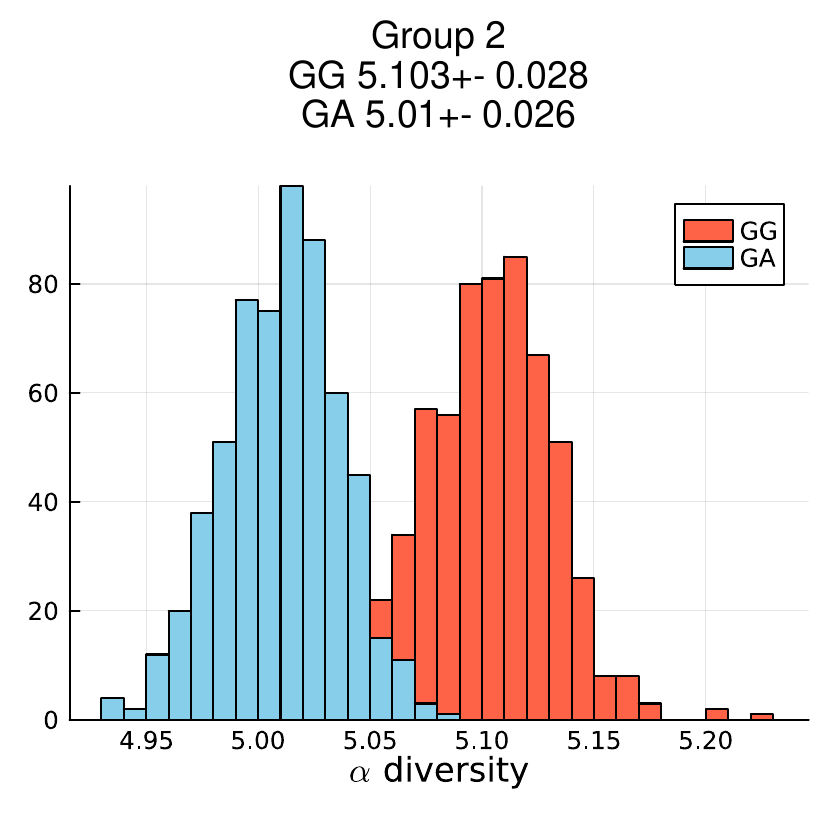}
    \includegraphics[width=0.24\linewidth]{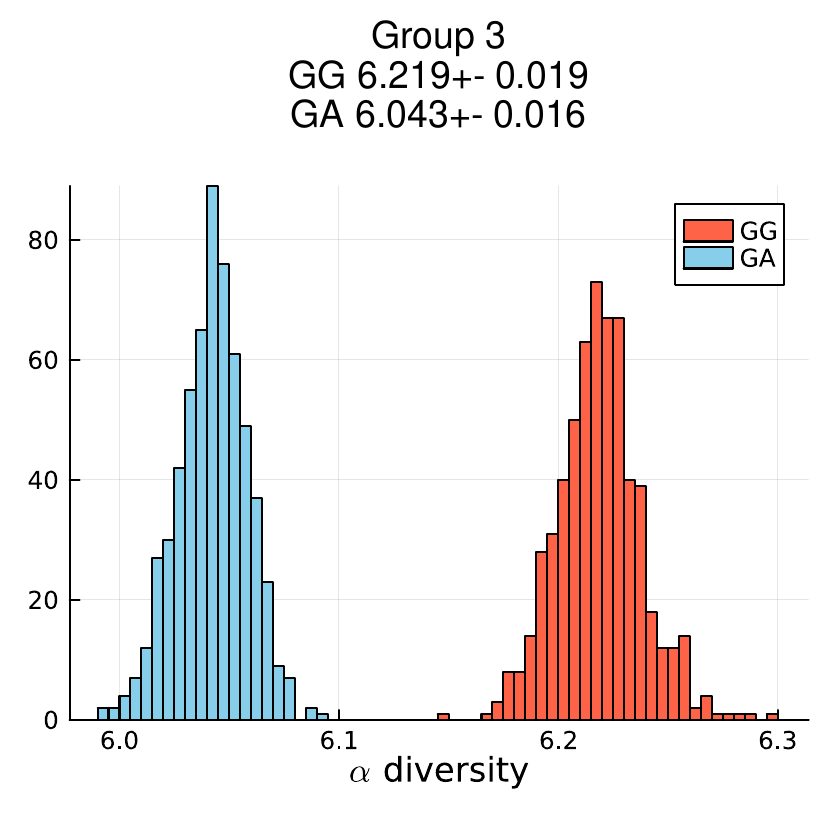}
    \includegraphics[width=0.24\linewidth]{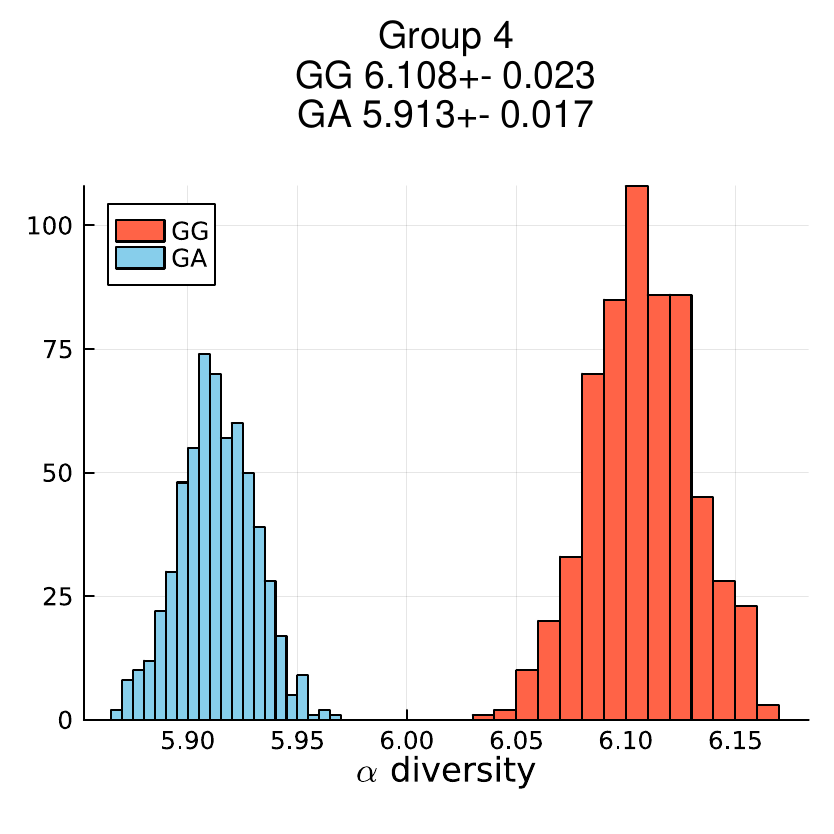}
    \includegraphics[width=0.24\linewidth]{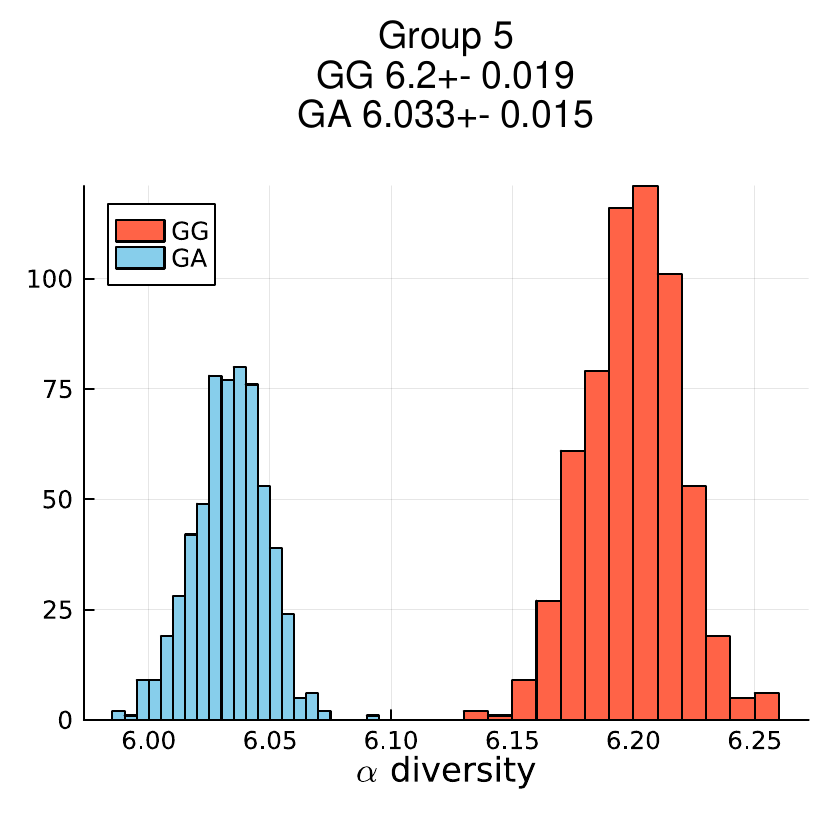}
    \includegraphics[width=0.24\linewidth]{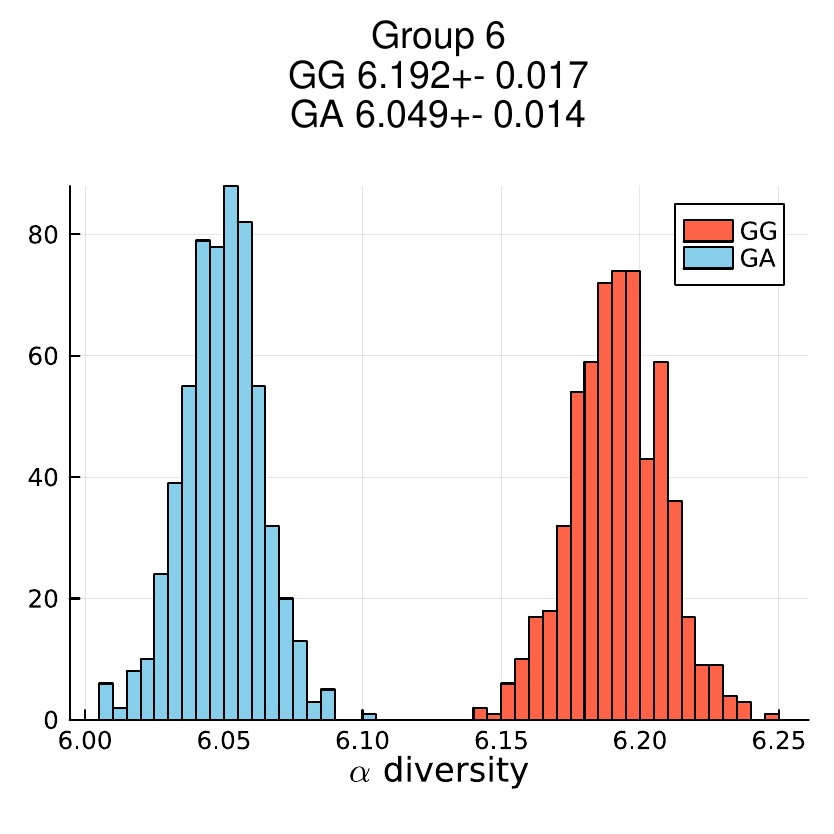}
    \includegraphics[width=0.24\linewidth]{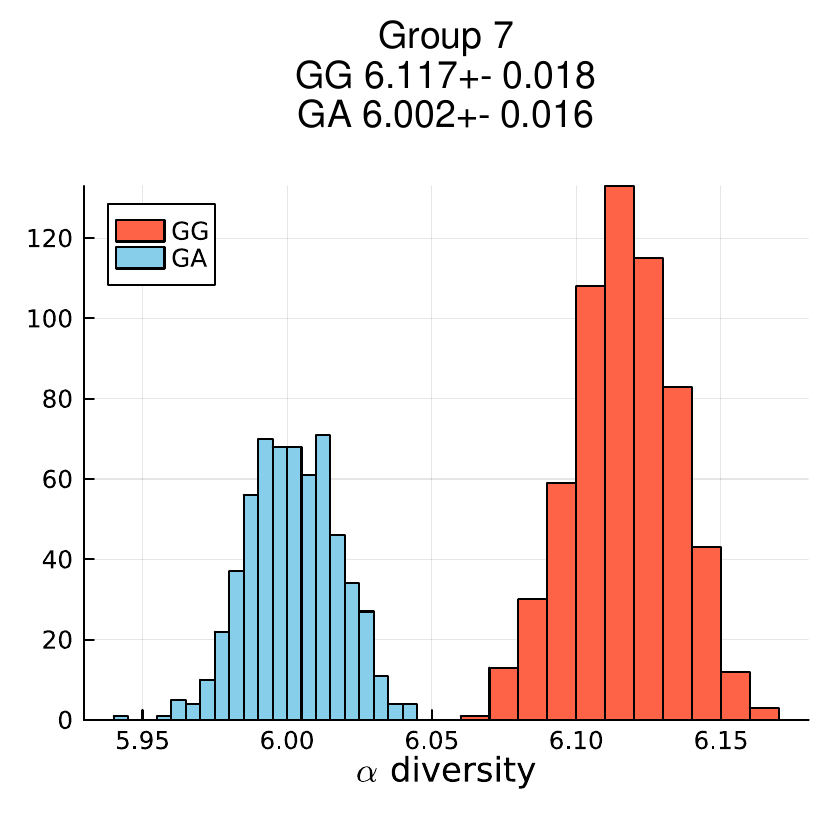}
    \includegraphics[width=0.24\linewidth]{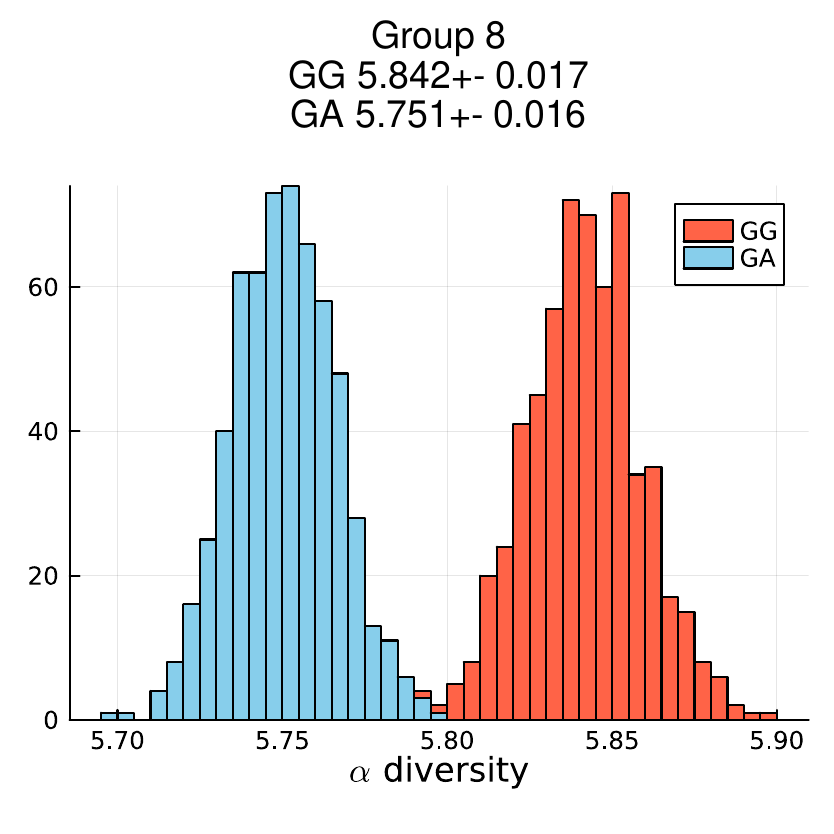}
    \includegraphics[width=0.24\linewidth]{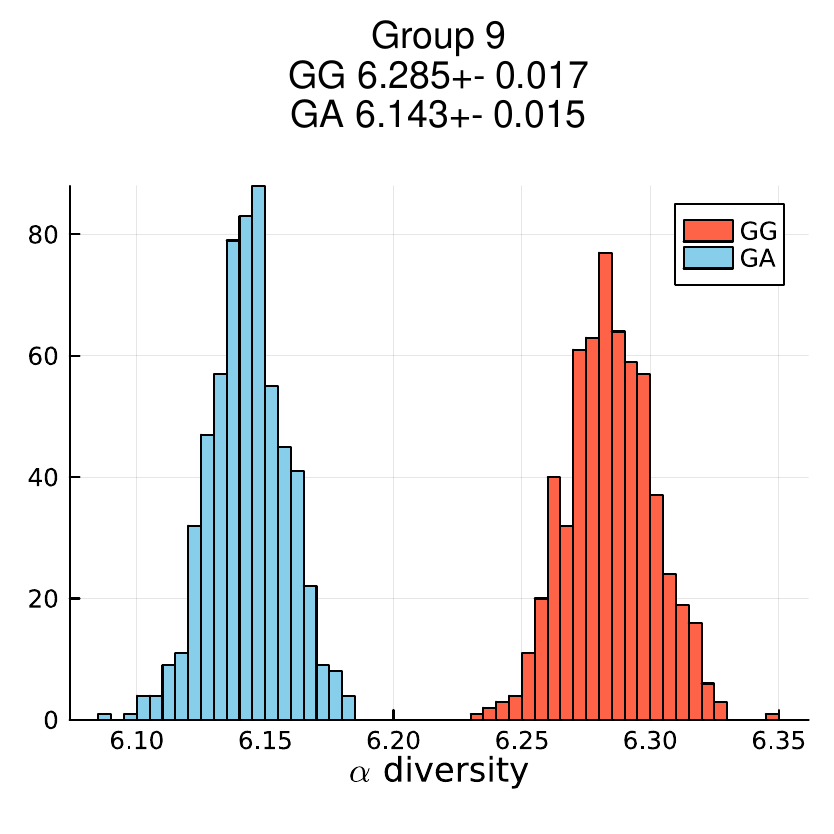}
    \includegraphics[width=0.24\linewidth]{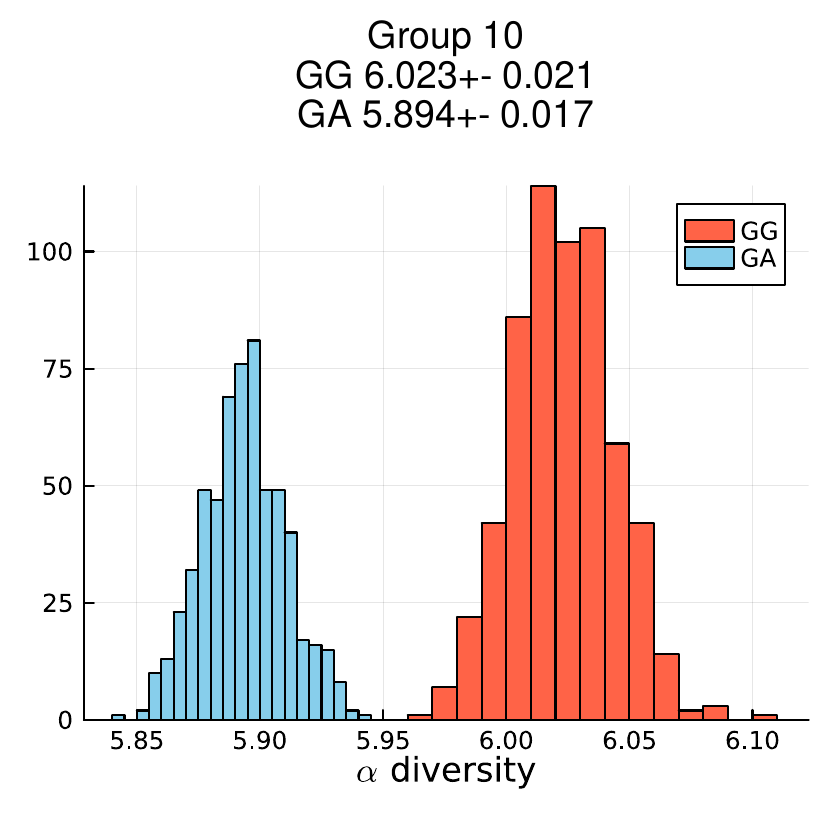}
    \includegraphics[width=0.24\linewidth]{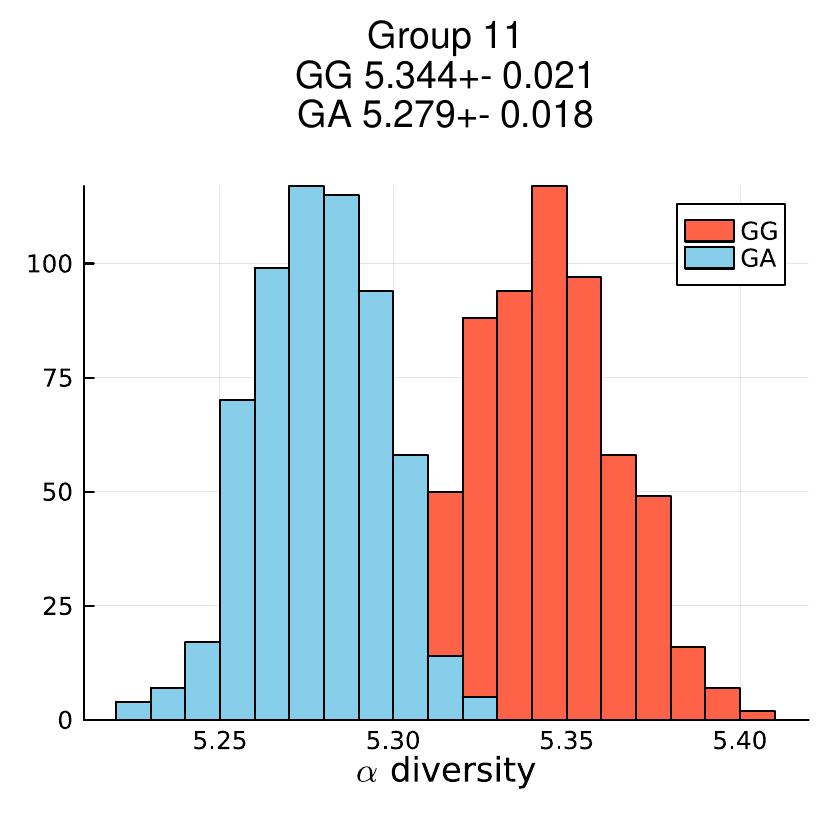}
    \includegraphics[width=0.24\linewidth]{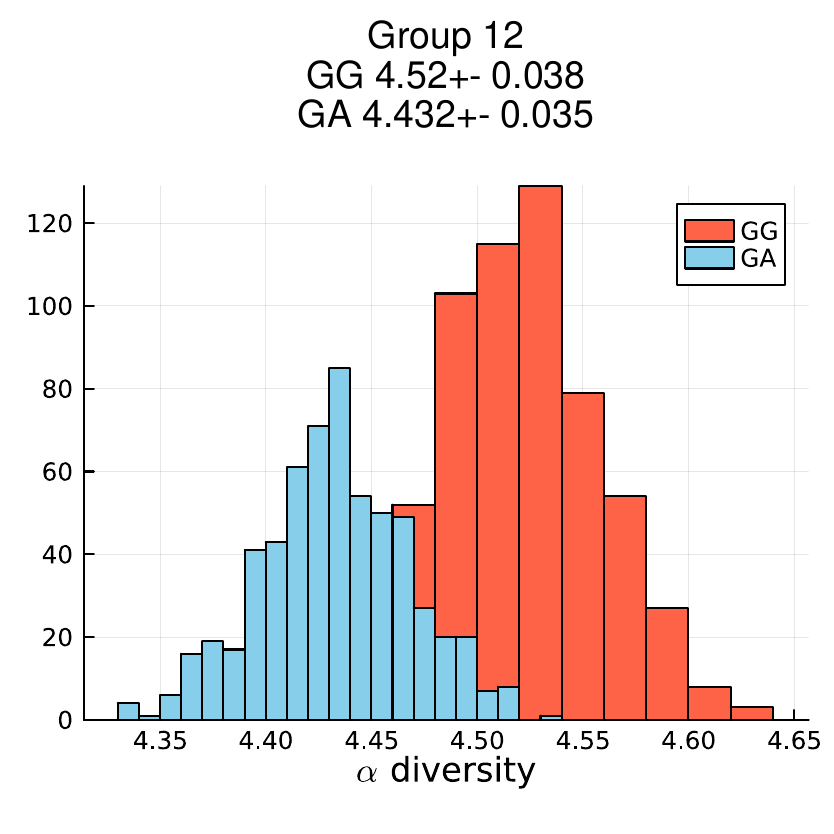}
    \caption{Posterior $\alpha$ diversities measured as Shannon entropies for the microbiome data.}
    \label{fig:microbiome_alpha_divs}
\end{figure}

Figure~\ref{fig:microbiome_test_ll} shows the test log-likelihood values of GG and Gamma HIBP models computed for microbiome data, where GG significantly outperforms Gamma. Figures~\ref{fig:microbiome_posterior_gg} and~\ref{fig:microbiome_posterior_gamma} show the posterior distributions of the parameters inferred by the GG and Gamma HIBP models on the microbiome data. Figure~\ref{fig:microbiome_cnts_ks} presents a group-wise comparison of the FoF distributions predicted by both models against those from the real test data. In Figure~\ref{fig:microbiome_alpha_divs}, we display the posterior distributions of group-wise Shannon entropy derived from the GG and Gamma HIBP models.

\newpage

%\newpage
\bibliographystyle{agsm}
\bibliography{bibliography}

\end{document}